%% file: 3DMPE_arxiv_2026.tex
\documentclass[runningheads]{llncs}

\usepackage{eccv}

\usepackage{eccvabbrv}

\usepackage{graphicx}
\usepackage{booktabs}

\usepackage[hidelinks]{hyperref}

\usepackage{orcidlink}

\usepackage{algorithm}
\usepackage{algpseudocode}
\usepackage{multirow}
\usepackage{tabularray}
\usepackage{adjustbox}

\usepackage{subcaption}
\usepackage{wrapfig}

\DeclareMathOperator{\cD}{\mathcal{D}}
\DeclareMathOperator{\cP}{\mathcal{P}}

\DeclareMathOperator{\cX}{\mathcal{X}}

\DeclareMathOperator{\mD}{\mathbf{D}}

\DeclareMathOperator{\mX}{\mathbf{X}}

\DeclareMathOperator{\RR}{\mathbb{R}}

\begin{document}

% ---------------------------------------------------------------
% TODO REVIEW: Replace with your title
% \title{Author Guidelines for ECCV Submission} 
\title{3DMPE: 3D Multi-Perspective Embedding}

% TODO REVIEW: If the paper title is too long for the running head, you can set
% an abbreviated paper title here. If not, comment out.
\titlerunning{3DMPE: 3D Multi-Perspective Embedding}

% TODO FINAL: Replace with your author list. 
% Include the authors' OCRID for the camera-ready version, if at all possible.
\author{Vahan Huroyan\inst{1}
% \orcidlink{0000-0002-1596-6962} 
\and
Md Rahat-uz-Zaman\inst{2}
% \orcidlink{0000-0001-6728-7569} 
\and
Stephen Kobourov\inst{3}
% \orcidlink{0000-0002-0477-2724}
}

% TODO FINAL: Replace with an abbreviated list of authors.
% \authorrunning{F.~Author et al.}
% % First names are abbreviated in the running head.
% % If there are more than two authors, 'et al.' is used.

% % TODO FINAL: Replace with your institution list.
\institute{Saint Louis University, St. Louis MO 63103, USA \\
\email{vahan.huroyan@slu.edu}
\and
University of Utah, \\
\email{rahat.zaman@utah.edu}
\and
Technical University of Munich, \\
\email{stephen.kobourov@tum.de}
}
\maketitle

\begin{abstract}
We study 3D point cloud reconstruction from multiple partially observed
2D projections. Given two or more projections of an unknown 3D point
cloud, together with cross-view point correspondences and visibility
information, our goal is to recover a consistent 3D configuration when
different views contain different subsets of points. We propose
\emph{3D Multi-Perspective Embedding} (3DMPE), an optimization-based,
training-free method that reconstructs the 3D point cloud and, in the
variable-projection setting, jointly estimates the projection maps.
3DMPE extends Multi-Perspective Simultaneous Embedding to accommodate
missing points and incomplete pairwise distance information across
views. We consider both fixed-projection and variable-projection
settings. Unlike learning-based reconstruction methods that infer shape
from raw images and often depend on training data, 3DMPE operates on
geometric observations with established correspondences and does not
require category-specific training. Experiments on ShapeNet and Pix3D
evaluate reconstruction quality using Chamfer Distance, Earth Mover
Distance, and RMSE-Optimize-Align (ROA), and examine the effects of
initialization, the number of views, point visibility, and several noise
regimes, including noisy distances and erroneous correspondences. The
results demonstrate that 3DMPE can effectively reconstruct point clouds
from partial multi-view geometric observations.

\end{abstract}

% ---------------------------------------------------------------

\section{Introduction}

\begin{figure*}[t]
    \centering
    \includegraphics[width=0.70\textwidth]{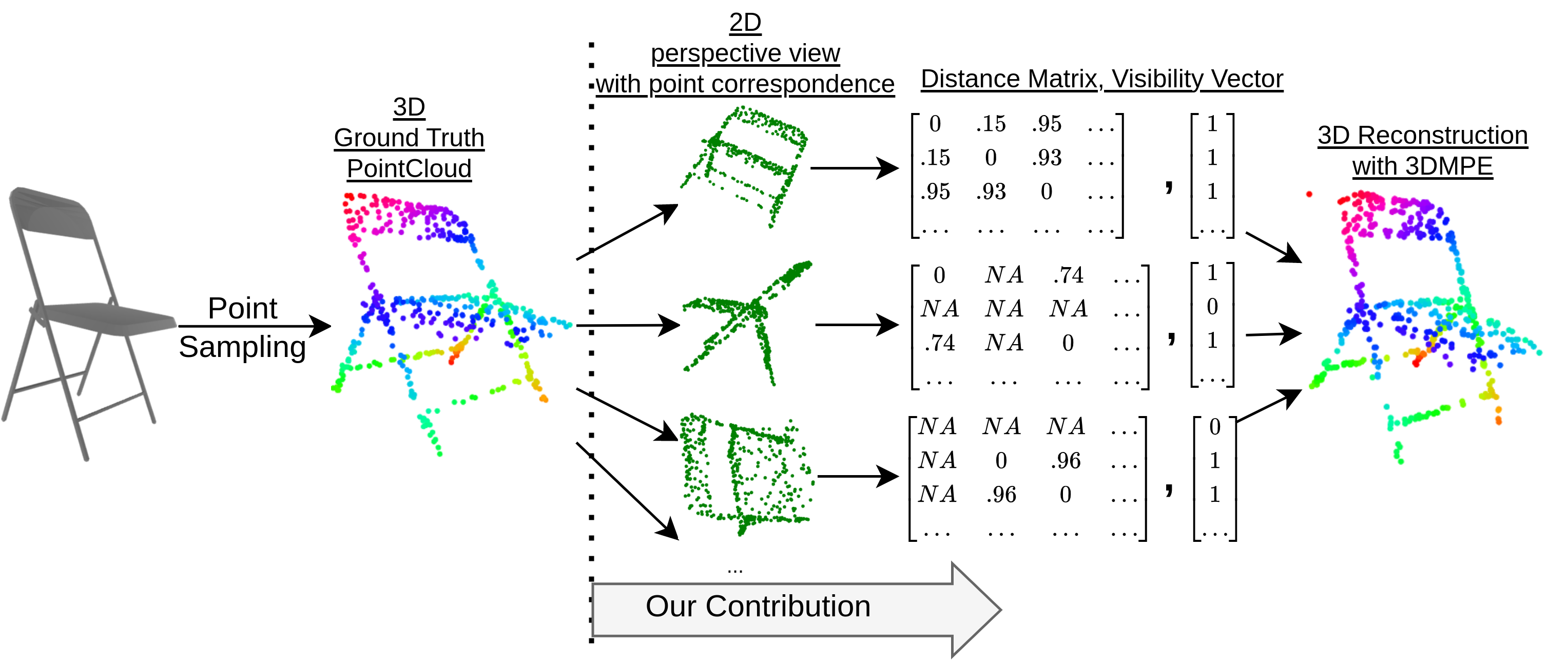}
    \caption{Full pipeline of 3DMPE. Given multiple 2D snapshots and the visibility vectors ($1$ if a point is present in a view and $0$ otherwise), in this example $3$, but the algorithm works for arbitrary number of inputs, we first compute the pairwise distance matrices for each snapshot the algorithm reconstructs the 3D point cloud.}
    \label{fig:contribution}
\end{figure*}

Understanding the 3D structure of objects from limited observations is a fundamental problem in computer vision, with applications in robotics, autonomous driving, precision agriculture, gaming, and virtual and augmented reality. The 3D reconstruction problem aims to infer the three-dimensional structure of an object or scene from one or more 2D images. In general, this problem is ill-posed and highly sensitive to noise and missing information.
In the noise-free case, the problem is called triangulation~\cite{hartley2006multiple}. Yet, noise is unavoidable. Our setup assesses how different noise regimes affect our algorithm's performance.

In practice, 3D reconstruction is often addressed using Structure-from-Motion (SfM) pipelines \cite{ozyesil2017survey, haming2010the, bianco2018evaluating}, which process raw images through feature detection, correspondence matching, camera pose estimation, and triangulation. Modern systems such as COLMAP \cite{schonberger2016structure} provide end-to-end implementations of this pipeline and are widely used in practice.

In this work, we study a more structured reconstruction problem corresponding to the geometric recovery stage after point correspondences have been established. Specifically, the input consists of multiple partially observed 2D projections of an unknown 3D point cloud, together with point-to-point correspondences and visibility information across views. A point may be absent from some views because of occlusion or limited visibility, and consequently the pairwise distance information is incomplete. We do not address feature detection or correspondence estimation from raw images; instead, we study how accurately the underlying 3D point cloud can be recovered once these geometric observations are available.

Our formulation also considers two settings. In the fixed-projection setting, the projection viewpoints are given as part of the input. In the variable-projection setting, the viewpoints are unknown and must be estimated jointly with the 3D point cloud. The goal in both cases is to recover 3D coordinates that are consistent with all observed 2D distances.

To address this problem, we propose \emph{3-Dimensional Multi-Perspective Embedding} (3DMPE), an optimization-based method for reconstructing 3D point clouds from multiple partially observed 2D projections. 3DMPE is inspired by Multi-Perspective Simultaneous Embedding (MPSE) \cite{hossain2021multi}, which jointly embeds multiple datasets described by pairwise distance matrices. However, standard MPSE assumes complete pairwise distance information in every view and therefore cannot be directly applied when points are absent from some projections. We modify the MPSE objective to explicitly account for visibility and missing pairwise distances, thereby enabling reconstruction from partial observations.

Unlike learning-based methods \cite{mandikal20183D, fan2017apoint, wang2024dust3r} that infer 3D shape from raw images, 3DMPE does not require object-specific training data. Instead, it exploits geometric consistency across views once correspondences are available. Thus, the two classes of methods address complementary input settings rather than being directly comparable in all experimental regimes.

Our main contributions are:

\begin{itemize}
    \item We formulate 3D point cloud reconstruction from partially observed 2D projections with known correspondences and visibility information.
    \item We introduce 3DMPE, including fixed-projection and variable-projection formulations that extend MPSE to incomplete distance observations.
    \item We evaluate the method on ShapeNet~\cite{shapenet2015} and Pix3D~\cite{pix3d} and study its behavior under changes in initialization, number of views, point visibility, input size, and several noise models.
\end{itemize}
\cref{fig:contribution} illustrates the 3DMPE pipeline.

% ---------------------------------------------------------------

\section{Related Work}
\label{sec:rel_work}

The 3D reconstruction problem has been studied for several decades. %In different times, there were different directions and suggested solutions to the problem. 
The timeline can be generally split into two periods: focus on understanding the geometry and the structure of the 3D object from given 2D images, followed by deep learning methods; see survey by Han et al.~\cite{han2021image}. 
The first period includes Stereo-based approaches~\cite{hartley2006multiple}, 
PDE-based approaches ~\cite{zhu2022pde}, and SfM~\cite{ozyesil2017survey, haming2010the, bianco2018evaluating, schonberger2016structure}.
%with its subproblems.
% For these type of methods, there is a final step that usually includes a surface reconstruction from point clouds, as discussed by Berger et al.~\cite{berger2017survey}.
For such methods, a final step often involves surface reconstruction from point clouds, as discussed by Berger et al.~\cite{berger2017survey}.

The second period is characterized by deep learning methods, starting with
%by using encoder-decoder based architectures.
%The initial developments of deep learning models included 
voxel-based volumetric representations of 3D objects. 
Wu et al.~\cite{wu20153d} represent a geometric 3D shape as a probability distribution 
% (of binary variables) 
on a voxel grid and uses a convolutional deep belief
network for the reconstruction. 
Wu et al.~\cite{wu2016learning} propose a model by leveraging volumetric convolutional networks and generative adversarial networks.
Gidhar et al.~\cite{girdhar2016learning} propose a TL-embedding network to reconstruct volumetric shapes from RGB images.
Tatarchenko et al.~\cite{tatarchenko2017octree} use an octree representation along with deep convolutional decoder architecture to generate volumetric 3D outputs.
Other directions of voxel-based multi-view and single-view reconstruction are introduced by Kar et al.~\cite{kar2017learning} and Tulsiani et al.~\cite{tulsiani2017multi}.

Recent work in this domain explores surface-based representations.
%for example 3D point clouds. 
%Some works in this are include
Fan et al.~\cite{fan2017apoint} propose a PSGN architecture (loss function and learning paradigm), to predict multiple plausible 3D point clouds from an input image.
Mandikal et al.~\cite{mandikal20183D} propose an auto-encoder based approach, where instead of learning just one reconstruction, they predict multiple reconstructions that are consistent with the input view. Haoqiang et al.~\cite{fan2017apoint}, 
Insafutdinov and Dosovitskiy~\cite{insafutdinov2018unsupervised} train a convolutional network for shape and make predictions from a single image by minimizing the reprojection error.
Lin et al.~\cite{lin2018learning} use 2D convolutional operations for the 3D structure prediction task from multiple viewpoints.
Sridhar et al.~\cite{sridhar2019multiview} propose a dense 3D shape reconstruction from a variable number of RGB views of previously unobserved object instances. 
Mandikal and Babu~\cite{mandikal2019dense} propose a deep pyramidal network for point cloud reconstruction. %The network hierarchically predicts point clouds of increasing resolution.

% {\color{red} 
% Miclea et al.~\cite{8751135}, Cheng~\cite{1716121}, Hamid et al.~\cite{hamid2021stereo}, Altingövde~\cite{ALTINGOVDE2022113460}, Neubert et al.~\cite{932923} have performed 3D reconstruction task from the output of active stereo vision systems. The task is hard in the sense that the input is only limited to two images, but easier in the sense that these two images are taken from a very similar region. The reconstruction task is only limited to the visible surface from where the images are taken.
% } 

Miclea et al.~\cite{8751135}, Cheng~\cite{1716121}, Hamid et al.~\cite{hamid2021stereo}, Altingövde~\cite{ALTINGOVDE2022113460} and Neubert et al.~\cite{932923} study the 3D reconstruction task from the output of active stereo vision systems, where the input is only limited to two images, taken from nearby positions.
Other approaches to this and related problems include Michalkiewicz et al.~\cite{michalkiewicz2020asimple}, who use a linear decoder, obtained from the PCA on the signed distance function of the surface, to obtain the 3D reconstruction. 
Lei et al.~\cite{jiahui2020learning} design a neural network that generates high-quality parametric 3D surfaces that are consistent between the given views. 
Klokov et al.~\cite{klokov2019probabilistic} propose a probabilistic reconstruction network which expresses image-conditioned 3D shape inference through a family of latent variable models.
Choy et al.~\cite{choy20163dr2n2} propose a unified approach for single and multi-view 3D object reconstruction.
Their proposed RNN architecture learns a mapping from a set of images of objects to their 3D shapes. 
Rocco et al.~\cite{Rocco_2017_CVPR}, Seki et al.~\cite{Seki_2017_CVPR}, Zbontar et al.~\cite{Zbontar2015Stereo} and Zhu et al.~\cite{8613838} use deep neural network for matching feature points in images to find correspondences. 
% Such methods can be used in a preprocessing step to generate input for 3DMPE.
Recent methods such as DUSt3R~\cite{wang2024dust3r}, MASt3R~\cite{leroy2024mast3r}, and VGGT~\cite{wang2025vggt} jointly estimate correspondences and geometric information from images. These methods are complementary to 3DMPE and could provide the correspondences or geometric observations required by our optimization-based framework.

% {\color{red} Although not directly relevant, Rocco et al.~\cite{Rocco_2017_CVPR}, Seki et al.~\cite{Seki_2017_CVPR}, Zbontar et al.~\cite{Zbontar2015Stereo} and Zhu et al.~\cite{8613838} used deep neural network for matching images' feature points to find correspondences. These methods can fit as preprocessing steps of images to generate input in 3DMPE.}

One drawback of such deep learning methods is that they  work well only on objects similar to those in the training sets.
% , as illustrated by the following specific examples.
Such several approaches to reconstruct/estimate human hands include~\cite{yu2021local, kulon2019single}.
% where Kulon et al.~\cite{kulon2019single} use mesh convolutions to reconstruct 3D hand shapes, based on a single image.
% Yu et al.~\cite{yu2021local} solve the 3D point cloud reconstruction and 3D pose estimation of
% the human hand based on a single image.
%Another example  bird 3D shape reconstruction problem is Badger et al.~\cite{badger20203d}.
To be successful, such methods need large and diverse training data.
%The need for training data of similar objects is not the only drawback but also these sets need to be large and diverse. 
% {\color{red} 
% These trained models cannot be used to reconstruct an unseen object coming from different distributions. Moreover, deep learning methodologies have their own security vulnerabilities for cyberattacks \cite{barreno2006can, barreno2010security, su2019onepixel}.
% }
% {\color{red} 
% % These trained models cannot be used to reconstruct an unseen object.
% Moreover, deep learning methodologies have their own security vulnerabilities for cyberattacks \cite{barreno2006can, barreno2010security, su2019onepixel}.
% }
We remark, that our approach does not require a training data and we do not learn a specific type of object. Instead, we require multiple snapshots of the object with some registered annotations on it and our proposed algorithm manages to find an embedding for these annotation points in 3D that recover the 3D structure of the underlying object.

\subsection{Multi-Perspective Simultaneous Embedding}
%(MPSE)~\cite{hossain2021multi}}

Here we summarize MPSE~\cite{hossain2021multi}, which aims to visualize datasets and graphs based on multiple pairwise distances between its instances.
The goal of MPSE is to compute coordinates of points in 3D, as well as provide different views into the data, by means of 2D projections that preserve each of the given distance matrices.
Given $K$ matrices $D^{(1)}, \dots, D^{(K)}$ of size $N \times N$ with non-negative entries, also referred to as the \textbf{pairwise distance matrices},
MPSE aims to find $\{x_1, \dots, x_N\} \in \RR^3$
as well as $K$ possible subspaces (or orthogonal projection matrices $P^{(1)}, \dots, P^{(K)}$), such that the distances between $i$-th and $j$-th points in $k$-th pairwise distance matrix $D^{(k)}_{ij}$
is well approximated by the distance between $x_i$ and $x_j$ in the corresponding k-th subspace. 
That is, $\Vert P^{(k)} x_i - P^{(k)} x_j \Vert$  is as close to $D^{(k)}_{ij}$ as possible for $1 \le k \le K$ and $1 \le i, j \le N$.
MPSE achieves this goal with the help of the multi-perspective MDS stress function:
\begin{equation}
\label{eq:MPSE_stress}
    S(\cX,\cP;\cD) = \sum_{k=1}^K \sum_{i > j} \left( \mD^{(k)}_{i j} - \Vert P^{(k)}(x_i) - P^{(k)}(x_j) \Vert \right)^2
\end{equation}
\noindent The minimizer ($\{ x_1, x_2, \dots, x_N \} \subset \RR^3$) of~\eqref{eq:MPSE_stress} is the solution of MPSE  based on the given distance matrices.
%The $\{ x_1, x_2, \dots, x_N \} \subset \RR^3$ minimizer of the MPSE stress function \eqref{eq:MPSE_stress} is the solution of MPSE which is based on the distance matrices by solving the equation. 
There are two variants of MPSE: fixed projections (the projection viewpoints are given) and variable projections (the projection viewpoints are unknown).
With fixed projections the projection subspaces are part of the input
and MPSE only  minimizes~\eqref{eq:MPSE_stress} with respect to $\{x_1, x_2, \dots, x_N\}$.
MPSE with variable projections finds the coordinates $\{x_1, x_2, \dots, x_N\}$ along with best possible subspaces (projections) $P^{(1)}, P^{(2)}, \dots, P^{(K)}$ for which the pairwise distance matrices are well approximated.
%In the variable projection setting both the embedding and viewpoints need to be optimized: \eqref{eq:MPSE_stress} with respect to $\{x_1, x_2, \dots, x_N\}$ and $P^1, P^2, \dots, P^K$.  
% Note, that MPSE cannot be directly used for 3D reconstruction as it assumes that all pairwise distance matrices contain all pairwise distances from every given viewpoint. This is not a realistic assumption for 3D reconstructions as different viewpoints obscure different subsets of points.
% In \cref{sec:mpse_hidden} we show how to use the MPSE idea in this more general setting. \\

% ---------------------------------------------------------------

\section{3D Multi-Perspective Embedding}
\label{sec:mpse_hidden}

In this section, we describe our 3-Dimensional Multi-Perspective Embedding (3DMPE) approach for 3D point cloud reconstruction. Recall, that we assume multiple 2D viewpoints of an unknown 3D point cloud. For each 2D point cloud, we compute pairwise distance matrices for the visible points. As different subsets of points are hidden in different views, we need to handle pairwise distance matrices with missing entries. 
%The reason that we need to handle pairwise distance matrices with missing entries is that for the multi-view 3D reconstruction problem, not all the views have access to all the keypoints, thus the pairwise distance matrices are likely to be incomplete.
A hidden (missing) point implies that the entire corresponding row and column are missing from the pairwise distance matrix.
%We remark, that our proposed algorithm can handle both of the above mentioned cases for missing entries.

We formally define the 3DMPE problem as follows. Let $N$ be the number of points in the unknown 3D point cloud that 3DMPE must reconstruct, given $K$ 3D viewpoints captured by pairwise distance matrices between subsets of points, $\mD^{(1)}, \dots \mD^{(K)}$ (different entries missing in different matrices due to occlusion).  
Define a visibility vector $\alpha^k \in \RR^N$ for each view $1 \le k \le K$, where $\alpha^{(k)}_i = 1$ if the $i$-th keypoint is present in the $k$-th view and $\alpha^{(k)}_i = 0$ otherwise. 3DMPE
% looks for a solution that 
minimizes the stress function  below:
\begin{equation} 
\label{eq:MPSE_hidden_points}
S_{3DREC}(\cX,\cP;\cD) = \sum_{k = 1}^K \sum_{i > j} \alpha^{(k)}_{i} \alpha^{(k)}_{j} \left( \mD^{(k)}_{i j} - \Vert P^{(k)}(x_i) - P^{(k)}(x_j) \Vert \right)^2.
\end{equation}
Similar to MDS and MPSE, 3DMPE is also a non-convex problem. 
We remark, that if 
% points the 
$i$-th and $j$-th points are present in $k$-th view then both $\alpha^k_i$ and $\alpha^k_j$ are $1$ and their product is also 1. Otherwise (when either or both $i$-th or $j$-th points are missing from the $k$-th view), $\alpha^{(k)}_{i} \alpha^{(k)}_{j} = 0$.

The optimization problem in \eqref{eq:MPSE_hidden_points} can be considered in two different scenarios. The first and more simple scenario, which we refer to as reconstruction for fixed projections, assumes that we are given the viewpoints, that is we are given the projection matrices $P^{(1)}, \dots, P^{(K)}$ and we only need to optimize over the $x_1, \dots, x_N$. The optimization problem for this case is 
% {\scriptsize
\begin{equation}
    \label{eq:3Drec_fixed}
    \arg\min_{\cX} S_{3DREC} (\cX,\cP;\cD)
\end{equation}
% }
The second scenario is more complex, we refer to it as reconstruction for varying projections, which assumes that the projection matrices $P^{(1)}, \dots, P^{(K)}$ are also variable:
% The more complex optimization problem becomes:
% {\scriptsize
\begin{equation}
    \label{eq:3Drec_varying}
    \arg \min_{\cX, \cP} S_{3DREC} (\cX,\cP;\cD)
\end{equation}
% }
To solve \eqref{eq:3Drec_fixed} and \eqref{eq:3Drec_varying} we use Stochastic Gradient Descent (SGD) \cite{Robbins1951AStochastic} with smart initialization,
% from~\cite{hossain2021multi} 
which is also used as baseline; see \cref{sec:evaluation}. 
% All the parameters and implementation details are included in \cref{sec:3Drec_algorithm}. 
% which also summarizes the full algorithm that we use to solve \eqref{eq:3Drec_fixed} and \eqref{eq:3Drec_varying}.

% \todo{discuss and analyze smart initialization}

\subsection{Algorithm}
\label{sec:3Drec_algorithm}
%{\color{red} In this section, we summarize the algorithm of 3DMPE.
% MPSE with hidden points for 3D reconstruction. 
\begin{algorithm}[t]
\caption{3DMPE for Fixed Projections by SGD}
\label{algo:fixed}
\begin{algorithmic}
\Require Pairwise Distance Matrices: $\cD = \{ \mD^{(1}), \dots, \mD^{(K)}$, \\
Visibility vectors $\{v^{(1)}, \dots v^{(K)} \}$, \\
Initial embedding $\mX_0$, initial learning rate $\mu_0$, stochastic constant $c$, iteration number $T$, projection matrices $P$.
\For{$t=1,2,\dots,T$}
\State $\xi_t \sim \Xi(\mD, c)$
\State $\mX_t = \mX_{t-1} - \mu_{t-1} \nabla_{\mX} S^2_{\xi_t} (\mX_{t-1},\mathbf{P};\mathbf{D})$ 
\State $\mu_t = \mu(\mX_{t-1}, \mX_t,\xi_t)$
\EndFor
\Ensure $\mX_T$
\end{algorithmic}
\end{algorithm}
The 3DMPE algorithms (fixed projections and variable projections) are summarized in \cref{algo:fixed} and \cref{algo:varying}, respectively. The algorithmic details are  discussed in \cref{sec:algorithmic_details}.
We use smart initialization
for the initial 3D embedding and minibatch SGD with early termination~\cite{yao2007early} on the gradient and stress values. Implementation details are discussed in \cref{sec:implementaion}, dataset details are in \cref{sec:dataset} and evaluation details are in \cref{sec:evaluation}.

\begin{algorithm}[t]
\caption{3DMPE with varying perspectives by SGD}
\label{algo:varying}
\begin{algorithmic}
\Require Pairwise Distance Matrices: $\cD = \{ \mD^{(1)}, \dots, \mD^{(K)}\}$, \\
Initial embedding, perspective parameters: $\mX_0$, \\ $\mathbf{Q}_0$, 
Visibility vectors $\{v^{(1)}, \dots v^{(K)} \}$,
Initial learning rates $\mu_{\mX}$, and $\mu_Q$, 
Stochastic constant: $c$, iteration number: $T$.
\For{$t=1,2,\dots,T$}
\State $\xi_t \sim \Xi(\mD, c)$
\State Compute $\nabla S^2_\xi (X_{t-1},\mathbf{Q}_{t-1})$ and $\nabla S^2_\xi (X_{t},\mathbf{Q}_{t})$
\State Compute $\mu_t$.
\State $\mX_t = \mX_{t-1} - \mu_{X,0} \nabla_{\mX} S^2_\xi (\mX,\mathbf{Q}_t)$.
\State $\mathbf{Q}_t = \Pi(\mathbf{Q}_t - \mu_{\mathbf{Q}_t} \nabla_Q S^2_\xi)$
\State $\mu^X_t = \mu(\mX_{t-1}, \mX_t,\xi_t)$ 
\State $\mu^Q_t = \mu(\mathbf{Q}_{t-1},\mathbf{Q}_t,\xi_t)$
\EndFor
\Ensure $X_\textrm{final}$, $\mathbf{Q}_\textrm{final}$.
\end{algorithmic}
\end{algorithm}

\subsection{Details of Algorithms~\ref{algo:fixed} and \ref{algo:varying}}
\label{sec:algorithmic_details}

Algorithms~\ref{algo:fixed} and~\ref{algo:varying} are described in \cref{sec:3Drec_algorithm} and here we provide some additional details.
%As we have already mentioned in \cref{sec:3Drec_algorithm}, 
Similar to MPSE~\cite{hossain2021multi}, the objective functions for fixed projections~\eqref{eq:3Drec_fixed} and varying projections~\eqref{eq:3Drec_varying} are non-convex. Thus, a gradient descent algorithm is not guaranteed to find the global minimum.
A good implementation of  SGD, however, has been shown to consistently find high quality solutions~\cite{zheng2019graph}.
%(possibly a good enough point of local minima).
We use SGD with smart initialization to solve the optimization problems in~\eqref{eq:3Drec_fixed} and~\eqref{eq:3Drec_varying}.
SGD uses the following sampling mechanism: First, we define the following random variable $\xi \sim \Xi(\mD, c)$, which samples an incomplete pairwise dissimilarity matrix based on a given dissimilarity matrix $\mD$. That is, $\xi$ itself is a pairwise distance matrix, for which,  each entry of $\mD$ is included in $\xi$ with probability $c$; otherwise, it is $0$.
Based on this sample from the pairwise distance matrix, we define:
\begin{equation}
S_{\xi} (\cX, \cP; \cD) = S_{3DREC} (\cX, \cP; \xi).
\end{equation}
Note, that here $\xi = \left[ \xi^1, \dots, \xi^K \right],$ and for $1 \le k \le K$, $\xi^k \sim \Xi(\mD^k, c)$.
Similar to~\cite{hossain2021multi}, the adaptive step size is chosen by adaptive scheme~\cite{malitsky2020adaprive}. For Algorithm~\ref{algo:fixed}, the adaptive step size $\mu_{\mX}$ is defined as:
\begin{multline}
\mu_{\mX} ({\mX}_t,{\mX}_{t-1},\xi_{t}) =\\
\frac{ \left( \mX_t - \mX_{t-1}\right)^T \left(\nabla_{\mX} S_{\mathbb{\xi}_t}^2 (\mX_t, \cP; \cD) - \nabla_{\mX} S_{\mathbb{\xi}_t}^2(\mX_{t-1}, \cP; \cD)\right)}
{ \left \Vert \nabla_{\mX} S_{\mathbb{\xi}_t}^2(\mX_t, \cP; \cD) - \nabla_{\mX} S_{\mathbb{\xi}_t}^2(\mX_{t-1}, \cP; \cD) \right \Vert^2}, 
\end{multline}
where $\mX_t, \mX_{t-1}$ are the embeddings at steps $t$ and $t-1$, $\xi_t$ are the sampled pairwise distance matrices at iteration $1 \le t \le T$ and 
$\nabla_{\mX} S_{\mathbb{\xi}_t}^2 (\mX_t)$ 
is the gradient of $S_{\mathbb{\xi}_t}^2$ at iteration $1 \le t \le T.$
For Algorithm~\ref{algo:varying}, $\mu_t^{Q}$ is defined similarly.

\subsection{Baseline Approach and Evaluation Metrics}
\label{sec:evaluation}

We propose the following basic 3D point cloud reconstruction idea from given 2D projections as a baseline approach:
Given multiple pairwise distance matrices, we suggest computing their average and using Multi-Dimensional Scaling (MDS)~\cite{shepard1962analysis} to embed the point cloud in 3D based on the combined proximity distance matrix. We refer to this reconstruction as the \textbf{baseline} reconstruction and use it to compare against 3DMPE. 
To initialize the 3DMPE algorithm for stochastic gradient descent (SGD), we have two options: random initialization, referred to as 3DMPE with random initialization, or use the baseline method to initialize the SGD, which we call 3DMPE with smart initialization. We will compare these two approaches in later sections.
In \cref{sec:numerical_initialization} we numerically demonstrate the effect of smart initialization on the performance of 3DMPE.

We use two standard metrics to evaluate the quality of the 3D point cloud reconstruction when compared to the underlying 3D point cloud:

\noindent The Earth Mover Distance (EMD)~\cite{rubner2000earth} between point clouds $X$ and $\hat{X}$ is defined as:
\begin{equation}
\label{eq:EMD}
d_{EMD}(X, \hat{X}) = \min\limits_{\phi: X \xrightarrow{} \hat{X}} \sum_{x \in X} \Vert x-\phi(x)\Vert_2 
\end{equation}
where $\phi: X \xrightarrow{} \hat{X}$ is a bijection.

\begin{figure}[t]
    \centering
    \includegraphics[width=0.8\linewidth]{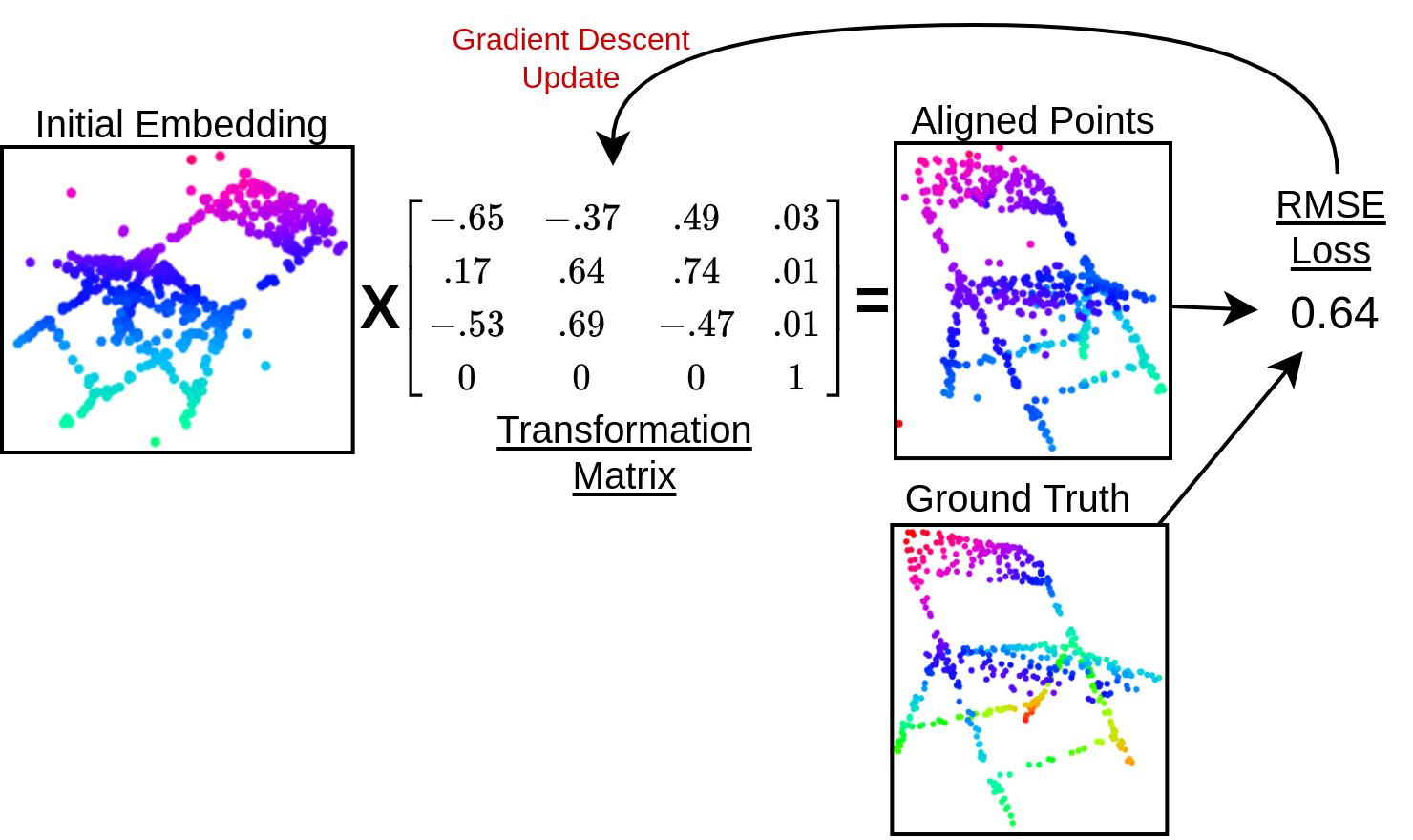}
    \caption{Pipeline for ROA metric. Here 0.64 is the ROA value for current iteration of gradient descent update loop.}
    \label{fig:roa}
\end{figure}

\noindent The second one is the Chamfer Distance (CD) between  point clouds $X$ and $\hat{X}$ is defined as:
\begin{equation}
\label{eq:chamfer} 
d_{CD} (X, \hat{X}) 
= \sum_{x \in X} \min\limits_{\hat{x} \in \hat{X}} \Vert x-\hat{x} \Vert_2^2 + \sum_{\hat{x} \in \hat{X}} \min\limits_{x \in X} \Vert x-\hat{x} \Vert_2^2.
\end{equation}
Note that neither metric (EMD and CD) takes into account  rotations and translations, thus requiring an initial alignment of the two point clouds.
%the output of the algorithm with the ground truth and then apply these metrics for evaluation.
The Iterative Closest Point (ICP)~\cite{Zhang2020} algorithm is often used for such alignments, e.g., ~\cite{mandikal20183D}. As ICP can 
% is known to 
get stuck in local minima, we improve the alignment by
% with the help of 
point-to-point correspondence information. Specifically, we align 4 points using~\eqref{eq:4pointalign}.
\begin{equation}
\label{eq:4pointalign}
    T = AX^{-1},
\end{equation}
where $A$ and $X$ are the 4 selected points from the two point clouds in $\RR^{4 \times 4}$ and $T$ is the corresponding transformation matrix in $\RR^{4 \times 4}$ for homogeneous points.
Using RANSAC~\cite{Fischler1981RANSAC}, this process is repeated 1000 times, by selecting $T$ that minimizes~\eqref{eq:dist}:
\begin{equation}
\label{eq:dist}
\mathcal{L}(x, y) = \frac{\sum{||x_i - y_i||_2}}{n}
\end{equation}
In addition to the EMD and CD metrics, we also use a customized RMSE-Optimize-Align (ROA) metric; see \cref{fig:roa}. Given two point clouds, $X$ and $\hat{X}$ we compute the Root-Mean-Square Error (RMSE) for all corresponding pairs of points, with the best possible homogeneous space transformation matrix $T$ applied to the points of $\hat{X}.$
The optimization problem to align the points can be written as:
\begin{equation}
\label{eq:rao}
RMSE^2 = \min\limits_T \frac{1}{N} \sum\limits_{i=1}^{N} \Vert X_i - T \hat{X}_i \Vert_2^2.
\end{equation}
% where $T$ is the homogeneous space transformation matrix.
The problem can be tackled with SGD on the objective function.
%We coin the name of this metric as RMSE-Optimize-Align (ROA) metric and report it for our experiments. 
%An overview of the pipeline of this metric calculation is demonstrated in \cref{fig:roa}. 
A closed form solution to RMSE can be given by SVD; see the Appendix.

% ---------------------------------------------------------------

\input{tables/allimages.tex}

\section{Experiments}
\label{sec:experiments}

We evaluate 3DMPE in both the fixed and variable-projection settings on the ShapeNet and Pix3D datasets. Our experiments address four questions: 
(i) how accurately does 3DMPE reconstruct 3D point clouds relative to our MDS-based baseline; 
(ii) how do initialization, the number and diversity of views, point visibility, and point-cloud size affect reconstruction quality and runtime;
(iii) how robust is the method to noisy distances and erroneous correspondences; and 
(iv) how can 3DMPE be used within an image-based reconstruction pipeline.

Unless stated otherwise, we run minibatch SGD for at most $300$ iterations and terminate early when either the objective value or gradient norm falls below $(10^{-4})$. 
We use Chamfer Distance (CD), Earth Mover Distance (EMD), and ROA, as defined in \cref{sec:evaluation}, to evaluate reconstruction quality. Our primary controlled comparison is with the proposed MDS-based baseline. We additionally include published results for learning-based methods and an exploratory comparison with COLMAP for context; these methods use different inputs and assumptions and are therefore not directly comparable to 3DMPE.

\subsection{Datasets and Experimental Setup}
\label{sec:dataset}
 
In the experiments reported here, we use the normalized mesh of objects (\emph{.obj} format) from the \textbf{ShapeNET}~\cite{shapenet2015} and \textbf{Pix3D}~\cite{pix3d} datasets. From each 3D object, we uniformly sample dense point clouds of as many points as required by the specific experiment (see~\cite{weisstein1999triangle} for details).
Next, we specify the rotation matrices for each projection. The general rotation matrix $R$ by applying rotation angles $\alpha$, $\beta$, and $\gamma$ in x, y, z axis is given in~\eqref{eq:rotation}. 
For the sake of brevity we use $c\theta$ and $s\theta$ to denote $\cos(\theta)$ and $\sin(\theta)$, respectively.

% {\scriptsize
\begin{equation}
% R &= R_z(\alpha) R_y(\beta) R_z(\gamma) \label{eq:rotation} \\
% \label{eq:rotation}
% R &= 
% \begin{bmatrix}
% c\alpha & -s\alpha & 0\\
% s\alpha & c\alpha & 0\\
% 0 & 0 & 1
% \end{bmatrix}
% \begin{bmatrix}
% c\beta & 0 & s\beta \\
% 0 & 1 & 0\\
% -s\beta & 0 & c\beta
% \end{bmatrix}
% \begin{bmatrix}
% 1 & 0 & 0\\
% 0 & c\gamma & -s\gamma \\
% 0 & s\gamma & c\gamma 
% \end{bmatrix}\\
\label{eq:rotation}
R =
\begin{bmatrix}
c\alpha \cdot c\beta & c\alpha \cdot s\beta \cdot s\gamma - s\alpha \cdot c\gamma & c\alpha \cdot s\beta \cdot c\gamma + s\alpha \cdot s\gamma\\
s\alpha \cdot c\beta & s\alpha \cdot s\beta \cdot s\gamma + c\alpha \cdot c\gamma & s\alpha \cdot s\beta \cdot c\gamma - c\alpha \cdot s\gamma \\
-s\alpha & c\beta \cdot s\gamma & c\beta \cdot c\gamma 
\end{bmatrix}
\end{equation}
% }

Assuming that we need to generate $K$ projections from the angle range $[\theta_s,\theta_e]$ in all axes, we need $3$ angles, $\alpha_k$, $\beta_k$ and $\gamma_k$, for $1 \le k \le K$. We assume the band of angle in all axes is $\theta_r = \theta_e - \theta_s$. 
For $1 \le k \le K/2$, we randomly select the angles from range $[(k - 1) * \theta_r / K, k * \theta_r / K]$. 
We do the same for $K/2 < k \le K$ but for negated angles, so the range is $[-((k - K/2) - 1) * \theta_r / K, -(k - K/2) * \theta_r / K]$
(when $K$ is odd, we apply the same process for $K + 1$ and discard the last one). 
We also record the negations of these angles, as they are the input for 3DMPE with fixed projections. Unless stated otherwise, $\theta_s = 0^{\circ}$ and $\theta_e = 360^{\circ}$.
% In ideal cases, $[\theta_s,\theta_e]$ should span all around the point cloud (so $\theta_r = 360^{\circ}$).

We utilize the raytracing algorithm~\cite{Shirley2000Realistic} to the transformed point clouds~\cite{deul2010raytracing} to determine which points will be visible from a specified projection. Note, that not all points are visible in every projection; the ray tracing process helps us identify the visible points. 
With the rotation matrices already established, we position the rotated point cloud away from the origin while maintaining the viewpoint at the origin, casting rays toward the positive $z$-axis.
% Instead of specifying the viewpoints, we use this non-trivial approach as it makes 3DMPE simpler and more efficient for inputting the rotation matrices to 3DMPE (see last input of algorithm \ref{algo:fixed}).
% Finally, we 
We compute the pairwise distance matrices from the output 
% (2D projected point clouds) 
of the raytracing algorithm.

Before raytracing, we assign unique IDs to each point in the point cloud. During raytracing, we keep track of those IDs and keep them for the 2D projections. This way, point-to-point correspondence is reserved. 
The presence of a point in a perspective is recorded in a visibility vector ($1$ for presence and $0$ for absence); see \cref{fig:contribution}.

\subsection{Experiments on ShapeNet}
\label{sec:quantitative}

We run 3DMPE on ShapeNet and report the results
below. 
% We also experimented with 
% The results for Pix3D are in the supplementary material. 
For comparison, we identify the 3D objects used in the experiments of \cite{mandikal20183D} and use them in both our fixed and variable projections settings.
%After 3D reconstruction we use 
%Our chosen values for both minimum cost and minimum gradient, discussed in Section~\ref{sec:3Drec_algorithm}, are 0.0001. A maximum of 300 iteration of the minibatch stochastic gradient descent algorithm is used. The input projections are normalized.
%We run experiments for each object for both fixed projections as well as variable projections. 
% We run the algorithm on both the ShapeNet and Pix3D datasets with the following choice of parameters. The initial embedding for the MPSE algorithm is created randomly. For most of the experiments, we use batch size of 2048 3D points because we found it to converge fastest. Cylinder projection with smart initialization proposed in \cite{hossain2021multi} is used. We stop the MPSE algorithm based on the criteria of reaching a minimum gradient of 0.0001 or minimum stress function value of 0.0001. A maximum of 300 iterations in the MPSE algorithm is used and we take the point cloud embedding of the last iteration. We found to have better reconstruction accuracy if the input projected points are normalized. An adaptive learning rate method from \cite{DBLP:journals/corr/abs-1212-5701} is used for batch stochastic gradient descent algorithm of MPSE. All the experiments with these parameters are run twice, once with fixed angle projection with algorithm \ref{algo:fixed} and once with variable angle projection with algorithm \ref{algo:varying}. 
\cref{tab:allimages} demonstrates a qualitative comparison of 3DMPE alongside other 3D reconstruction algorithms (with different setups), baseline, as well as the ground truth.
Visual inspection suggests that 3D-LMNet and 3DMPE with fixed projection outperform the rest.
%both the baseline and PSGN and are visually very close the ground truth. 
%Note, that there is a slight orientation difference for each point cloud, which is due to the fact that the solutions of the 3D point cloud reconstruction problem is up to a global orientation.
%We remark that the solution to this problem is not necessarily unique even when factoring translation and rotation.
%
% One advantage of 3DMPE is that the algorithm does not assume certain parts of the point cloud, instead, it estimates all parts equally for the reconstruction process. 
% This can be noticed for the first (airplane object) and last (chair object) row in table \ref{tab:allimages}.
%
\cref{tab:allmetrics} confirms the visual observations from \cref{tab:allimages} with a quantitative comparison with the CD and EMD metrics.

\subsubsection{The Effect of Smart Initialization}
\label{sec:numerical_initialization}

% \begin{wrapfigure}{r}{0.4\textwidth}
\begin{figure}
    \centering
    \includegraphics[width=0.5\linewidth]{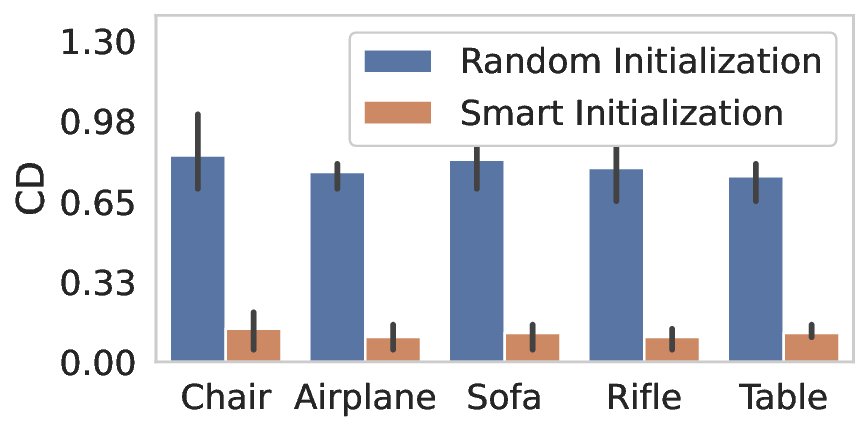}
    \label{fig:smart_init_noise}
    \caption{ Effect of initialization: blue  corresponds to random and orange to smart initialization.}
    \label{fig:smart_init_experiment}
\end{figure}
% \end{wrapfigure}

As the objective functions of 3DMPE  with fixed~\eqref{eq:3Drec_fixed} and varying projections~\eqref{eq:3Drec_varying} are non-convex, initialization is important for the SGD step.
In this section we analyze the effect of smart initialization (see \cref{sec:evaluation}) on the performance of 3DMPE.
We run 3DMPE with smart initialization and compare the results with that of random initialization for the same set of distance matrices.
For the random initialization we sample points uniformly at random from a unit cube.
For each experiment we take 4 instances from the ShapeNet dataset, sample 1024 points (see \cref{sec:dataset}), set the number of viewpoints to 5 with each point visible from 3 viewpoints, and set the maximum number of iterations per experiment to $200$. 
We repeat each experiment 4 times and report the average CD of reconstructions in \cref{fig:smart_init_experiment}, where blue corresponds to smart initialization and orange corresponds to random initialization. 
We observe that in all experiments smart initialization outperforms  random initialization by an order of magnitude. Moreover, the large Chamfer distance (CD) values indicate that random initialization might lead to local rather than global minima. \Cref{tab:smart_init} shows the 3DMPE reconstruction with and without smart initialization, which visually illustrates this point. We remark that smart initialization not only leads to better reconstruction, but  also leads to better running time of the algorithm, as fewer iterations are needed to convergence.

\begin{table*}
\centering
\caption{Example of different 3D objects from ShapeNet dataset with Random and Smart Initialization. The iterations of all experiments are limited to $200$. }
\label{tab:smart_init}
\begin{tabular}{cccc}
Input & Ground Truth & Random Initialization & Smart Initialization \\ \hline
\includegraphics[width=1.5cm,height=1.5cm,keepaspectratio]{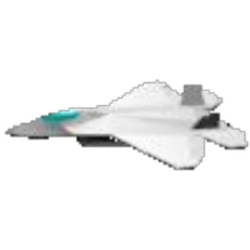}
& \includegraphics[width=1.5cm,height=1.5cm,keepaspectratio]{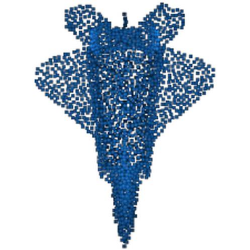} 
& \includegraphics[width=1.5cm,height=1.5cm,keepaspectratio]{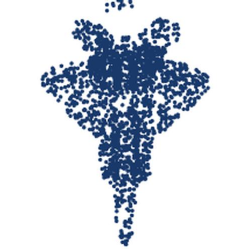} 
& \includegraphics[width=1.5cm,height=1.5cm,keepaspectratio]{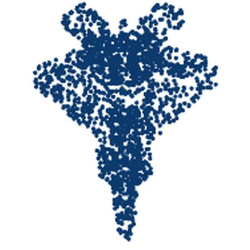} 
 \\

\includegraphics[width=1.5cm,height=1.5cm,keepaspectratio]{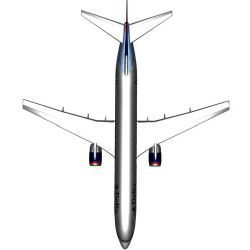}
& \includegraphics[width=1.5cm,height=1.5cm,keepaspectratio]{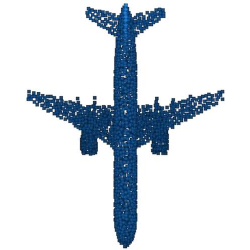} 
& \includegraphics[width=1.5cm,height=1.5cm,keepaspectratio]{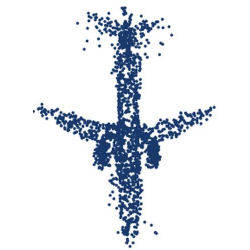} 
& \includegraphics[width=1.5cm,height=1.5cm,keepaspectratio]{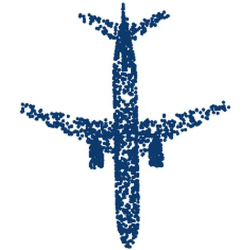} 
 \\

\includegraphics[width=1.5cm,height=1.5cm,keepaspectratio]{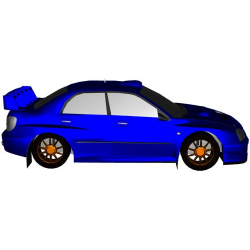}
& \includegraphics[width=1.5cm,height=1.5cm,keepaspectratio]{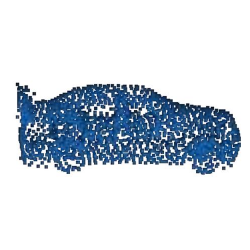} 
& \includegraphics[width=1.5cm,height=1.5cm,keepaspectratio]{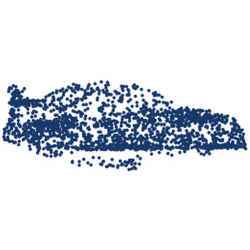} 
& \includegraphics[width=1.5cm,height=1.5cm,keepaspectratio]{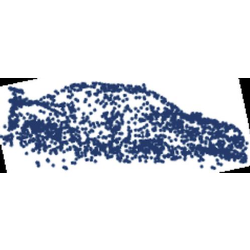} 
 \\

\includegraphics[width=1.5cm,height=1.5cm,keepaspectratio]{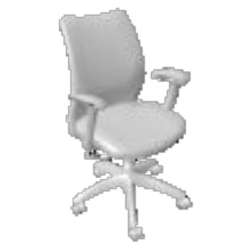}
& \includegraphics[width=1.5cm,height=1.5cm,keepaspectratio]{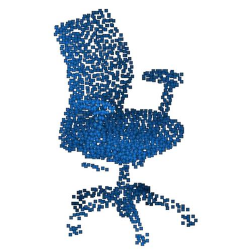} 
& \includegraphics[width=1.5cm,height=1.5cm,keepaspectratio]{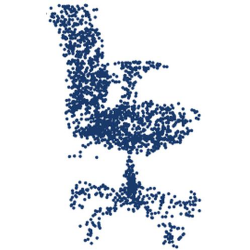} 
& \includegraphics[width=1.5cm,height=1.5cm,keepaspectratio]{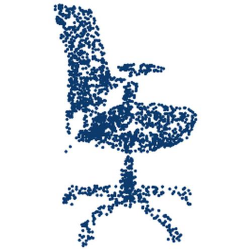} \\

\includegraphics[width=1.5cm,height=1.5cm,keepaspectratio]{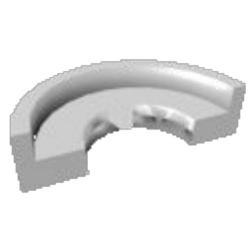}
& \includegraphics[width=1.5cm,height=1.5cm,keepaspectratio]{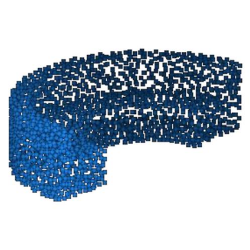} 
& \includegraphics[width=1.5cm,height=1.5cm,keepaspectratio]{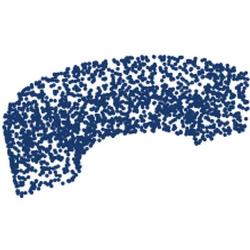} 
& \includegraphics[width=1.5cm,height=1.5cm,keepaspectratio]{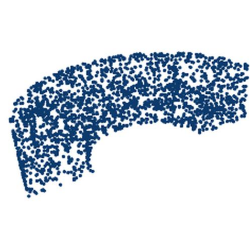} \\

\end{tabular}
\end{table*}

\input{tables/allmetrics.tex}

\subsubsection{Robustness of the Algorithm}
\label{sec:robustness}

\begin{figure}[t]
\centering

\begin{subfigure}[t]{0.49\linewidth}
    \centering
    \includegraphics[width=\linewidth]{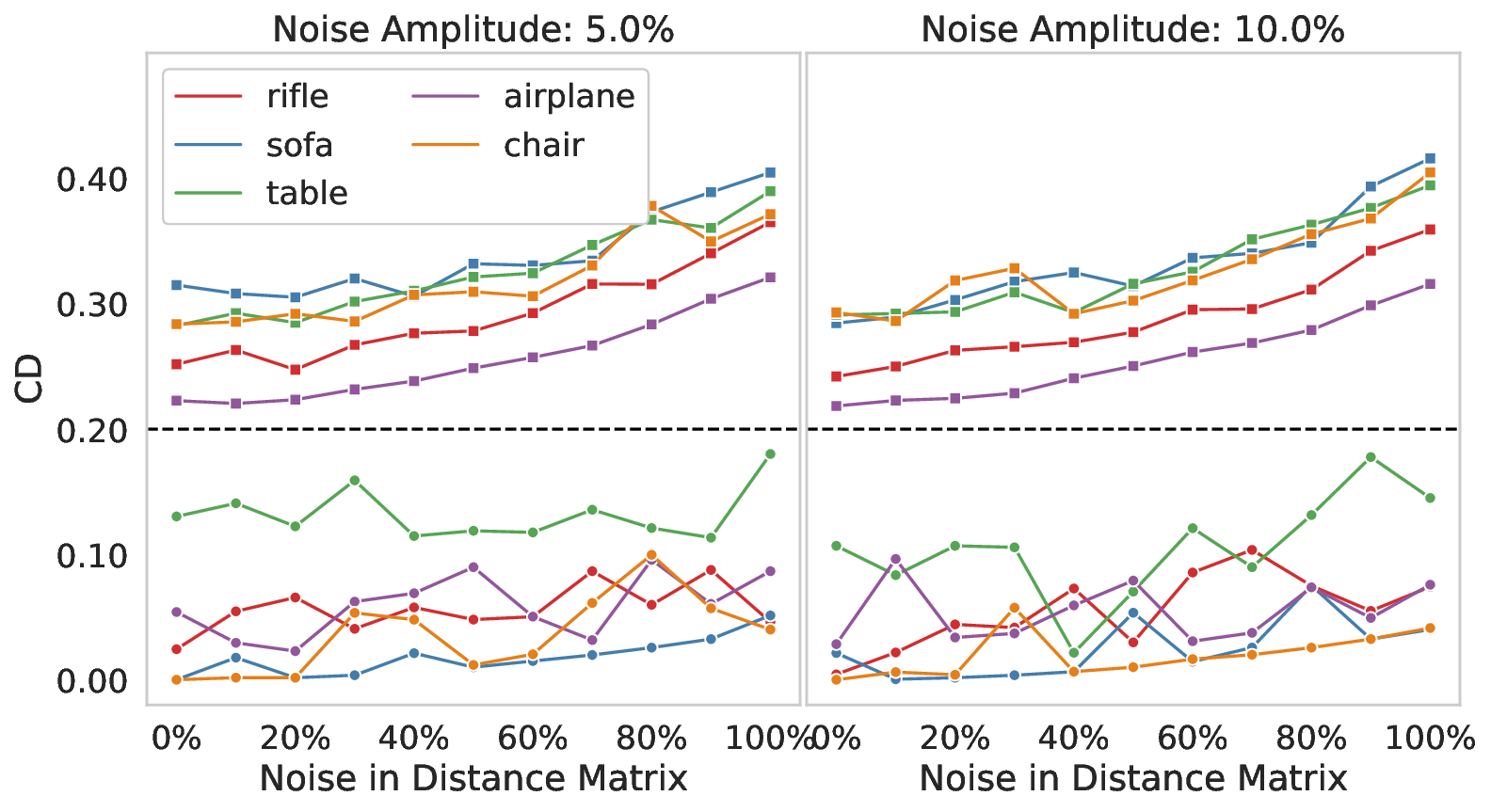}
    \caption{Additive Gaussian noise.}
    \label{fig:gaussian_noise}
\end{subfigure}
\hfill
\begin{subfigure}[t]{0.49\linewidth}
    \centering
    \includegraphics[width=\linewidth]{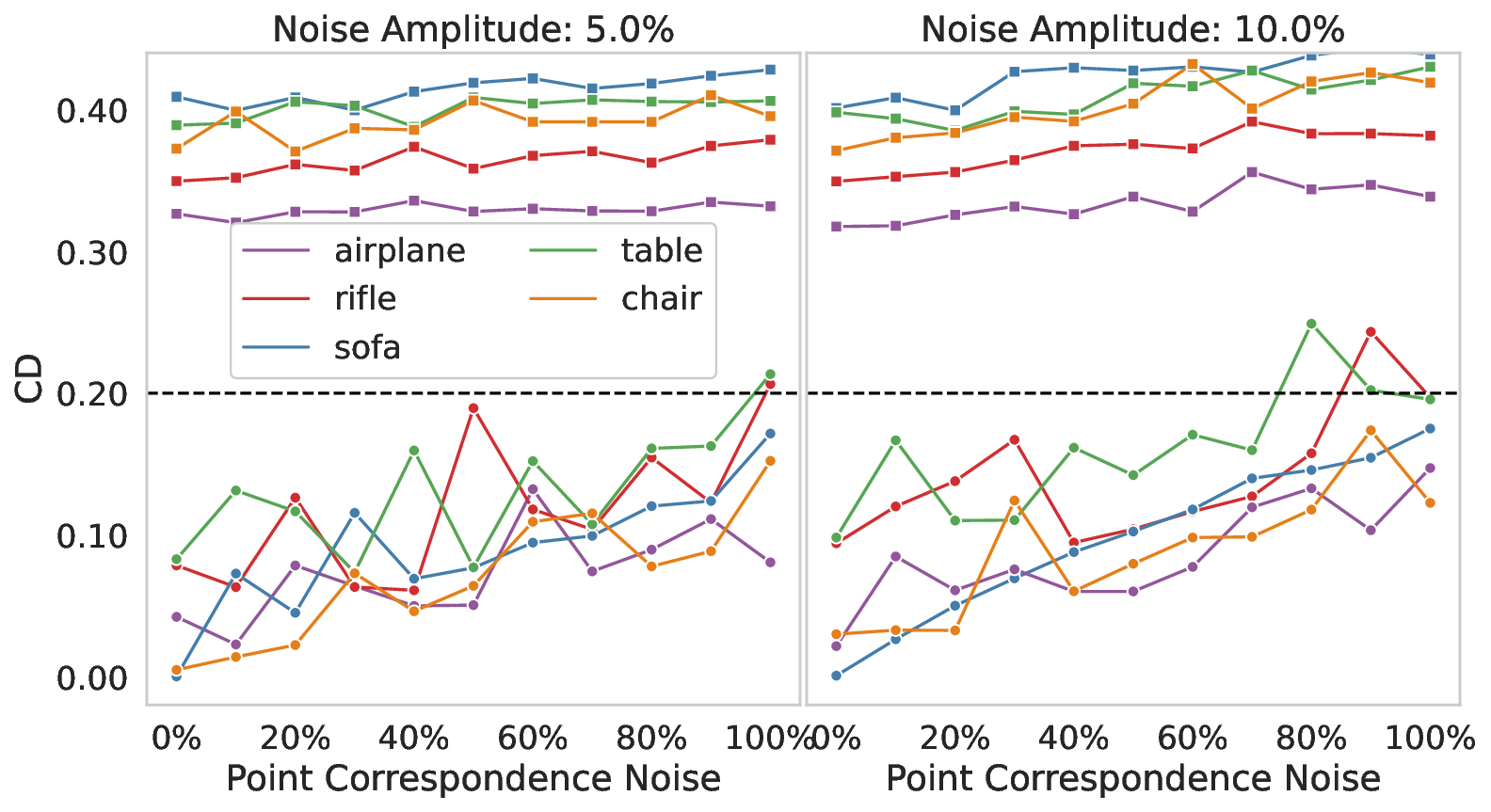}
    \caption{Erroneous correspondences.}
    \label{fig:perm_noise}
\end{subfigure}

\caption{
Noise robustness analysis of 3DMPE (1024 points, 5 viewpoints).
$x$-axis shows the percentage of corrupted points.
Top curves: baseline model (squares); bottom curves: 3DMPE (circles).
}
\label{fig:noise_analysis}
\end{figure}

For all experiments in this section, we take 5 instances from the ShapeNet and sample $n = 1024$ points, with 5 viewpoints, where each point is visible from 3 viewpoints. We have observed that the threshold value for CD under which the reconstruction is acceptable is $\approx 0.2$; see \cref{fig:gaussian_noise} and \cref{fig:perm_noise}.

We analyze the performance of 3DMPE under various noise regimes.
First, we consider additive Gaussian noise to the pairwise distance matrices.
% We take 5 instances from the ShapeNet dataset.
% For each object, we sample $n = 1024$ points according to Section~\ref{sec:dataset} and set the number of viewpoints to be 5 such that each point is visible from 3 viewpoints.
We compute the corresponding pairwise distance matrices and add Gaussian noise to them as follows:
For a given noise amplitude $0 \le p \le 1$ we uniformly pick $nq$ rows in distance matrices and add them Gaussian noise, $\mu = 0,$ $\sigma^2 = pd$, where $d$ is the diameter of the point cloud.
% maximum distance between two points for a given projection. 
We substitute the negative values by $0$
and make the matrices symmetric.
Next, we run 3DMPE with varying projections for these corrupted pairwise distance matrices and report the average reconstruction CD values in \cref{fig:gaussian_noise}. For $p$ in the range $[0.05,0.1]$, increasing the percentage of corrupted points results in a slight increase in CD values while staying in the desired range, i.e., the algorithm successfully recovers the shapes. 
In \cref{fig:gaussian_noise} we also report the values achieved by the baseline model for the experiment with square marked lines and note that the decrease in accuracy for the baseline model is much higher.

% In the first regime, we take 5 differently shaped 3D objects from the ShapeNet dataset and ran 4 experiments to take the average performance. 
% For each experiment 1024 points projected from 5 viewpoints, where each point is visible from 3 viewpoints. Before running 3DMPE with distance matrices, assuming the maximum distance between two points is $d$, we add noise to $n$ points in each distance matrix. 
% This noise is a uniform distribution, where the range is $0$ to $pd$, $0 \le p \le 1$. The left 4 subfigures of Figure~\ref{fig:noise_analysis} shows performance of 3DMPE for different values of $p$ (Noise Amplitude). X-axis denotes $n$, the number of points (rows and columns in distance matrices) that are corrupted by this noise. The lines with square markers are baseline performance, and lines with round markers are 3DMPE performance. 
% \begin{figure}[t]
%     \centering
%     \includegraphics[width=0.6\linewidth]{figures/dist_noise-gaussian.eps}
%     \label{fig:dist_noise}
%     \caption{
%     Additive Gaussian noise analysis of 3DMPE (1024 points, 5 viewpoints). 
%     $x$-axis shows the percentage of corrupted points. Top: baseline model (squares); bottom: 3DMPE (circles).
%     }
%     \label{fig:noise_analysis}
% \end{figure}
Next, we analyze the performance of 3DMPE under erroneous correspondences.
% when some of the correspondences between the points are erroneous.
% For this experiment, we take 5 instances from the ShapeNet dataset.
% For each object we sample $n = 1024$ points according to Section~\ref{sec:dataset} and set the number of viewpoints to be 5 such that each point is visible from 3 viewpoints.
Given $0 \le q \le 1$ (the fraction of points we aim to corrupt), we randomly pick $n q$ points and change the correspondence to a random point, at most $p d$ distance away from it in 3D space, where $d$ is the diameter of the point cloud.
% ($d$ denotes the maximum distance between points in a point cloud).
We compute the pairwise distance matrices and run 3DMPE with varying projections; see \cref{fig:perm_noise}: the $x$-axis corresponds to the value of $q$, the left subfigure corresponds to $p = 0.05$ and the right one corresponds to $p = 0.1$. 
We observe that 3DMPE is robust to inaccuracies in correspondences: even when almost all correspondences are wrong, as long as they are in some reasonable range, 3DMPE recovers the 3D point cloud.

% % {\color{red}
% % We select $n$ points and swap them with a random point that is at most $pd$ distant away in 3D space for a random projection of the points. 

% % Figure \ref{fig:perm_noise} shows a summary of these experiments where each subfigure represents a value of $p$ (Noise Amplitude) and X-axis shows different values of $n$ (Noise Amount). The lines with square markers are baseline performance, and the lines with round markers are 3DMPE performance. 
% % }

% \begin{figure}[t]
%     \centering
%     \includegraphics[width=0.6\linewidth]{figures/perm_noise.eps}
%     \caption{
%     Erroneous correspondences analysis of 3DMPE (1024 points, 5 viewpoints). 
%     $x$-axis shows the percentage of corrupted points.
%     Top: baseline model (squares); bottom: 3DMPE (circles).
%     }
%     \label{fig:perm_noise}
% \end{figure}

% We select different rotational angles ($\alpha$,$\beta$,$\gamma$) from a uniform distribution over a number of experiments. To analyze the impact of view angles of input, we use angles from uniform distribution of range $[0,2\pi]$ (viewpoints can be from any side), and slowly narrow down the range to $[0,0]$ (viewpoints can only be from one location).

\subsubsection{Scalability of the Algorithm}
\label{sec:scalability}

\begin{figure}[t]
     \centering
     \begin{subfigure}{0.24\linewidth}
         \centering
         \includegraphics[width=\textwidth]{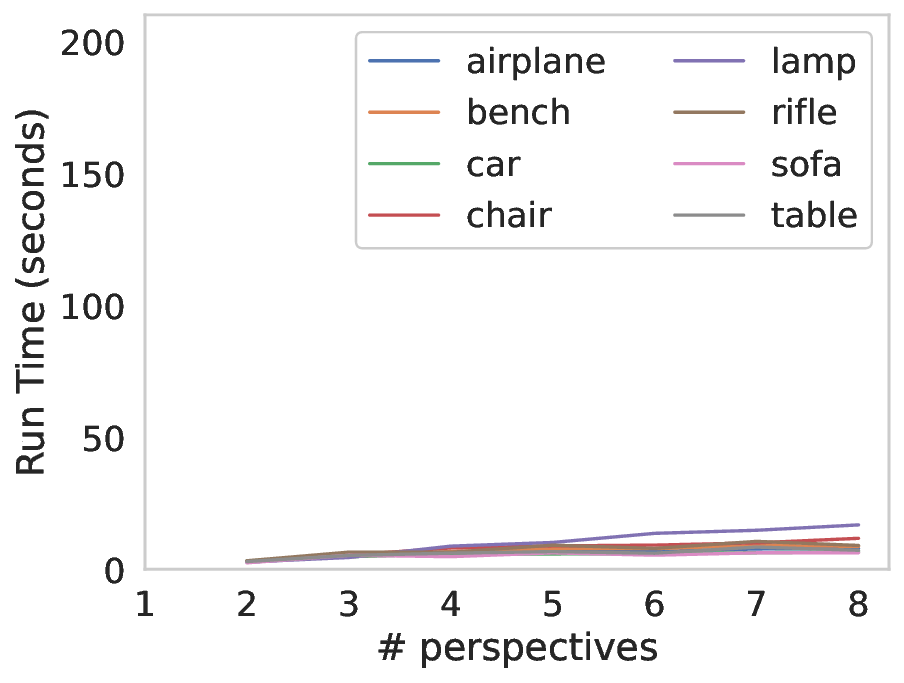}
         \caption{Fixed Proj.}
         \label{fig:viewpoints_fixed_Run_Time}
     \end{subfigure}
     % \hfill
     \begin{subfigure}{0.24\linewidth}
         \centering
         \includegraphics[width=\textwidth]{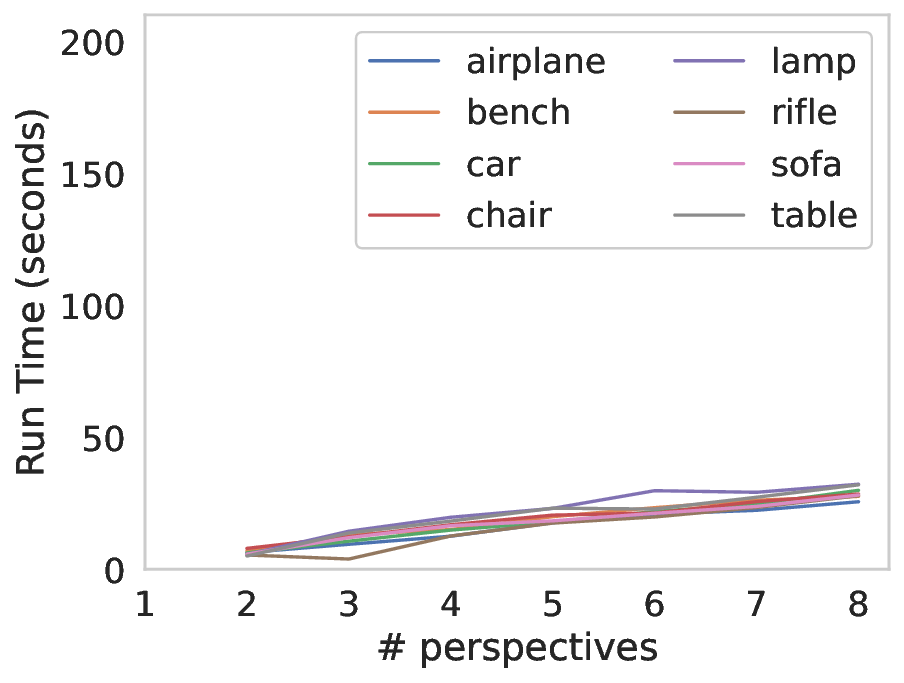}
         \caption{Variable Proj.}
         \label{fig:viewpoints_var_Run_Time}
     \end{subfigure}
     \begin{subfigure}{0.24\linewidth}
         \centering
         \includegraphics[width=\textwidth]{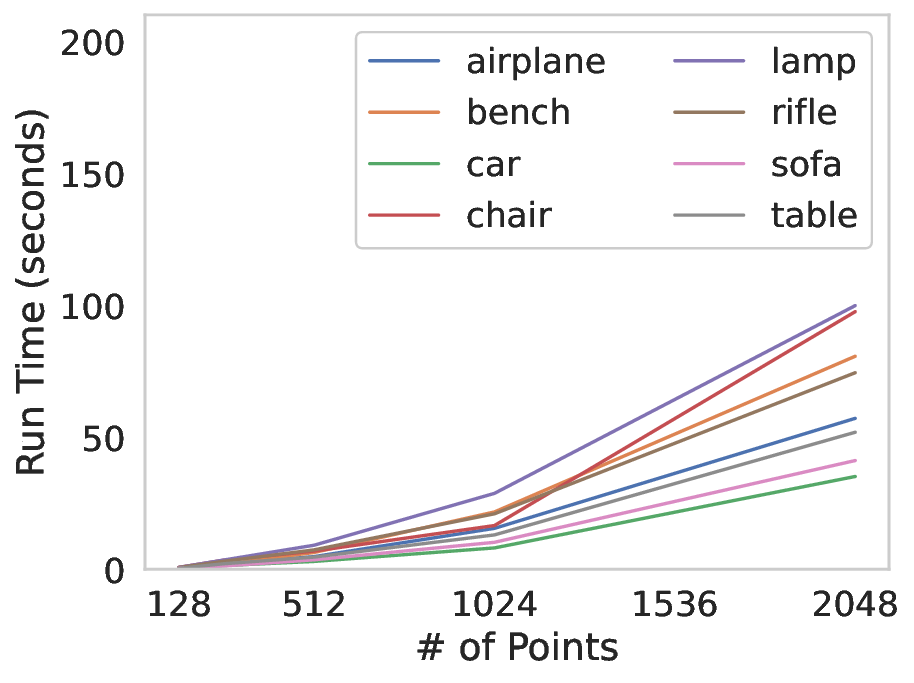}
         \caption{Fixed Proj.}
         \label{fig:points_fixed_Run_Time}
     \end{subfigure}
     % \hfill
     \begin{subfigure}{0.24\linewidth}
         \centering
         \includegraphics[width=\textwidth]{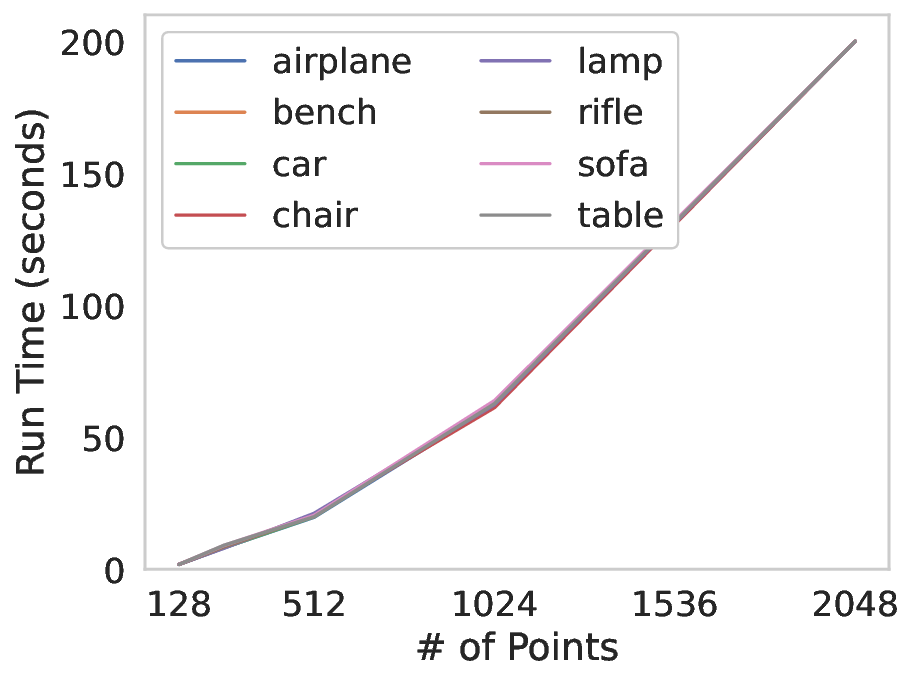}
         \caption{Variable Proj.}
         \label{fig:points_var_Run_Time}
     \end{subfigure}
    \caption{Runtime (seconds) of 3DMPE with the ShapeNet dataset: (a-b) show the impact of changing the number of perspectives (using 512 points) and (c-d) show the impact of changing the number of points (with 4 projections).}
    \label{fig:viewpoints_runtime}
\end{figure}

We next discuss the performance of 3DMPE (in both the fixed and variable projections setting) with different size point clouds, with different  number of viewpoints, with respect to accuracy and runtime.
%To test the scalability of both the runtime and accuracy of we track the runtime and all metrics of our algorithm for different experimental setups. 
\cref{fig:viewpoints_fixed_Run_Time} and \cref{fig:viewpoints_var_Run_Time} demonstrate the effect on runtime with more viewpoints for fixed and variable projections, respectively. 

In the fixed embedding setting, the effect of increasing the projections (from 2 to 8) is negligible. In the variable embedding setting there is an increase in runtime that is linear in the number of viewpoints (as in this setting the optimization needs to not only place the points but compute the viewpoints). \cref{fig:points_fixed_Run_Time} and \cref{fig:points_var_Run_Time} demonstrate the change in runtime of 3DMPE with fixed and varying projections as the number of points in the input point cloud increases.
Clearly the algorithm slows down for larger instances to around 50 seconds for instances with 2048 points in the fixed setting and around 200 seconds in the variable setting.

%linear increase in runtime. For 3DMPE with varying projections in Figure~\ref{fig:points_var_Run_Time} the increase is much larger (but still almost linear) than the one for 3DMPE with fixed projections in Figure~\ref{fig:points_fixed_Run_Time}. We remark that 
We note that 3DMPE with variable projections solves a more difficult optimization task and requires a greater number of iterations than the fixed projection setting. In this proof-of-concept implementation, we did not invest effort in optimizing 3DMPE for speed.

\begin{figure}
     \centering
     \begin{subfigure}{0.24\linewidth}
         \centering
         \includegraphics[width=\textwidth]{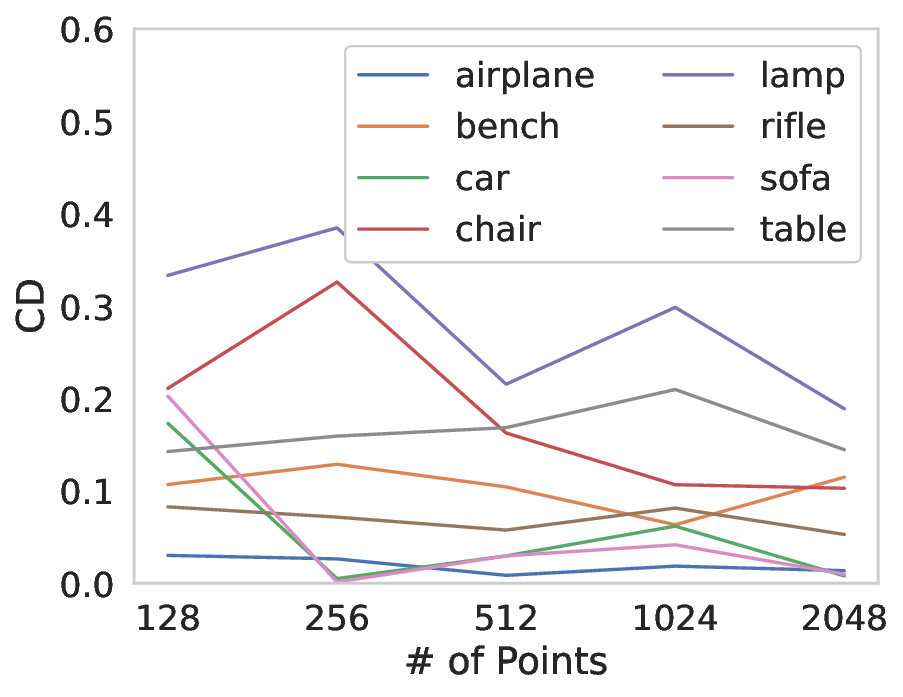}
         \caption{Fixed Proj.}
         \label{fig:points_fixed_Chamfer}
     \end{subfigure}
     % \hfill
     \begin{subfigure}{0.24\linewidth}
         \centering
         \includegraphics[width=\textwidth]{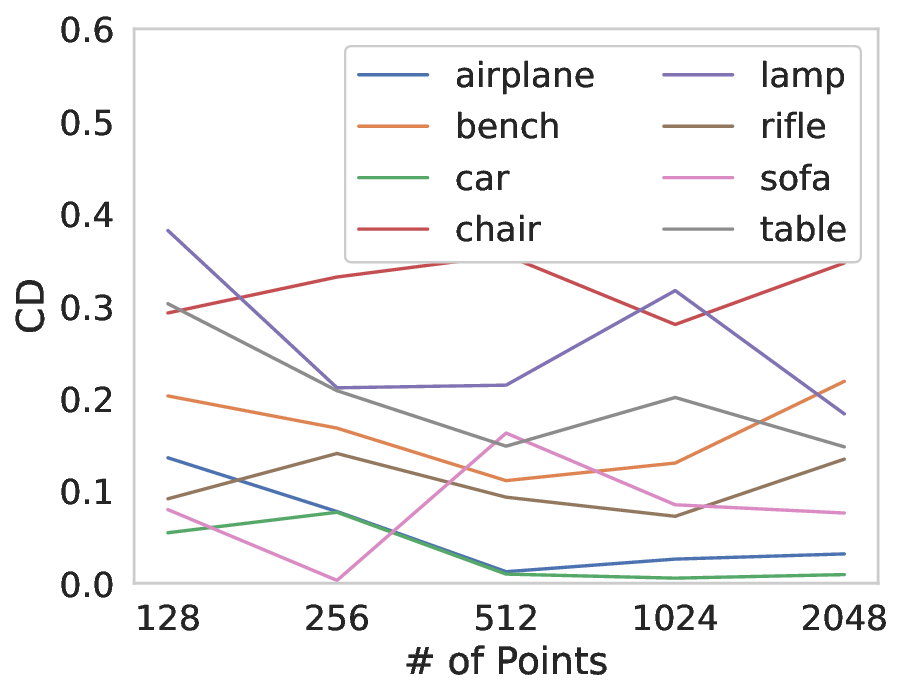}
         \caption{Variable Proj.}
         \label{fig:points_var_Chamfer}
     \end{subfigure}
     \begin{subfigure}{0.24\linewidth}
         \centering
         \includegraphics[width=\textwidth]{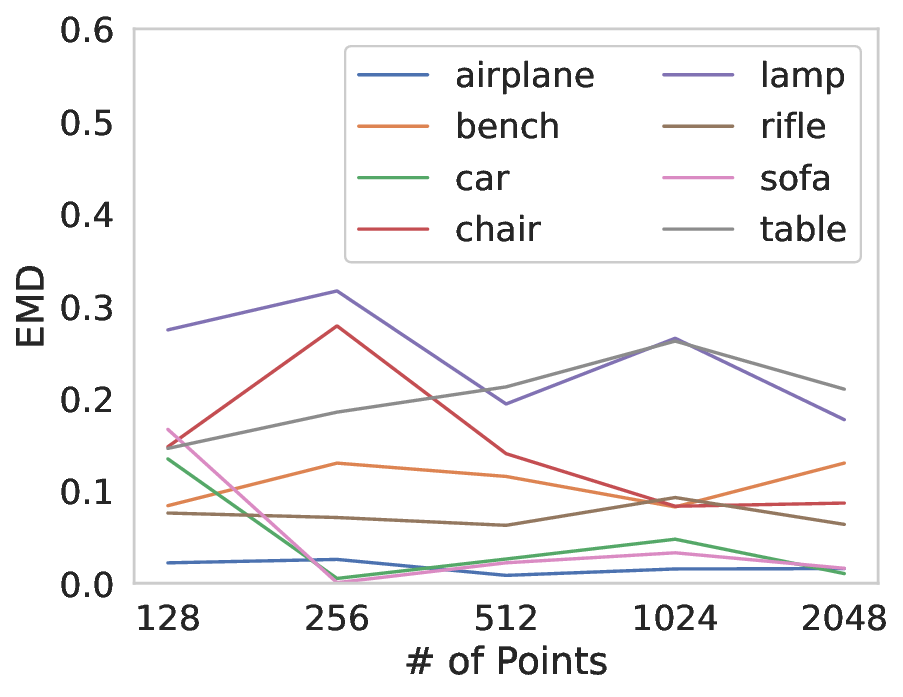}
         \caption{Fixed Proj.}
         \label{fig:points_fixed_EMD}
     \end{subfigure}
     % \hfill
     \begin{subfigure}{0.24\linewidth}
         \centering
         \includegraphics[width=\textwidth]{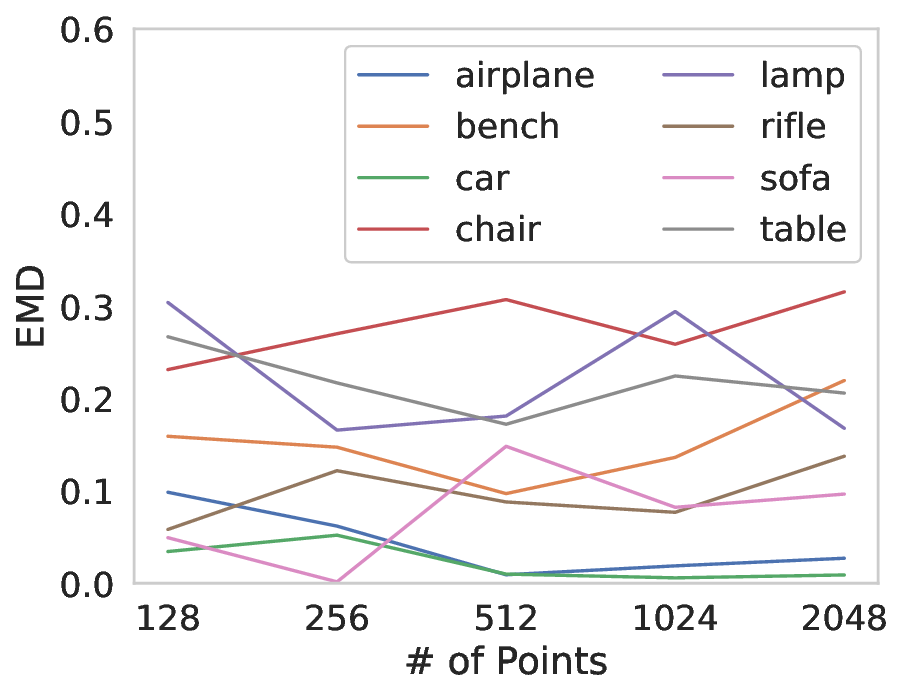}
         \caption{Variable Proj.}
         \label{fig:points_var_EMD}
     \end{subfigure}
    % \caption{EMD and CD values with varying number of points on ShapeNet data. Experiments are done with 4 viewpoints and each point visible from at least 3 viewpoints. }
    \caption{EMD and CD values with varying number of points on ShapeNet (4 viewpoints, each point visible from 3+ viewpoints). }
    \label{fig:points_metrics}
\end{figure}

\begin{figure}[t]
     \centering
     \begin{subfigure}{0.24\linewidth}
         \centering
         \includegraphics[width=\textwidth]{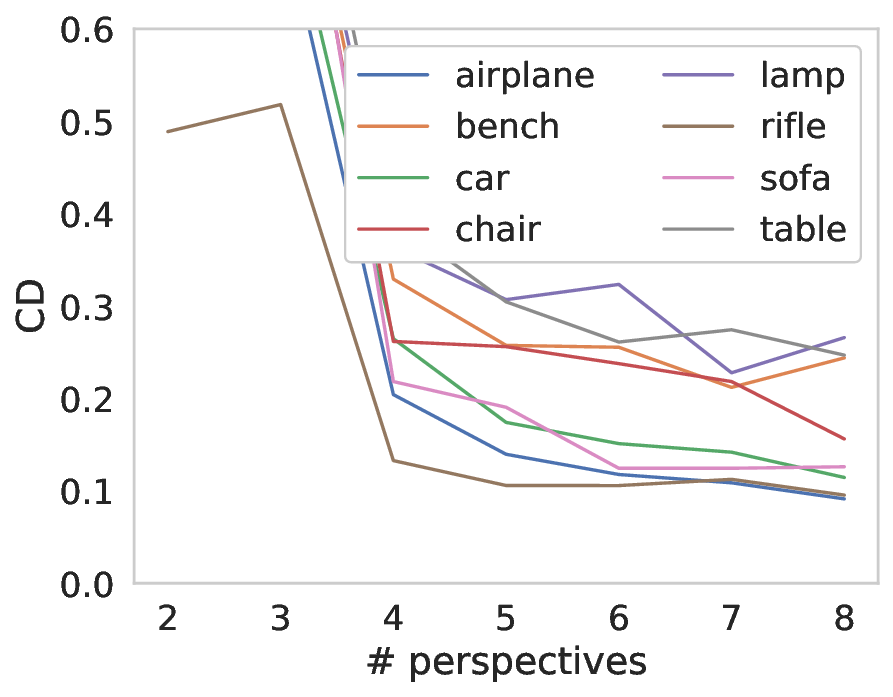}
         \caption{Fixed Proj.}
         \label{fig:viewpoints_fixed_Chamfer}
     \end{subfigure}
     % \hfill
     \begin{subfigure}{0.24\linewidth}
         \centering
         \includegraphics[width=\textwidth]{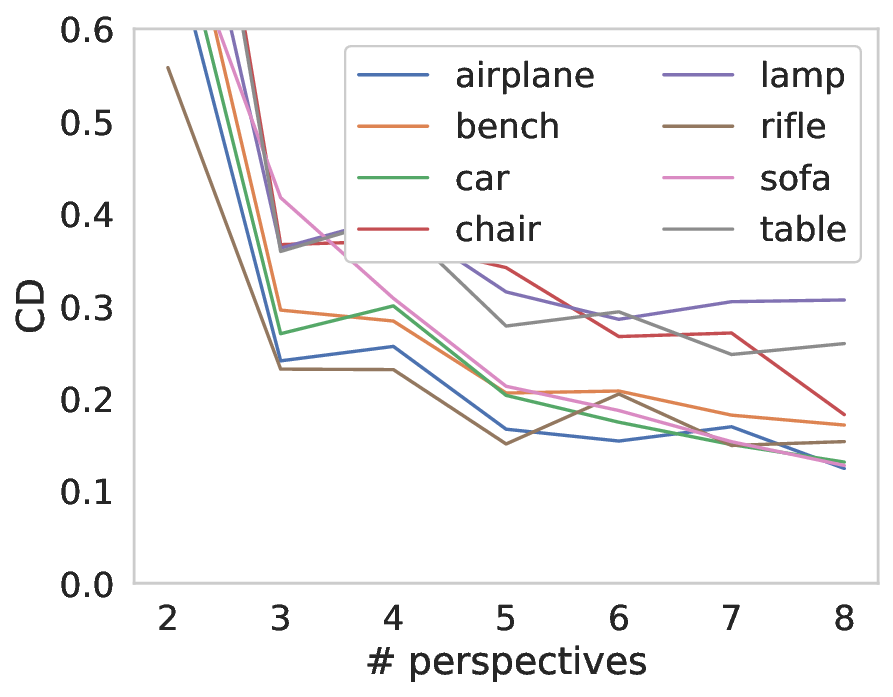}
         \caption{Variable Proj.}
         \label{fig:viewpoints_var_Chamfer}
     \end{subfigure}
     \begin{subfigure}{0.24\linewidth}
         \centering
         \includegraphics[width=\textwidth]{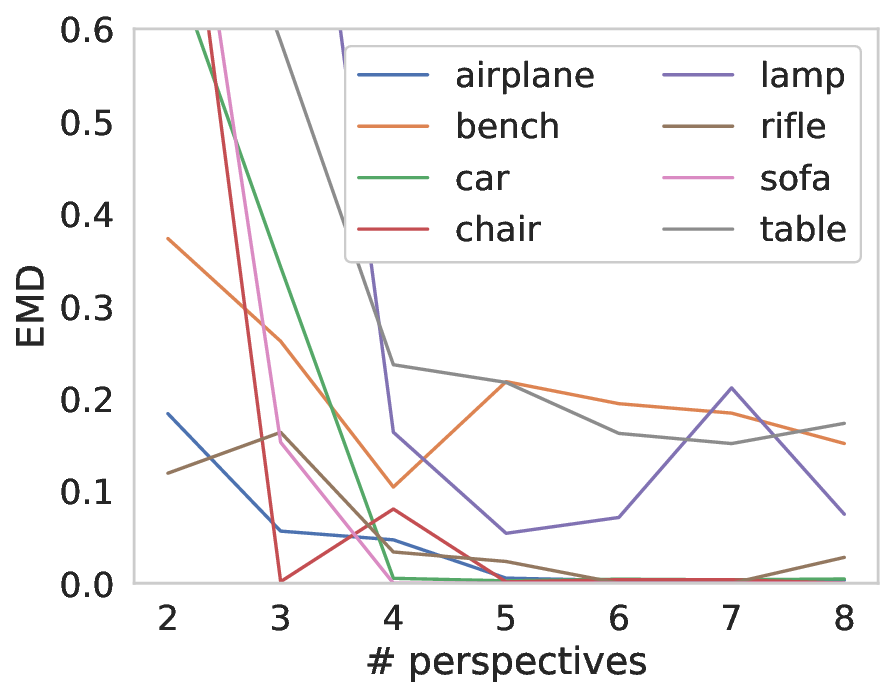}
         \caption{Fixed Proj.}
         \label{fig:viewpoints_fixed_EMD}
     \end{subfigure}
     % \hfill
     \begin{subfigure}{0.24\linewidth}
         \centering
         \includegraphics[width=\textwidth]{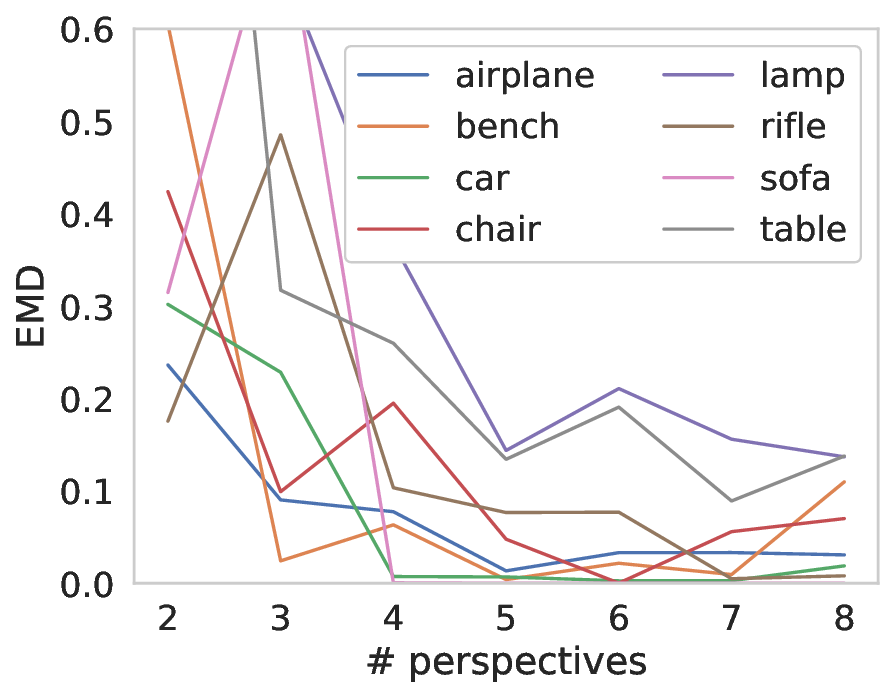}
         \caption{Variable Proj.}
         \label{fig:viewpoints_var_EMD}
     \end{subfigure}
    \caption{3DMPE metrics on ShapeNet for varying viewpoints (512 points; all visible for 2–3 views, otherwise each point appears in one fewer view).}
    % \caption{Metric analysis of 3DMPE on ShapeNet for varying number of viewpoints (512 points, each point is visible 1 less than the number of viewpoints, for 2 and 3 all points are visible).}
    \label{fig:viewpoints_metrics}
\end{figure}

Although the runtime of 3DMPE is affected by  increasing the number of points and viewpoints, we see no significant effect of these parameters on the accuracy of the 3DMPE for both fixed and variable settings; see \cref{fig:points_metrics}.
%demonstrates the CD, EMD and ROA metrics for fixed and variable projections. 
Further, the CD, EMD and ROA values for both 3DMPE variants are comparable, although 3DMPE with varying projections is a more complex problem to solve.
%(the projections/the viewpoints are unknown).
\cref{fig:viewpoints_metrics} demonstrates the effect of the increase of the number of perspectives on CD and EMD when the number of points is fixed. ROA metrics for the same experiments are in the appendix.
As expected,  reconstruction quality is poor with fewer than 3 perspectives. Typically 4 or 5 projections from different angles suffice for high-quality reconstruction.

%The raytracing algorithm behaves like a real-world camera. Thus, depending on the 3D object, the number of hidden points generated by the raytracing algorithm will vary. 
\begin{figure}[t]
     \centering
     \begin{subfigure}{0.24\linewidth}
         \centering
         \includegraphics[width=\textwidth]{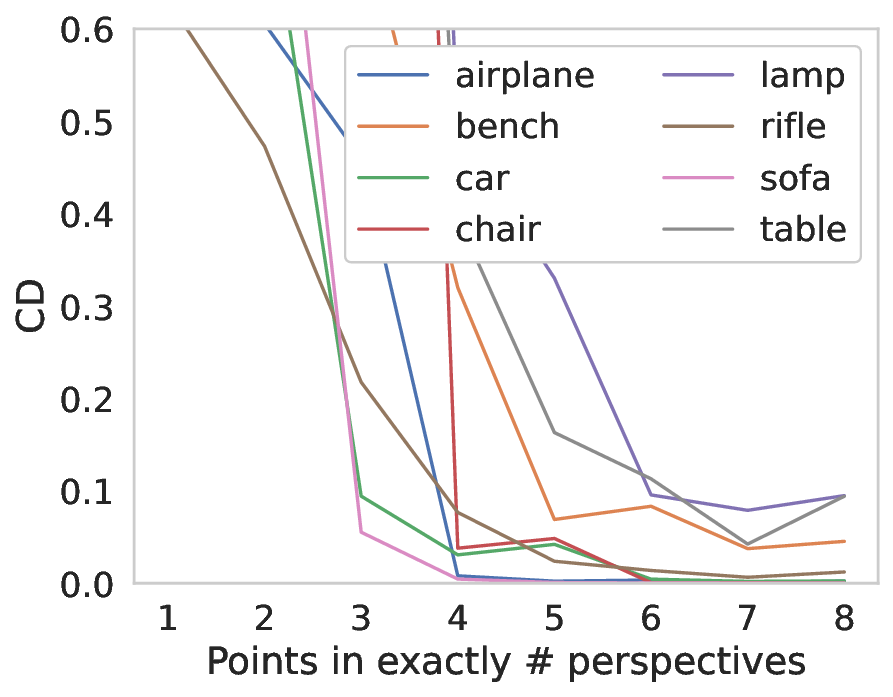}
         \caption{Fixed Proj.}
         \label{fig:visible_fixed_Chamfer}
     \end{subfigure}
     % \hfill
     \begin{subfigure}{0.24\linewidth}
         \centering
         \includegraphics[width=\textwidth]{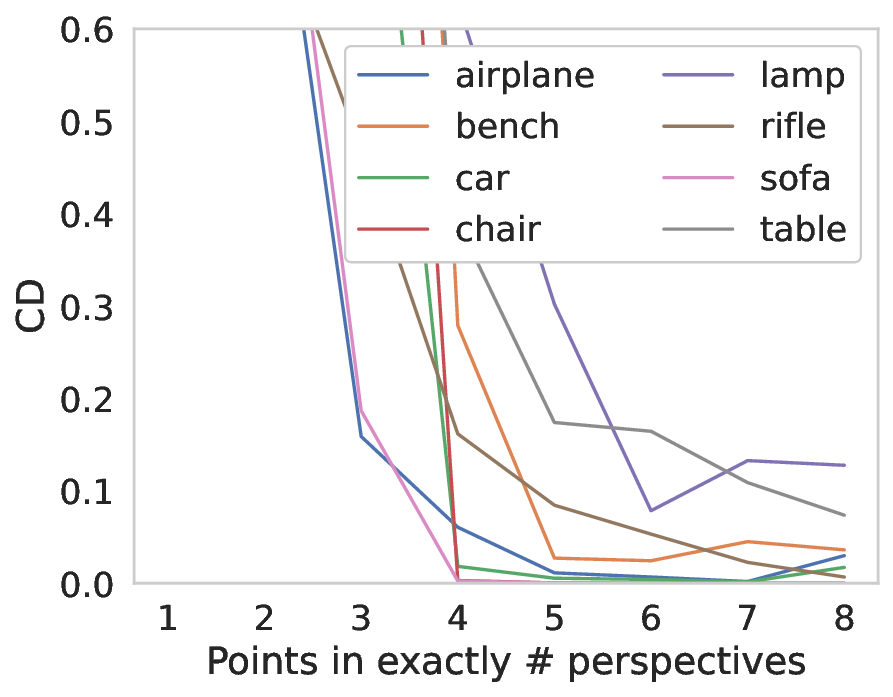}
         \caption{Variable Proj.}
         \label{fig:visible_var_Chamfer}
     \end{subfigure}
     \begin{subfigure}{0.24\linewidth}
         \centering
         \includegraphics[width=\textwidth]{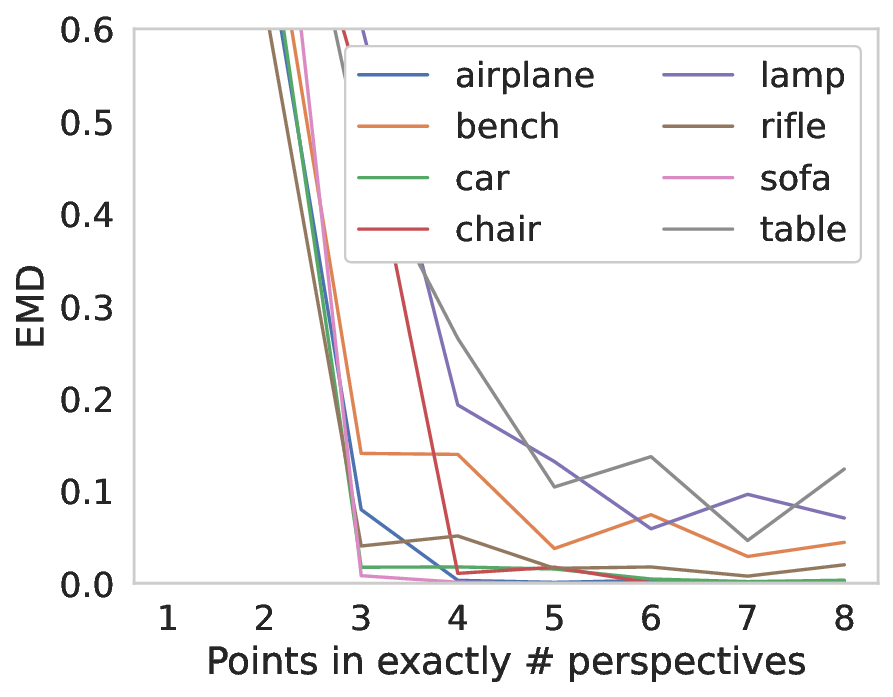}
         \caption{Fixed Proj.}
         \label{fig:visible_fixed_EMD}
     \end{subfigure}
     % \hfill
     \begin{subfigure}{0.24\linewidth}
         \centering
         \includegraphics[width=\textwidth]{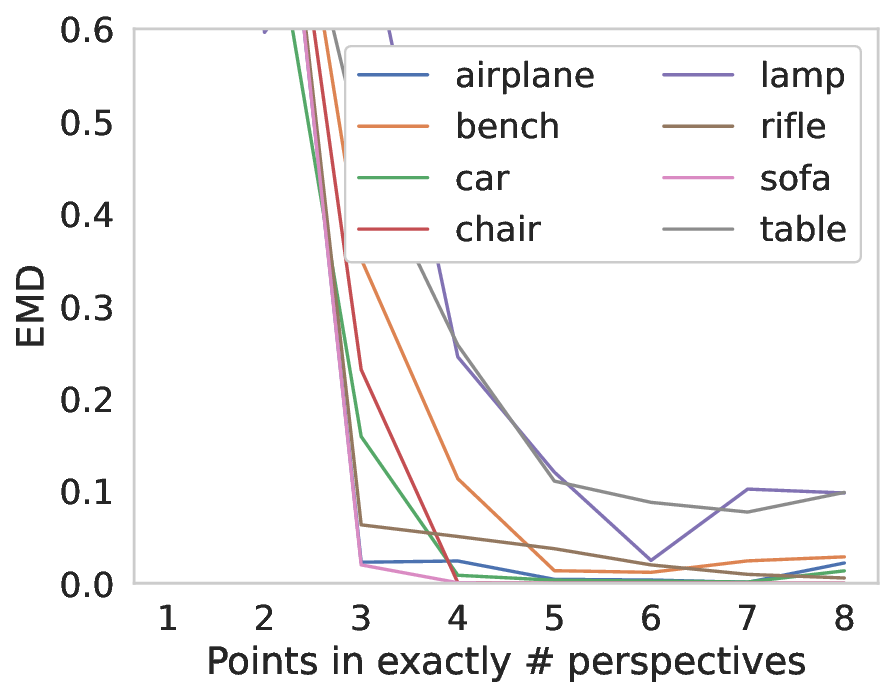}
         \caption{Variable Proj.}
         \label{fig:visible_var_EMD}
     \end{subfigure}
    \caption{Metric analysis of 3DMPE on ShapeNet dataset for points visible in a number of viewpoints. All experiments are done for 512 points with 8 viewpoints.}
    \label{fig:visible_metrics}
\end{figure}

\begin{figure}
     \centering
    \includegraphics[width=0.5\linewidth]{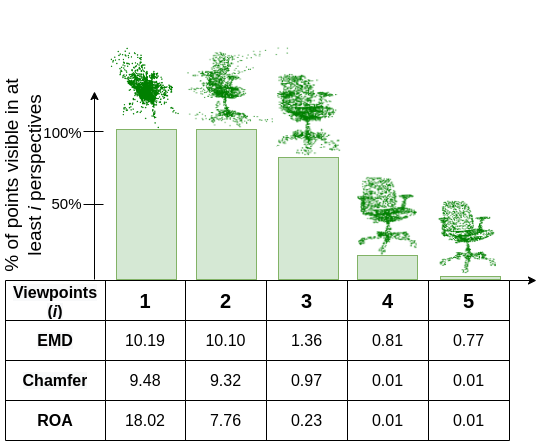}
    \caption{Points visible in at least  $1 \dots 5$ viewpoints for ray tracing with five equidistant views. 3DMPE reconstructions for the corresponding points are shown above each bar; reconstruction metrics are reported in the table.}
    \label{fig:raytrace_visible}
\end{figure}

We analyze the sensitivity of 3DMPE to the percentage of hidden points as follows:
fix a parameter $\vartheta$, which represents the number of points that we want to be visible in each projection, create the 2D projections from the given point cloud, and from each projection  remove $N-\vartheta$ point at random.
In \cref{fig:visible_metrics} we report the CD and EMD (ROA data in the appendix) for the point clouds generated by 3DMPE (with fixed and varying projections) for 512 points, ensuring that each point is visible in $1$ fewer than the total number of viewpoints (except 2 and 3 where all points are visible).
The experiments show that 3DMPE reconstructs the 3D point cloud properly if each point is visible from at least 3 projections. 
We remark, that this is feasible in real scenarios of photography. 
As demonstrated in \cref{fig:raytrace_visible}, even only for 5 projections, most of the points are present in at least 3 viewpoints. Thus, if there are more projections, the points will be visible from at least 3 viewpoints.

% \begin{wrapfigure}{r}{0.4\textwidth}
% \end{wrapfigure}

\subsubsection{Analysis of Angle Constraints.}

\begin{figure}
    \centering
    \includegraphics[width=0.5\linewidth]{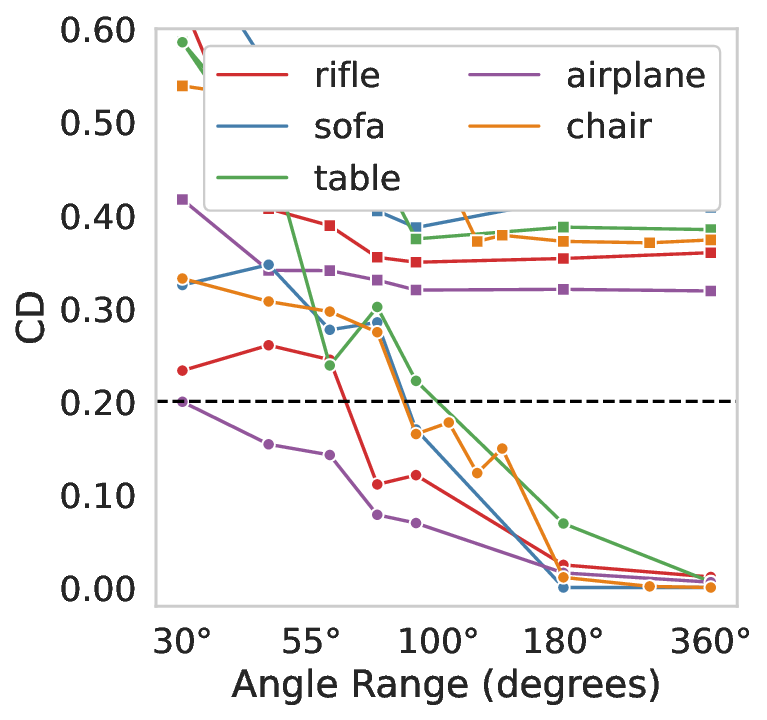}
    \caption{Performance of 3DMPE with different constraints on viewpoint positions. The $x$-axis (logarithmic scale) denotes the maximum angle distance between each viewpoint from the center of point cloud.}
    \label{fig:angle_noise}
\end{figure}

Here we briefly discuss the impact of angle constraints on the 3DMPE algorithm. We choose four 3D models and conduct an experiment where everything else is the same, except for the viewpoint angle constraints. For each experiments, point are visible in exactly 3 projections out of 5. 
Instead of using $\theta_s = 0$ and $\theta_e = 2\pi$ (so that $\theta_r=360^{\circ}$), we start from $[0, 2\pi]$, (viewpoints can be from any side), and slowly narrow down the range to $[0,0]$ (viewpoints can only be from one location). 
\cref{fig:angle_noise} shows the performance of 3DMPE for the four 3D models of the ShapeNet dataset, where the $x$-axis denotes $\theta_r$. Lines with square markers show the baseline, and lines with round markers show 3DMPE. As discussed in \cref{sec:robustness} the dashed line represents a margin above which the 3D reconstruction becomes unrecognizable. 
The figure indicates that for $\theta_r \ge 90^{\circ}$, 3DMPE can reconstruct the whole object.
%even with hidden points. Note that the x-axis is in logarithmic scale, and accuracy of 3DMPE is logarithmically decreasing with respect to angle constraint.

\subsection{Experiments on Pix3D}

In addition to experiments with examples from the ShapeNet dataset we also evaluated 3DMPE on examples from the Pix3D dataset~\cite{pix3d}; see \cref{tab:pix3dallimages}. 
% with the same 3D objects used in~\cite{mandikal20183D}. 
In all of these experiments we use 3DMPE with 2048 points and 4 or 5 viewpoints. We remark, that according to our experiments, the quality of reconstruction drastically increases as the number of viewpoints increases.
Note, that the exact orientations of the given examples in referred works are unknown. 
This is possibly due to the fact that these solutions of the 3D point cloud reconstruction problem are based on the relative position of the points rather than the absolute positions of those points. 
Thus, for both ShapeNet and Pix3D datasets, we visually aligned our baseline and 3DMPE reconstructions as close to the given images as possible.
Finally, according to \cref{tab:pix3dallimages},  3D-LMNet alongside both 3DMPE models performed the best. 
There are some instances where 3DMPE outperformed 3D-LMNet (e.g., the last table object in Table~\ref{tab:pix3dallimages}), and some cases where 3DMPE with variable projection performed poorly on certain portions of 3D models (e.g. the 1st sofa of Table~\ref{tab:pix3dallimages}).
% This can be overcome for 3DMPE with a higher number of viewpoints as input to the algorithm.
This issue is resolved by increasing the number of viewpoints.

\input{tables/pix3dallimages.tex}
\input{tables/allmetricspix3d.tex}

In Table~\ref{tab:allmetricspix3d} we present the quantitative comparison between 3DMPE, our proposed Baseline method, and the state-of-art 3D reconstruction models with both metrics CD and EMD. Similar to \cref{tab:allmetrics}, we note that 3DMPE with fixed projection performed the best.

\begin{figure*}
     \centering
     \begin{subfigure}{0.24\linewidth}
         \centering
         \includegraphics[width=\textwidth]{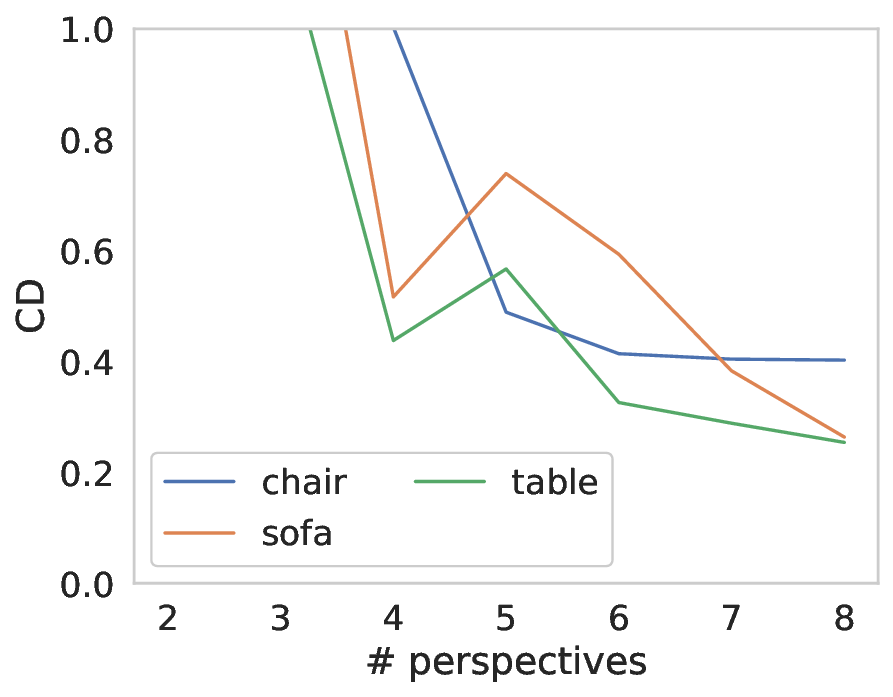}
         \caption{Fixed Proj.}
         \label{fig:pix3d_viewpoints_fixed_Chamfer}
     \end{subfigure}
     \begin{subfigure}{0.24\linewidth}
         \centering
         \includegraphics[width=\textwidth]{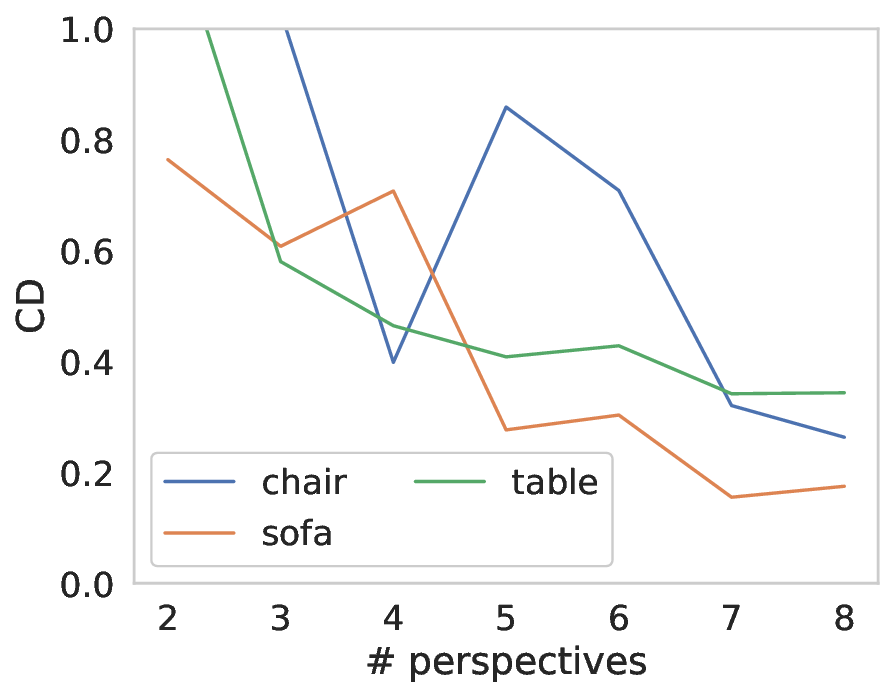}
         \caption{Variable Proj.}
         \label{fig:pix3d_viewpoints_var_Chamfer}
     \end{subfigure}
     \begin{subfigure}{0.24\linewidth}
         \centering
         \includegraphics[width=\textwidth]{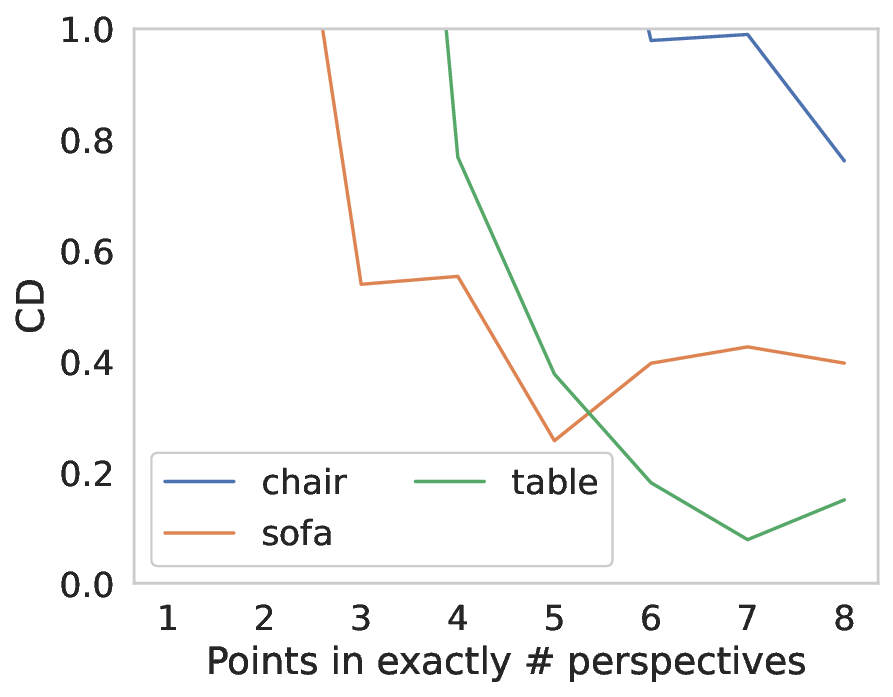}
         \caption{Fixed Proj.}
         \label{fig:pix3d_visible_fixed_Chamfer}
     \end{subfigure}
     \begin{subfigure}{0.24\linewidth}
         \centering
         \includegraphics[width=\textwidth]{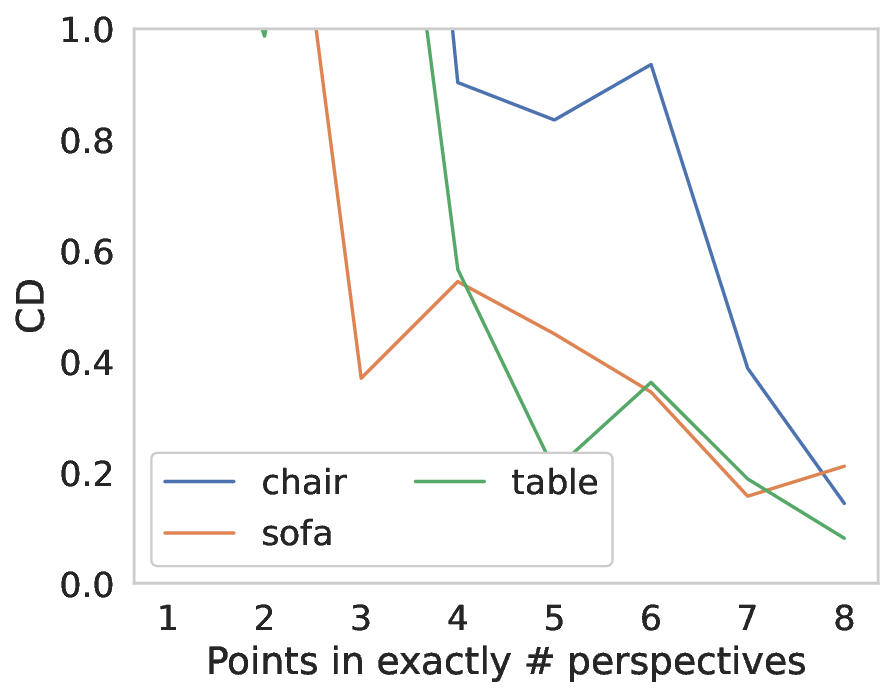}
         \caption{Variable Proj.}
         \label{fig:pix3d_visible_var_Chamfer}
     \end{subfigure}

     \begin{subfigure}{0.24\linewidth}
         \centering
         \includegraphics[width=\textwidth]{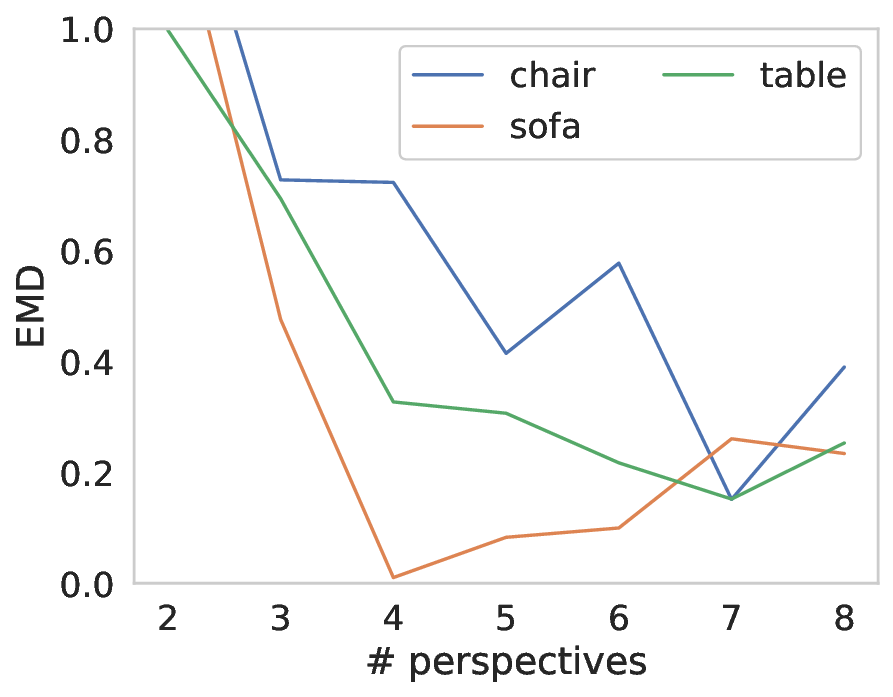}
         \caption{Fixed Proj.}
         \label{fig:pix3d_viewpoints_fixed_EMD}
     \end{subfigure}
     \begin{subfigure}{0.24\linewidth}
         \centering
         \includegraphics[width=\textwidth]{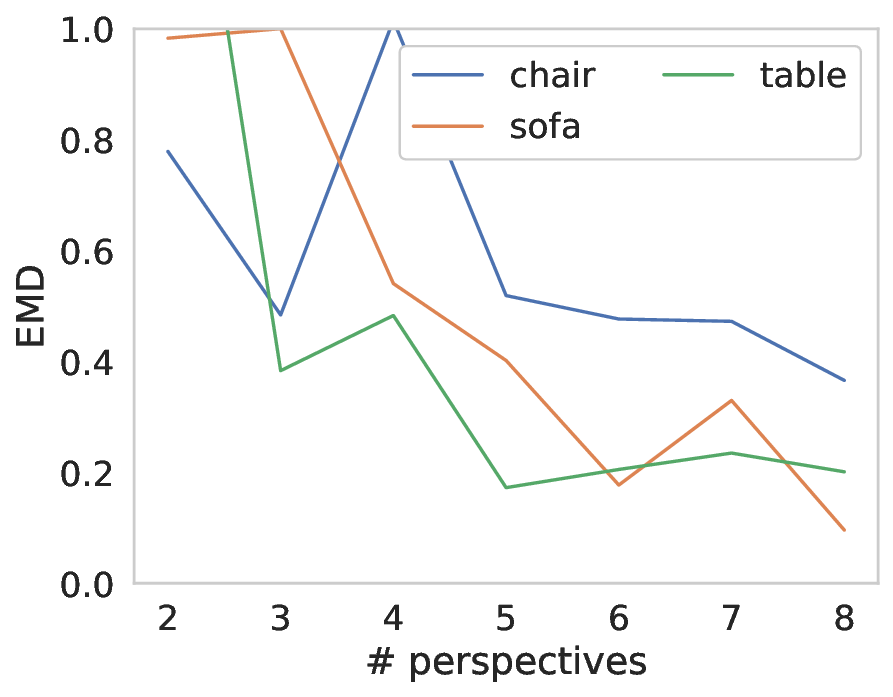}
         \caption{Variable Proj.}
         \label{fig:pix3d_viewpoints_var_EMD}
     \end{subfigure}
     \begin{subfigure}{0.24\linewidth}
         \centering
         \includegraphics[width=\textwidth]{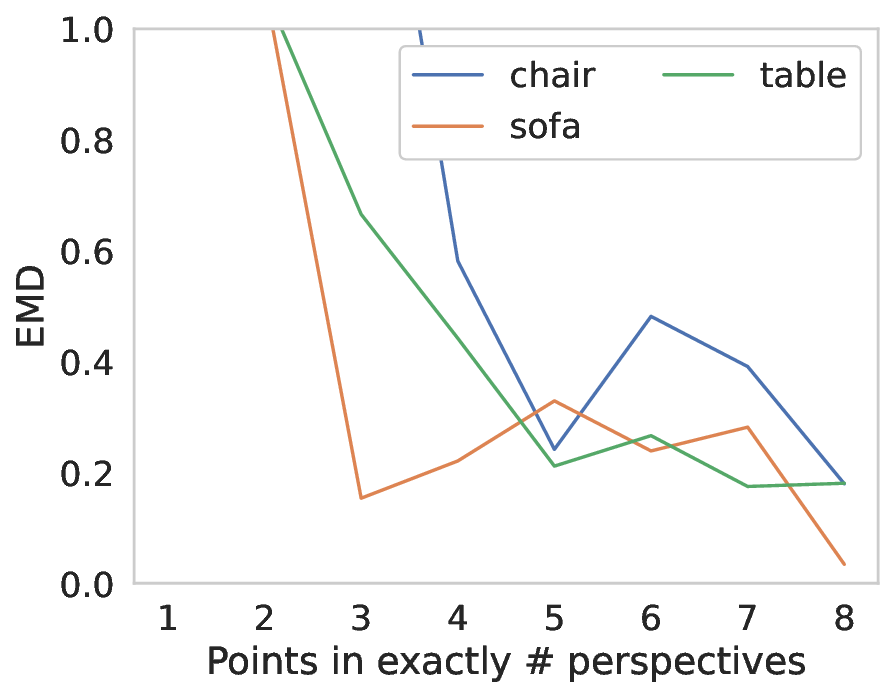}
         \caption{Fixed Proj.}
         \label{fig:pix3d_visible_fixed_EMD}
     \end{subfigure}
     \begin{subfigure}{0.24\linewidth}
         \centering
         \includegraphics[width=\textwidth]{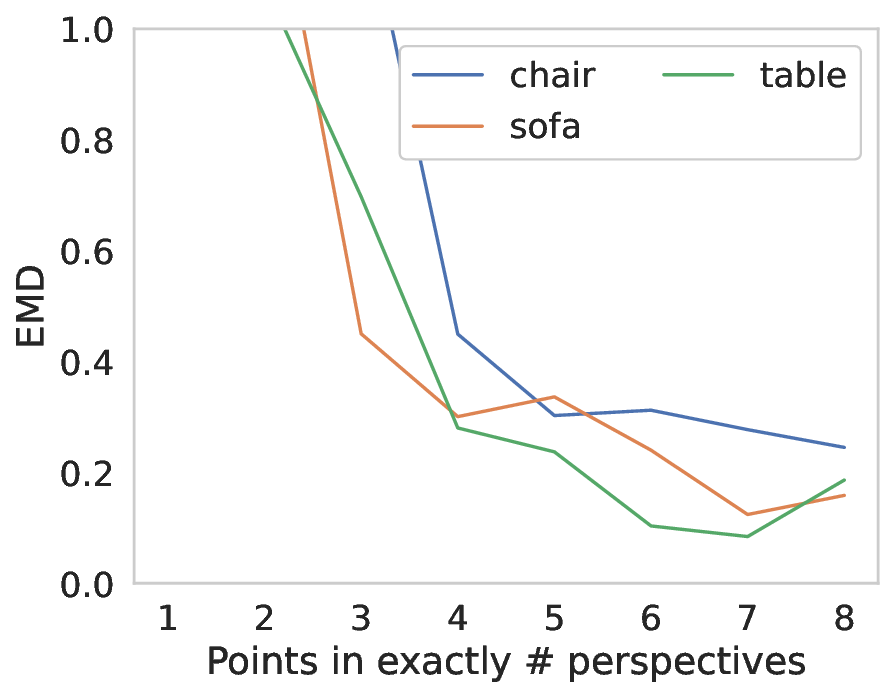}
         \caption{Variable Proj.}
         \label{fig:pix3d_visible_var_EMD}
     \end{subfigure}

     \begin{subfigure}{0.24\linewidth}
         \centering
         \includegraphics[width=\textwidth]{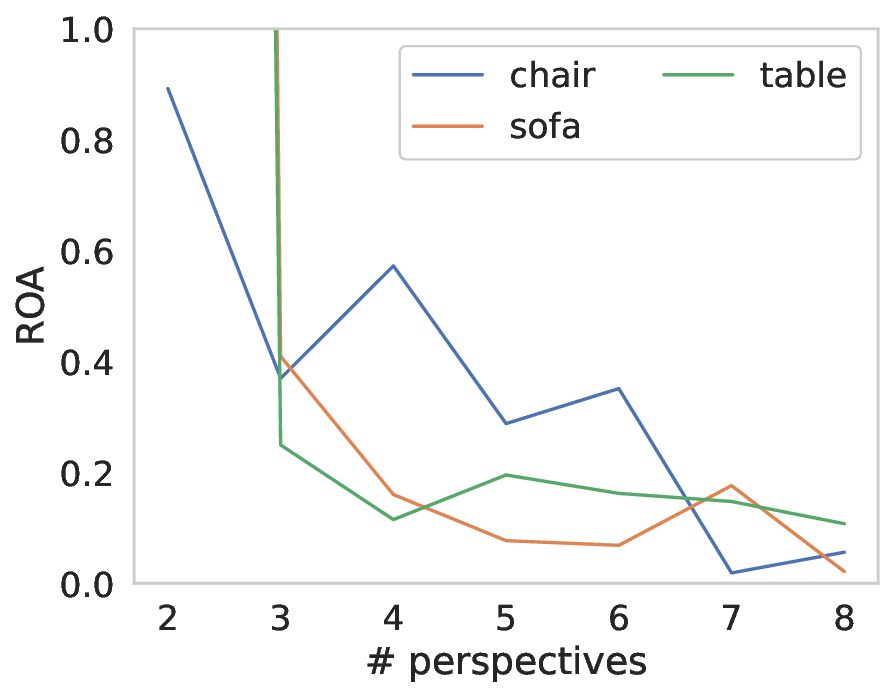}
         \caption{Fixed Proj.}
         \label{fig:pix3d_viewpoints_fixed_ROA}
     \end{subfigure}
     \begin{subfigure}{0.24\linewidth}
         \centering
         \includegraphics[width=\textwidth]{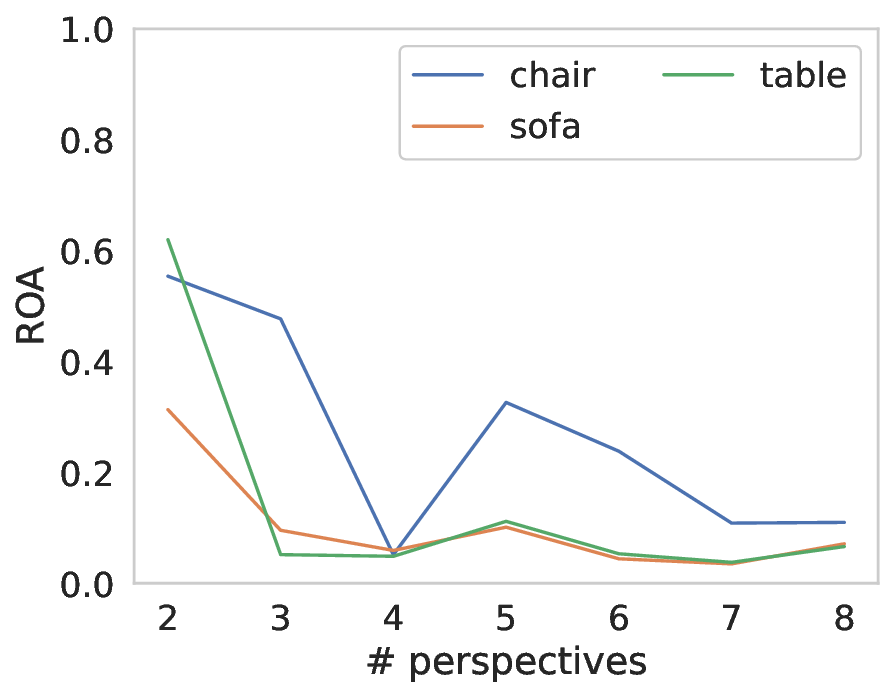}
         \caption{Variable Proj.}
         \label{fig:pix3d_viewpoints_var_ROA}
     \end{subfigure}
     \begin{subfigure}{0.24\linewidth}
         \centering
         \includegraphics[width=\textwidth]{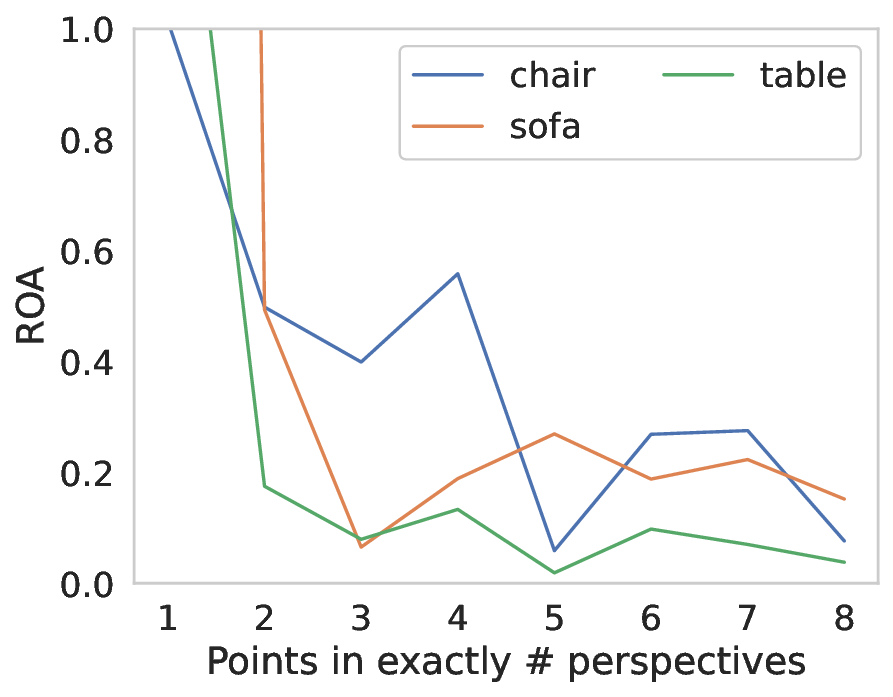}
         \caption{Fixed Proj.}
         \label{fig:pix3d_visible_fixed_ROA}
     \end{subfigure}
     \begin{subfigure}{0.24\linewidth}
         \centering
         \includegraphics[width=\textwidth]{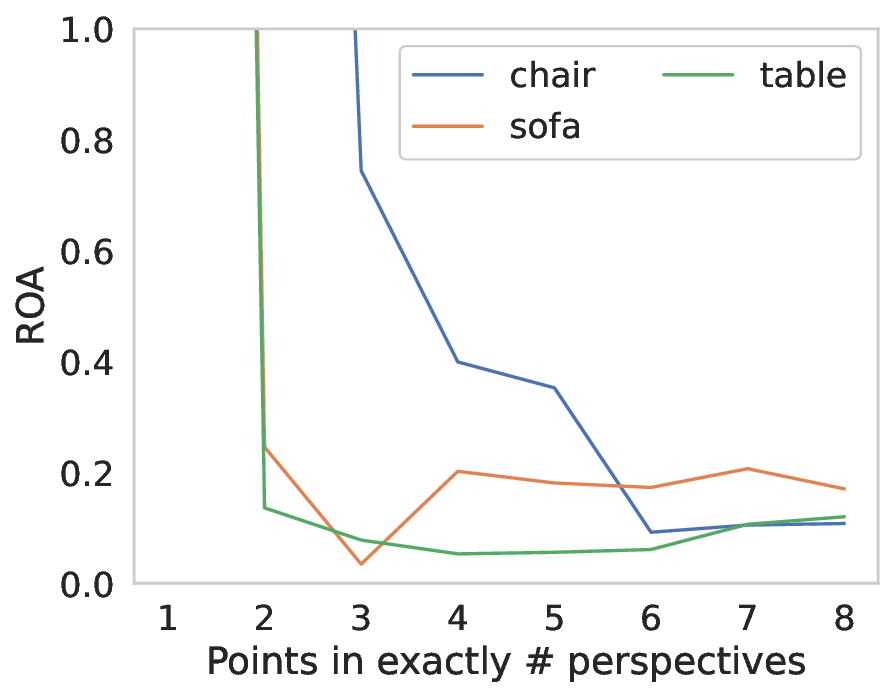}
         \caption{Variable Proj.}
         \label{fig:pix3d_visible_var_ROA}
     \end{subfigure}
    \caption{Metric analysis of 3DMPE on Pix3D dataset for varying viewpoints and visibility of points in those viewpoints. All experiments are done for 512 points. (a-d) shows the metric CD, (e-h) shows EMD and (i-l) shows ROA. For varying number of viewpoints (a-b, e-f, and i-j), each point is visible 1 less than the number of viewpoints (except 2 and 3 where all points are visible).}
    \label{fig:pix3d_viewpoints_metrics}
\end{figure*}

\begin{figure*}
     \centering
     \begin{subfigure}{0.32\linewidth}
         \centering
         \includegraphics[width=\textwidth]{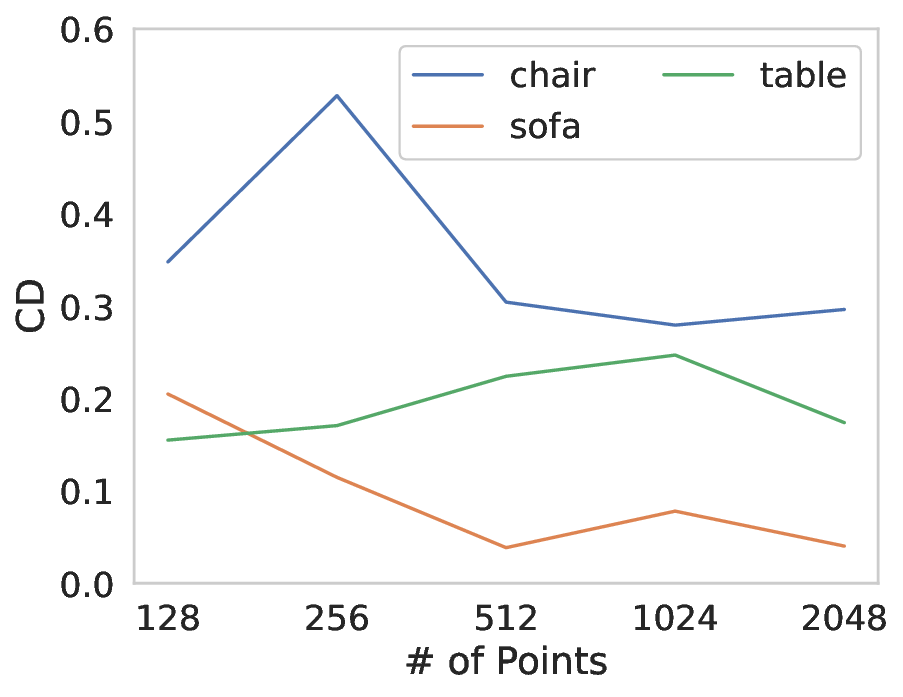}
         \caption{Fixed Projections}
         \label{fig:pix3d_points_fixed_Chamfer}
     \end{subfigure}
     \begin{subfigure}{0.32\linewidth}
         \centering
         \includegraphics[width=\textwidth]{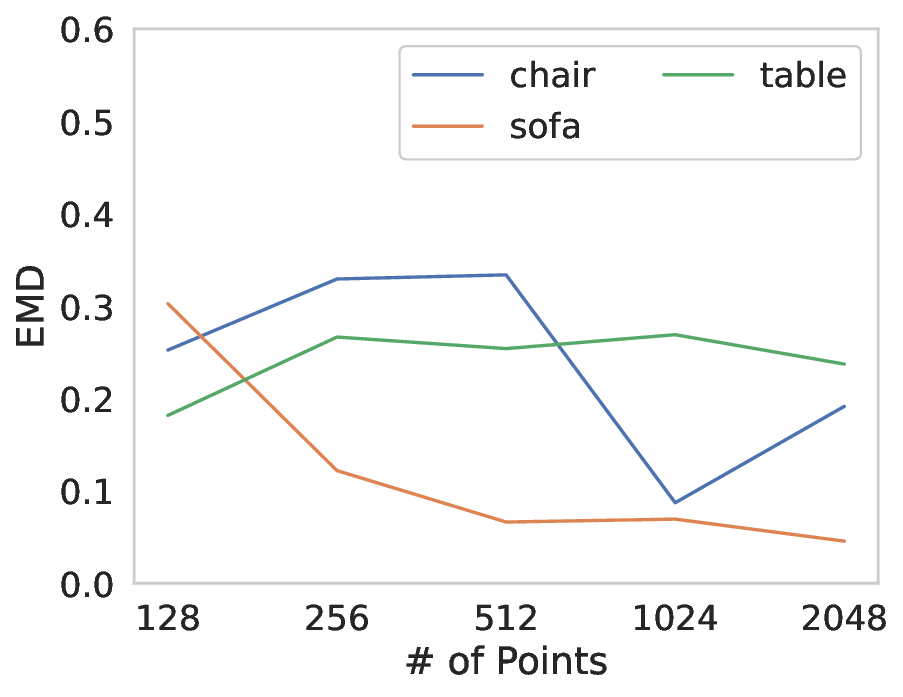}
         \caption{Fixed Projections}
         \label{fig:pix3d_points_fixed_EMD}
     \end{subfigure}
     \begin{subfigure}{0.32\linewidth}
         \centering
         \includegraphics[width=\textwidth]{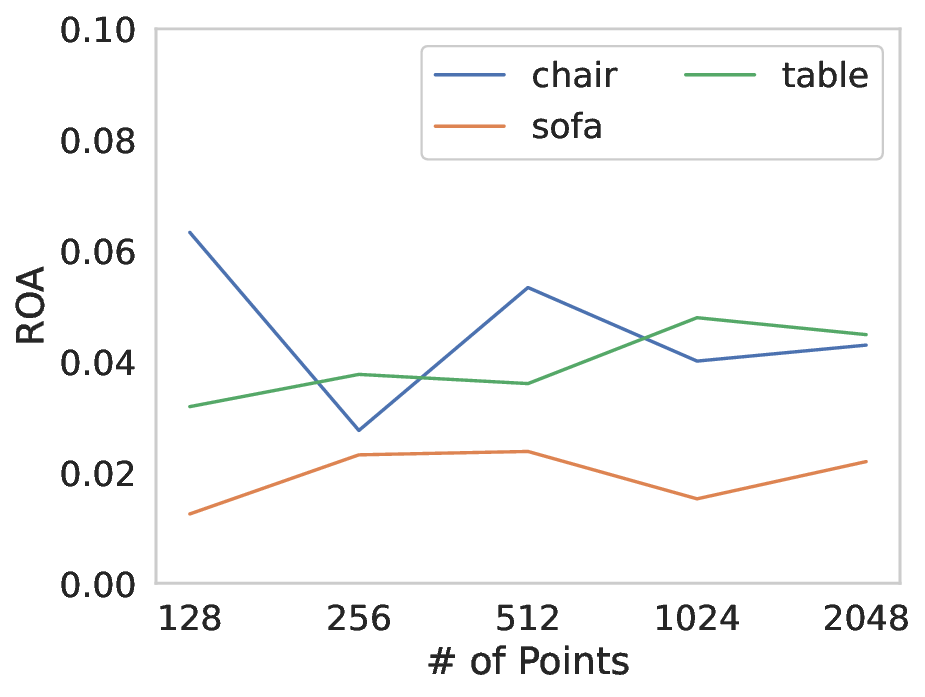}
         \caption{Fixed Projections}
         \label{fig:pix3d_points_fixed_ROA}
     \end{subfigure}
     
     \begin{subfigure}{0.32\linewidth}
         \centering
         \includegraphics[width=\textwidth]
         {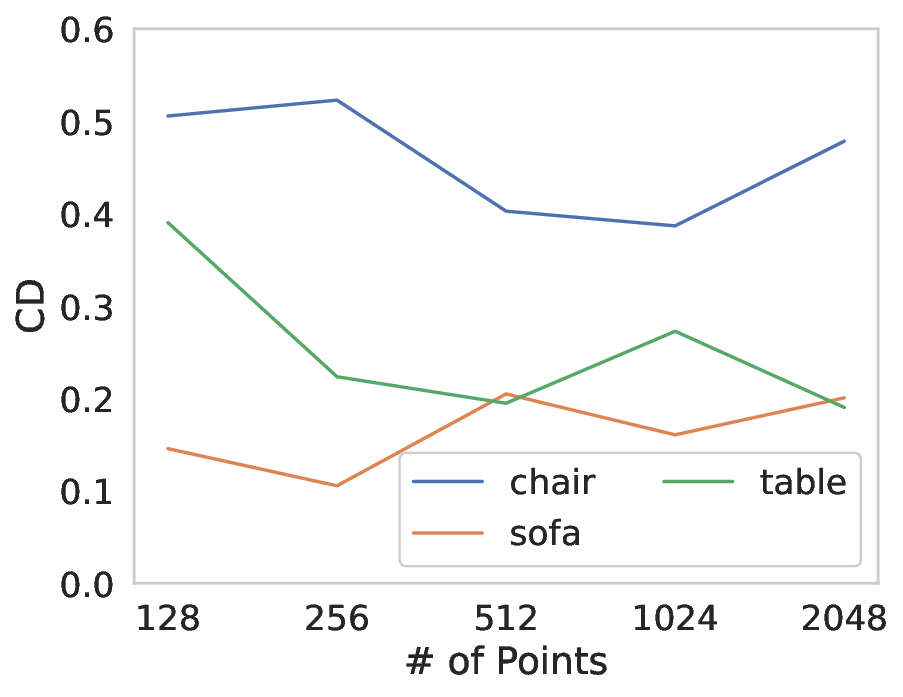}
         \caption{Variable Projections}
         \label{fig:pix3d_points_var_Chamfer}
     \end{subfigure}
     \begin{subfigure}{0.32\linewidth}
         \centering
         \includegraphics[width=\textwidth]{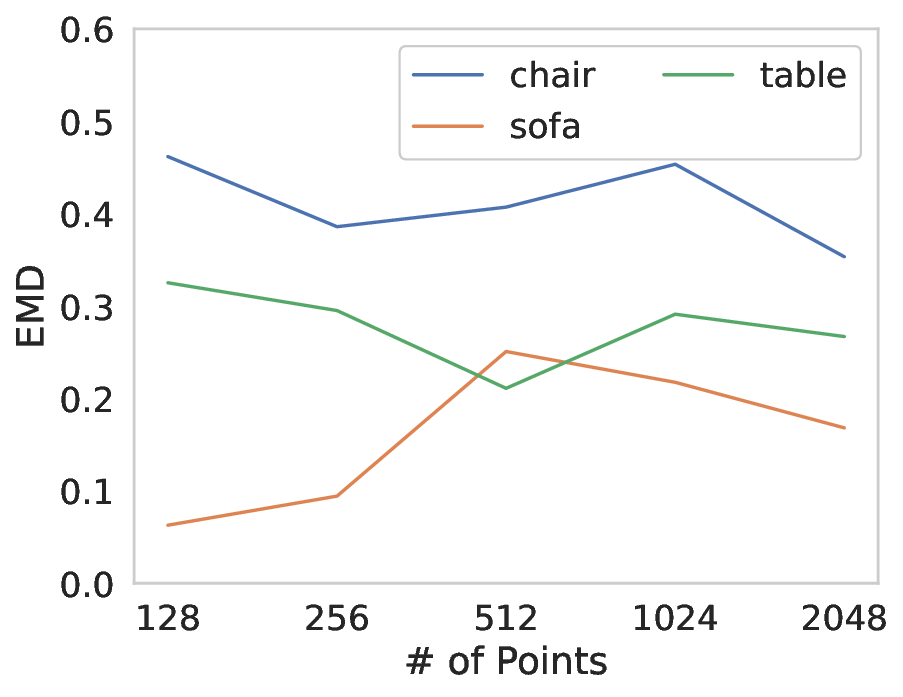}
         \caption{Variable Projections}
         \label{fig:pix3d_points_var_EMD}
     \end{subfigure}
     \begin{subfigure}{0.32\linewidth}
         \centering
         \includegraphics[width=\textwidth]{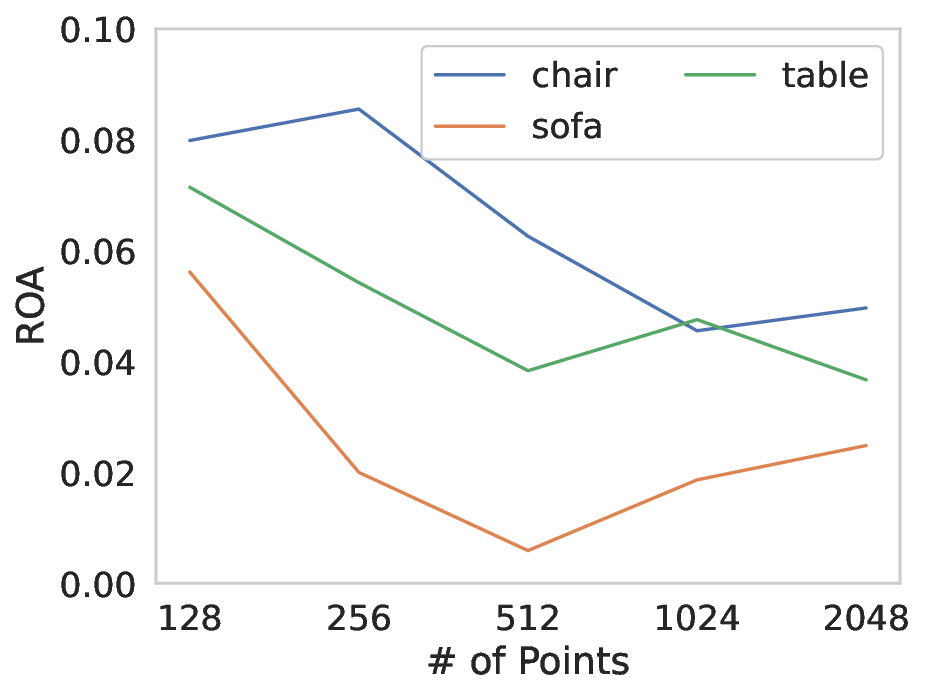}
         \caption{Variable Projections}
         \label{fig:pix3d_points_var_ROA}
     \end{subfigure}
    \caption{EMD, CD and ROA metrics on a varying number of points on Pix3D dataset. Each experiment is done with 4 viewpoints and each point is visible from at least 3 viewpoints. }
    \label{fig:pix3d_points_metrics}
\end{figure*}

\cref{fig:pix3d_viewpoints_metrics} and \cref{fig:pix3d_points_metrics} demonstrate the performance and runtime metrics for the Pix3D dataset for 3DMPE with both fixed and varying projections. The figures and numbers convey similar impressions to those with the ShapeNet data discussed in \cref{sec:scalability} (main text).

\begin{figure*}
     \centering
     \begin{subfigure}{0.24\linewidth}
         \centering
         \includegraphics[width=\textwidth]{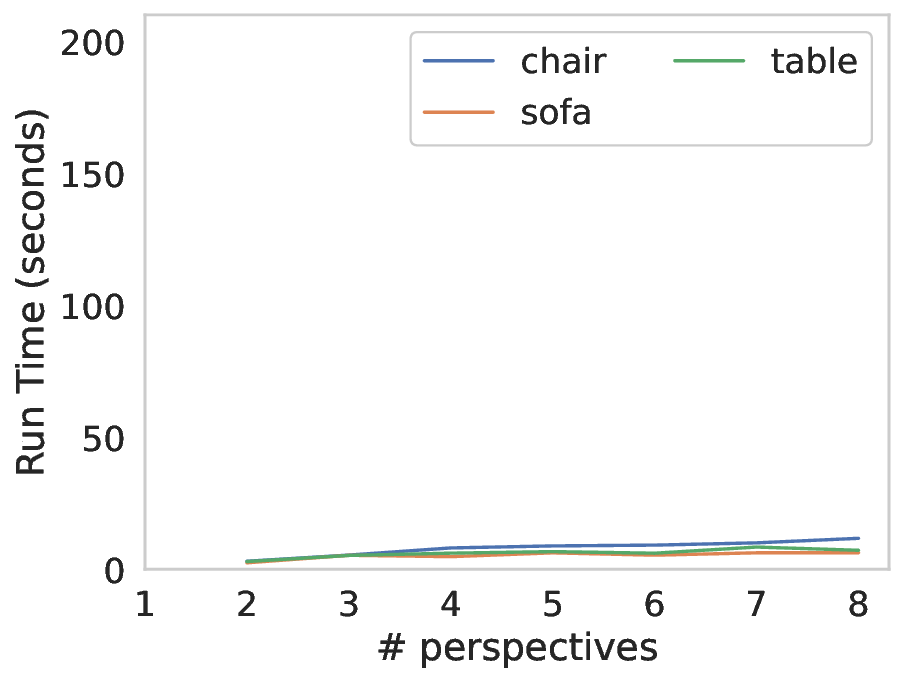}
         \caption{Fixed Proj.}
         \label{fig:pix3d_viewpoints_fixed_Run_Time}
     \end{subfigure}
     \begin{subfigure}{0.24\linewidth}
         \centering
         \includegraphics[width=\textwidth]{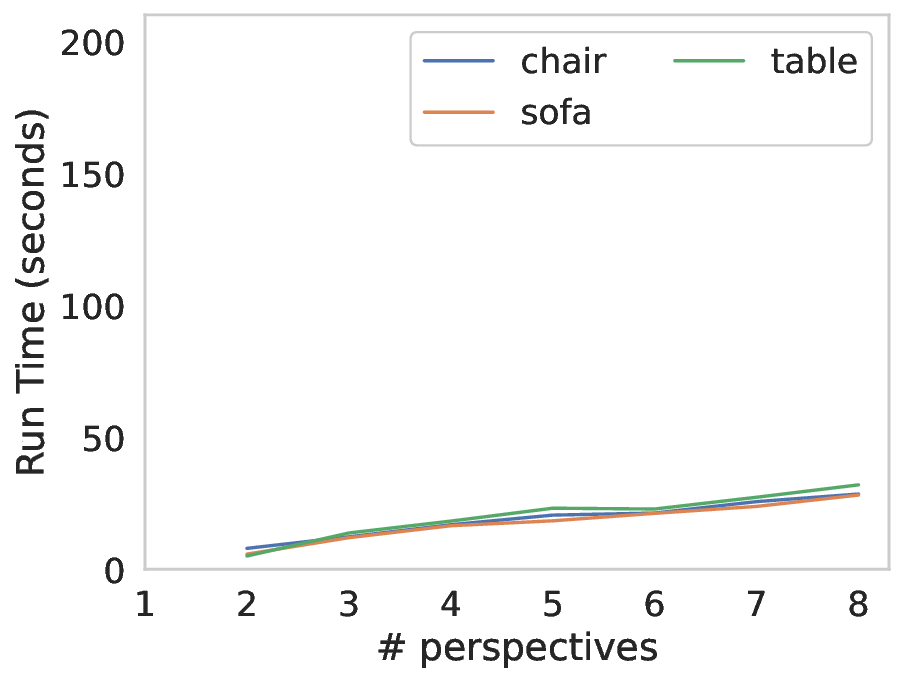}
         \caption{Variable Proj.}
         \label{fig:pix3d_viewpoints_var_Run_Time}
     \end{subfigure}
     \begin{subfigure}{0.24\linewidth}
         \centering
         \includegraphics[width=\textwidth]{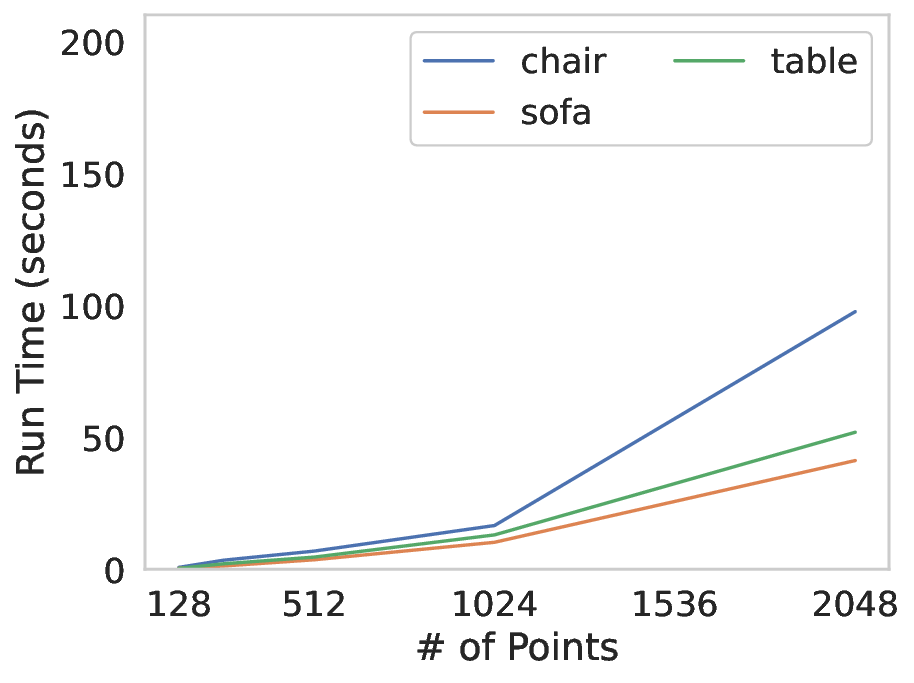}
         \caption{Fixed Proj.}
         \label{fig:pix3d_points_fixed_Run_Time}
     \end{subfigure}
     \begin{subfigure}{0.24\linewidth}
         \centering
         \includegraphics[width=\textwidth]{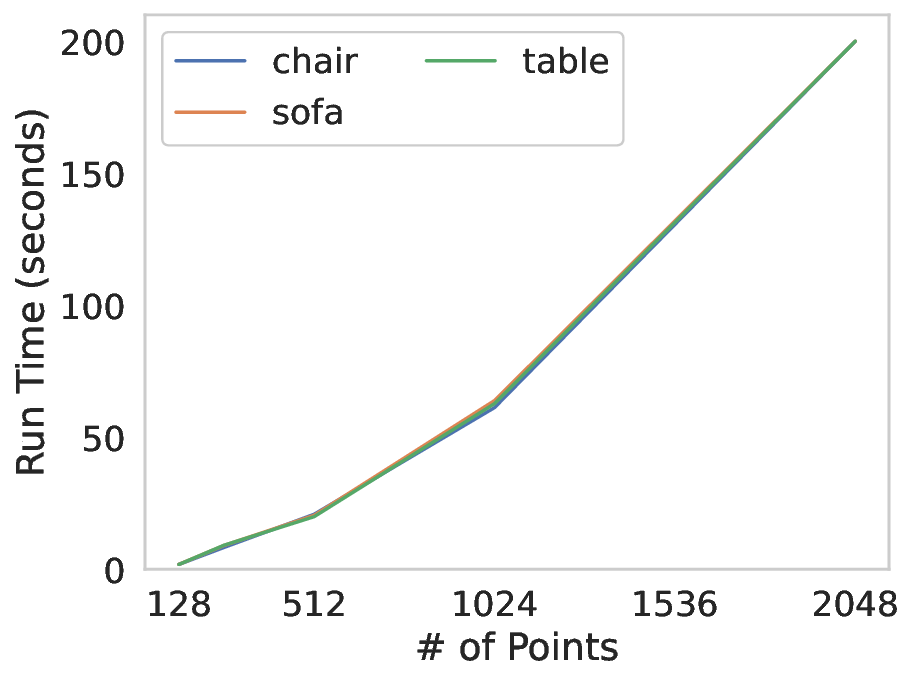}
         \caption{Variable Proj.}
         \label{fig:pix3d_points_var_Run_Time}
     \end{subfigure}
    \caption{Runtime (seconds) of 3DMPE on Pix3D dataset: (a-b) show the impact of changing the number of perspectives (using 512 points) and (c-d) show the impact of changing the number of points (with 4 projections).}
    \label{fig:pix3d_viewpoints_runtime}
\end{figure*}

\subsection{Exploratory Integration with COLMAP.}
\label{sec:colmap_comparison}

COLMAP~\cite{schonberger2016structure} is a full image-based Structure-from-Motion pipeline, whereas 3DMPE is for point clouds and assumes that point correspondences and visibility information are already available. We therefore explored whether COLMAP could be used to obtain the correspondences required by 3DMPE.

\begin{figure}
    \centering
    \includegraphics[width=0.45\linewidth]{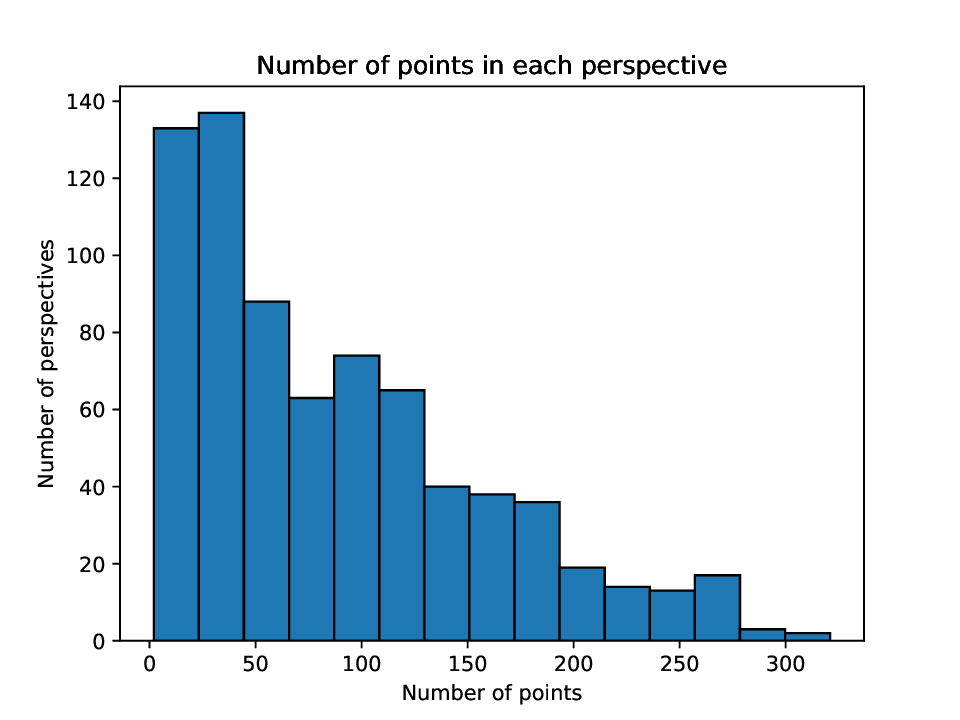}
    \caption{Maximum correspondences per perspective. With 703 perspectives and 65,373 feature points, the low correspondence count leads to poor reconstruction despite the noise robustness of 3DMPE.}
    \label{fig:sift_points}
\end{figure}

We first applied COLMAP v3.9 to benchmark image collections~\cite{realworlddataset,colmap_datasets}, using its SIFT~\cite{sift}-based feature extraction and matching pipeline. The resulting feature tracks were sparse, which led to poor 3DMPE reconstructions. \cref{fig:sift_points} shows the distribution of correspondence counts across 703 images of an airplane model, illustrating that most views contain relatively few matched points.

We next conducted a controlled experiment using a ShapeNet mesh. We rendered 250 images from different viewpoints using Blender~\cite{blender} and modified the rendering materials to provide sufficient texture variation for feature matching. COLMAP often failed to initialize reliably when fewer than 200 images were used. In contrast, for 3DMPE we generated four partially observed projections and used the known point correspondences inherited from the underlying mesh.

% \begin{wrapfigure}{r}{0.4\textwidth}
% \end{wrapfigure}

Although the two pipelines use substantially different inputs and assumptions, this experiment illustrates the practical role of correspondence quality. For the selected ShapeNet object, COLMAP reconstructed a point cloud with Chamfer Distance $23.1$ using $250$ rendered images, whereas 3DMPE achieved Chamfer Distance $5.79$ from four projections with known correspondences; see \cref{tab:colmap}. These results should not be interpreted as a direct benchmark comparison, but rather as evidence that reliable correspondence extraction remains a central challenge for integrating 3DMPE into an end-to-end image-based reconstruction pipeline.

\begin{table}[t]
\centering
\caption{ShapeNet chair reconstruction using COLMAP and 3DMPE.}
\label{tab:colmap}
\begin{tblr}  {
    % colspec = {|c|c|c|c|c|c|c|},
    colspec = {c c c c c c},
    rowspec = {Q[c,m]Q[c,m]Q[c,m]},
    rowsep = 0.5pt,
    hlines = {0.5pt},
    % vlines = {black, 1pt},
}
Pipeline & Input & \# Perspectives & Reconstruction & EMD & CD \\
COLMAP
& \adjustbox{valign=m}{\includegraphics[width=1.0cm,keepaspectratio]{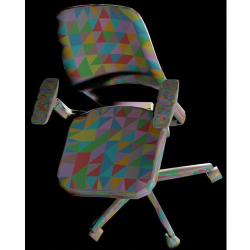}}
& 250
& \adjustbox{valign=m}{\includegraphics[width=1.0cm,keepaspectratio]{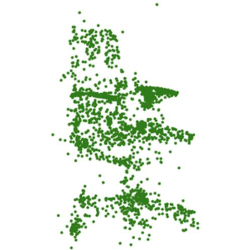}}
& 19.7
& 23.1 \\

3DMPE
& \adjustbox{valign=m}{\includegraphics[width=1.0cm,keepaspectratio]{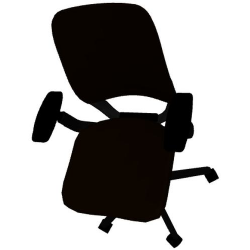}}
& 4
& \adjustbox{valign=m}{\includegraphics[width=1.0cm,keepaspectratio]{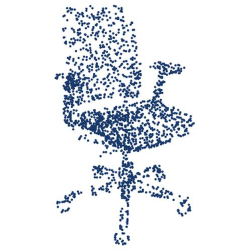}}
& 5.09
& 5.79 \\
\end{tblr}
\end{table}
% \lstinputlisting[
%     caption=Example Python Code,
%     label=lst:py:blender,
%     title=\lstname,
% ]{blender.py}

\subsection{Implementation Details}
\label{sec:implementaion}

We implemented 3DMPE in Python. Our implementation relies on the \texttt{pyoint} package for ICP alignment and on the \texttt{scikit-learn} manifold module for baseline methods. Numerical computations were performed using \texttt{NumPy}, \texttt{SciPy}, and \texttt{pandas}. We used \texttt{trimesh} for reading and sampling both the ShapeNet and Pix3D datasets, and TensorFlow for the alignment algorithm.

We conducted approximately 11{,}000 experiments across different parameter settings. 
The source code is available at \href{https://github.com/rahatzamancse/3DMPE}{https://github.com/rahatzamancse/3DMPE}.

%-------------------------------------------------------------------------

\section{Conclusions, Limitations, and Future Work}

We presented 3DMPE, a training-free method for reconstructing a 3D point cloud from two or more partially observed 2D projections with cross-view point correspondences. We considered fixed-projection and variable-projection settings, in which the projection maps are known or jointly estimated, respectively. Experiments on ShapeNet and Pix3D evaluated reconstruction quality under varying initialization schemes, numbers of views, levels of point visibility, and noise in distances and correspondences.

A primary limitation of 3DMPE is its assumption that cross-view point correspondences are available. Thus, 3DMPE is not an end-to-end image-to-3D reconstruction method; rather, it addresses the geometric reconstruction stage after correspondence estimation. 
In preliminary experiments using correspondences obtained from a COLMAP-based pipeline~\cite{schonberger2016structure}, unreliable feature matches produced inputs that were insufficient for accurate reconstruction. Although our controlled experiments demonstrate robustness to local correspondence perturbations, integrating 3DMPE with robust feature matching, tracking, and camera-estimation methods remains an important direction for future work. At the same time, 3DMPE provides a flexible reconstruction module that does not require category-specific training data.

%-------------------------------------------------------------------------

% \section*{Acknowledgements}
% Please insert your acknowledgments here.

% \newpage 

% ---- Bibliography ----
%
% BibTeX users should specify bibliography style 'splncs04'.
% References will then be sorted and formatted in the correct style.
%
\bibliographystyle{splncs04}
\bibliography{3DMPE}

\clearpage
% \setcounter{page}{1}
% \maketitlesupplementary

% \onecolumn

\noindent \textbf{Appendix}

\section*{Alternative Computation of the ROA Metric}
\begin{figure*}
    \centering
    \begin{subfigure}{0.35\linewidth}
         \centering
         \includegraphics[width=\textwidth]{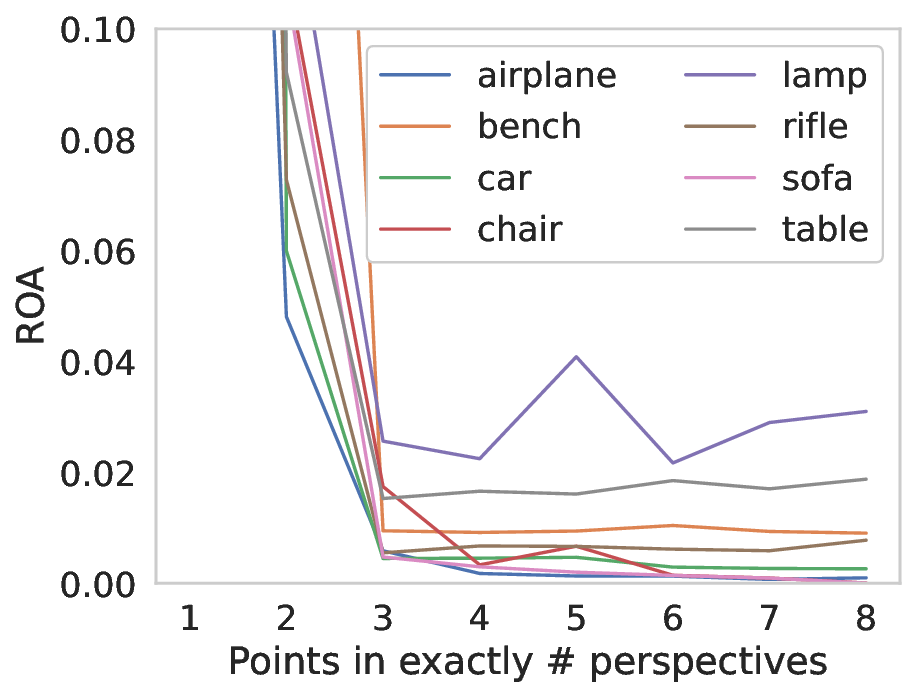}
         \caption{Fixed Projections}
         \label{fig:visible_fixed_ROA}
     \end{subfigure}
     \begin{subfigure}{0.35\linewidth}
         \centering
         \includegraphics[width=\textwidth]{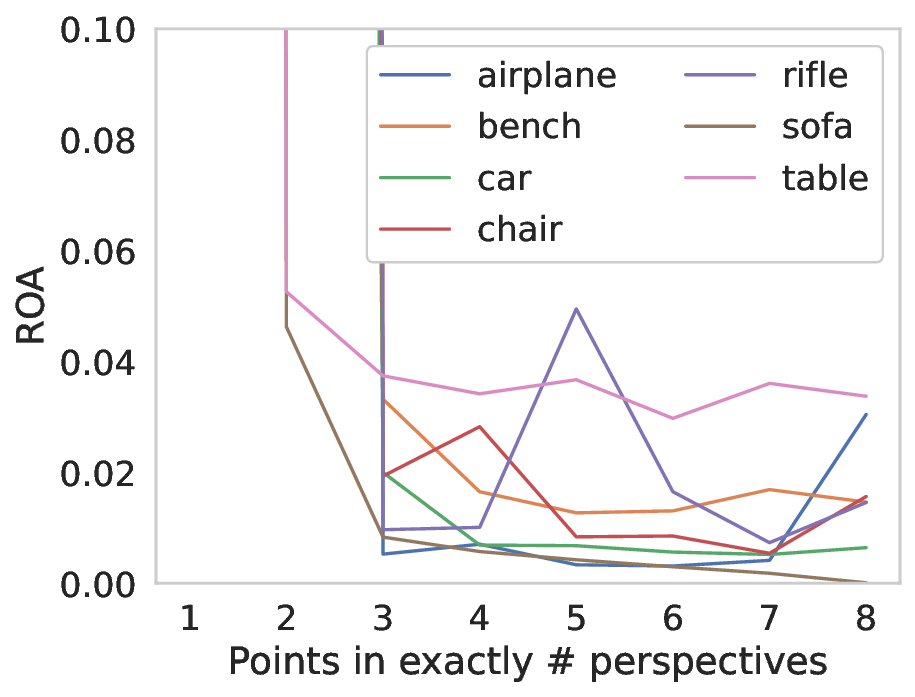}
         \caption{Variable Projections}
         \label{fig:visible_var_ROA}
     \end{subfigure}
     
     \begin{subfigure}{0.35\linewidth}
         \centering
         \includegraphics[width=\textwidth]{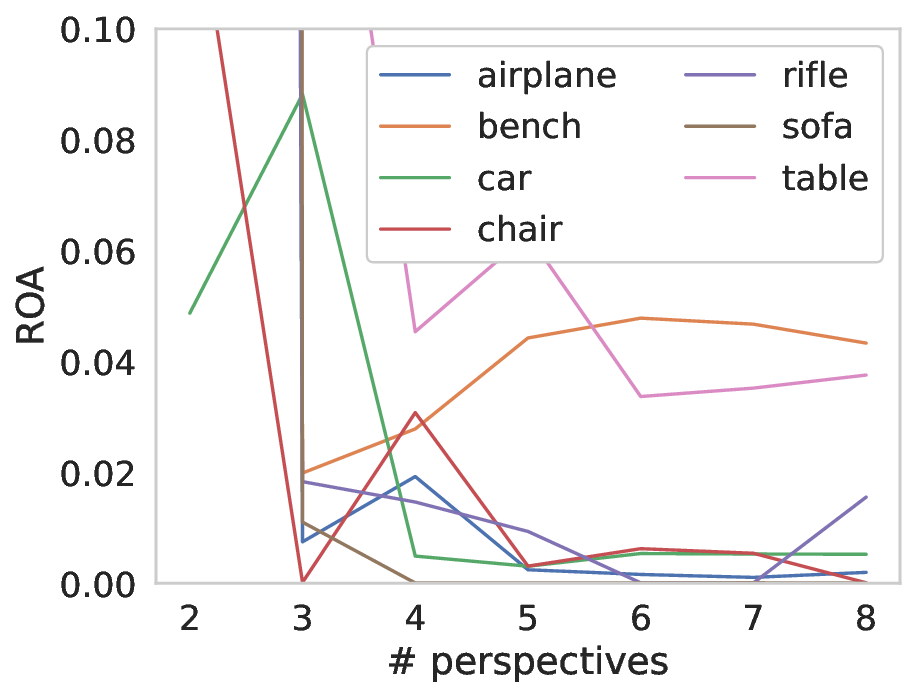}
         \caption{Fixed Projections}
         \label{fig:viewpoints_fixed_ROA}
     \end{subfigure}
     \begin{subfigure}{0.35\linewidth}
         \centering
         \includegraphics[width=\textwidth]{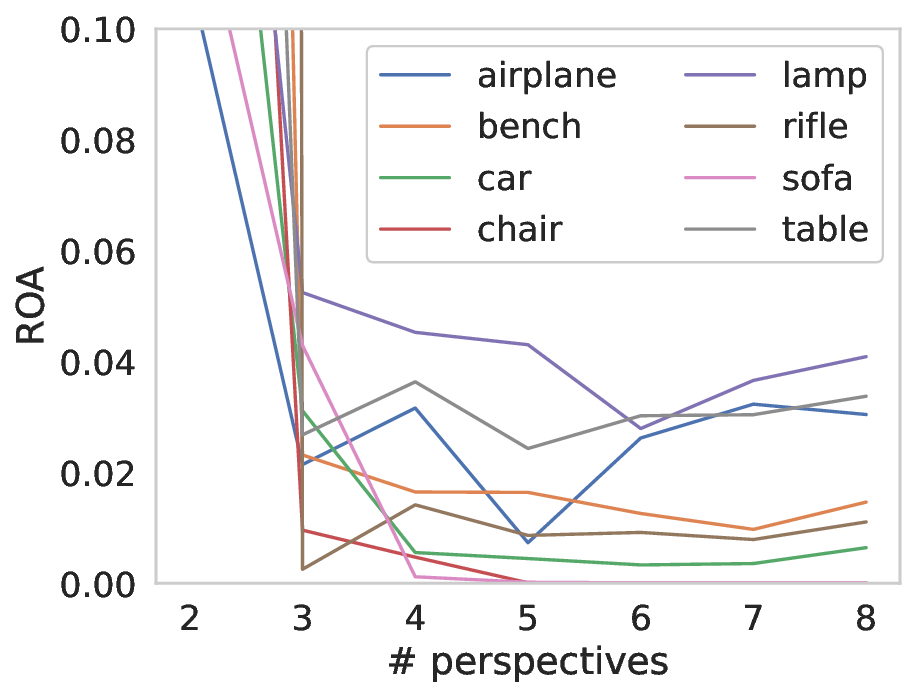}
         \caption{Variable Projections}
         \label{fig:viewpoints_var_ROA}
     \end{subfigure}
    \caption{ROA Metric analysis of 3DMPE on ShapeNet dataset for (a-b) points visible in a number of viewpoints and (c-d) varying number of viewpoints. All experiments are done for 512 points with 8 viewpoints. For (c-d) each point is visible 1 less than the number of viewpoints (except 2 and 3 where all points are visible).}
    \label{fig:visible_ROA}
\end{figure*}

In \cref{sec:evaluation}, we proposed the CD, EMD, and ROA metrics for evaluation. In addition to the iterative approach for computing ROA, here we discuss how one can compute the ROA using 
% For calculating ROA, we also used another approach with 
Singular Value Decomposition (SVD). 
In our experiments, both the iterative approach and SVD approach described below produce similar results, although the SVD approach takes a bit  longer to compute.
% but in general  that also gave the same result but avoids the iterative approach. 
The idea is as follows: We start by mean shifting both point clouds to the origin. 
That is, compute 
\begin{equation}
    \mu_{X} = \frac{1}{N} \sum_{i=1}^{N} X_i, \text{ } 
    \mu_{\hat{X}} = \frac{1}{N} \sum_{i=1}^{N} \hat{X}_i 
\end{equation}
and subtract them from the corresponding point clouds. We further compute the following matrix:
\begin{equation*}
    H = (X - \mu_{X})(\hat{X} - \mu_{\hat{X}})^T.
\end{equation*}
We continue by computing the Singular Value Decomposition (SVD)~\cite{10.2307/2132388} of $H$, 
\begin{equation}
[U, S, V] = SVD(H)
\end{equation}
We proceed by computing the rotation matrix $R \in \RR_{3 \times 3}$
\begin{equation}
R = VU^T 
\end{equation}
and the translation matrix 
\begin{equation}
t = \mu_{\bar{X}} - R \times \mu_{X}
\end{equation}
In this process, $X$ and $\bar{X}$ are the two point clouds to be aligned and $R$ and $t$ are the computed rotation and translation matrices for aligning the point clouds. We compute ROA values for several ShapeNet examples, using different numbers of perspectives while changing the number of viewpoints from which each point is visible, for both fixed and variable projections; see \cref{fig:visible_ROA}. 
% {\color{red} 
The results are consistent with those based on the CD and EMD (\cref{fig:viewpoints_metrics} and \cref{fig:visible_metrics}) metric -- 3D reconstruction is very accurate, as long as most points are visible from 3 or more viewpoints.
% }

\begin{figure*}
    \centering
    \begin{subfigure}{0.35\linewidth}
         \centering
         \includegraphics[width=\textwidth]{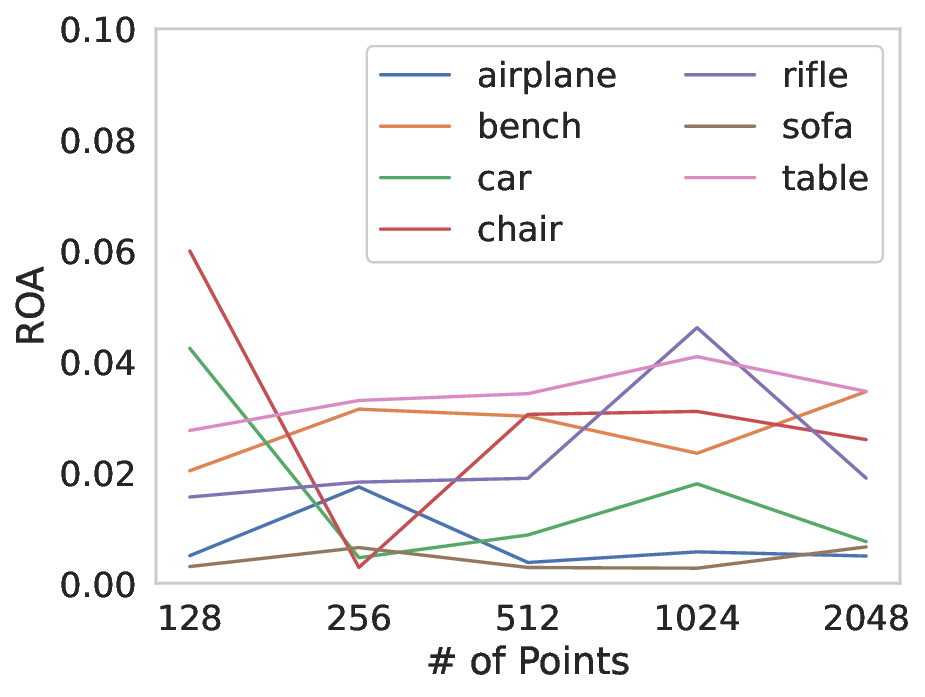}
         \caption{Fixed Projections}
         \label{fig:points_fixed_ROA}
     \end{subfigure}
     \begin{subfigure}{0.35\linewidth}
         \centering
         \includegraphics[width=\textwidth]{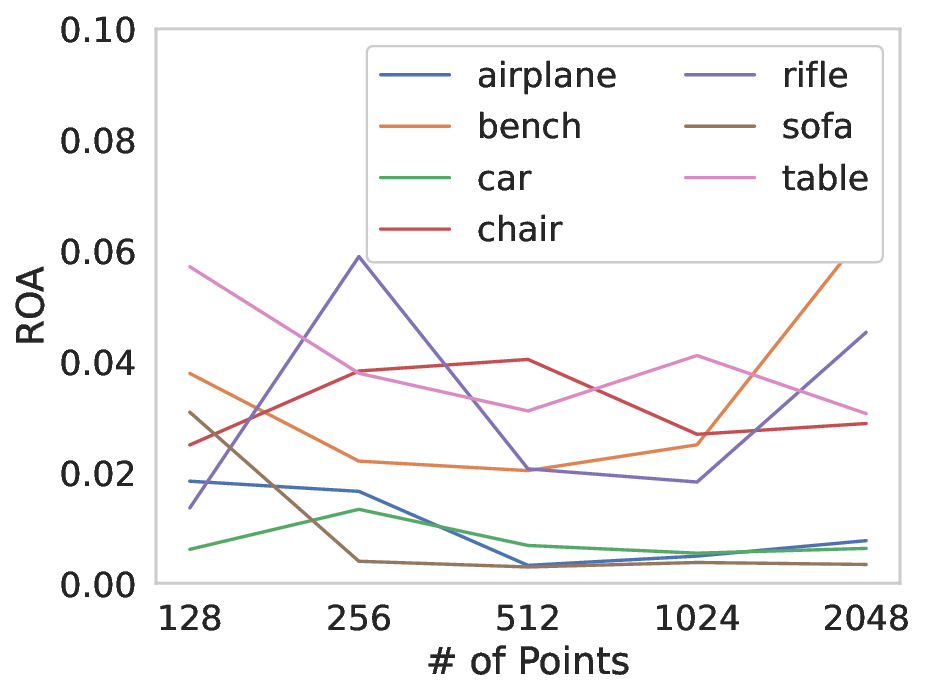}
         \caption{Variable Projections}
         \label{fig:points_var_ROA}
     \end{subfigure}
    \caption{ROA metrics on a varying number of points on ShapeNet dataset (similar to figure \ref{fig:points_fixed_Chamfer} and \ref{fig:points_var_Chamfer}). Each experiment is done with 4 viewpoints and each point is visible from at least 3 viewpoints. }
    \label{fig:points_ROA}
\end{figure*}

We also compute ROA values when varying number of points given as input to the reconstruction; see \cref{fig:points_ROA}. Here again the results are consistent with those based on the CD and EMD measures (\cref{fig:points_metrics}).

\end{document}

%% file: tables/allimages.tex
\begin{table}
\centering
\caption{3D reconstruction results for qualitative comparison between 3DMPE with fixed angle projections, PSGN, and 3D-LMNet. The baseline is used from \cite{mandikal20183D}.}
\label{tab:allimages}
\begin{tabular}{ccccccc}
Input & Ground Truth & PSGN \cite{fan2017apoint} & BL \cite{mandikal20183D} & 3D-LMNet \cite{mandikal20183D} & 3DMPE(F) & 3DMPE(V) \\ \hline
\includegraphics[width=0.80cm,height=0.80cm,keepaspectratio]{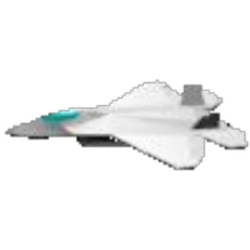}
& \includegraphics[width=0.80cm,height=0.80cm,keepaspectratio]{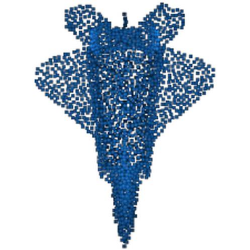} 
& \includegraphics[width=0.80cm,height=0.80cm,keepaspectratio]{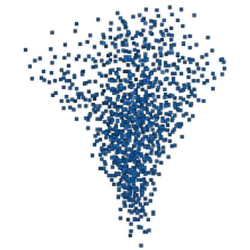} 
& \includegraphics[width=0.80cm,height=0.80cm,keepaspectratio]{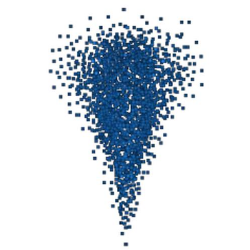} 
& \includegraphics[width=0.80cm,height=0.80cm,keepaspectratio]{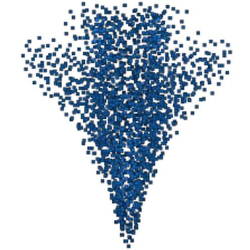} 
& \includegraphics[width=0.80cm,height=0.80cm,keepaspectratio]{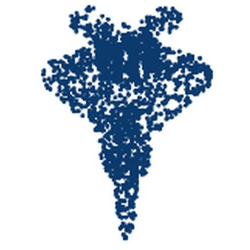}
& \includegraphics[width=0.80cm,height=0.80cm,keepaspectratio]{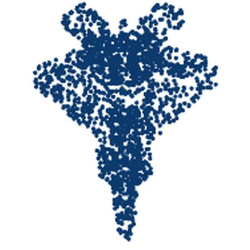} \\

\includegraphics[width=0.80cm,height=0.80cm,keepaspectratio]{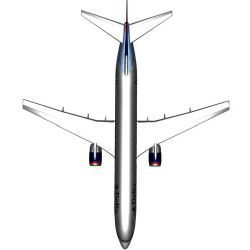}
& \includegraphics[width=0.80cm,height=0.80cm,keepaspectratio]{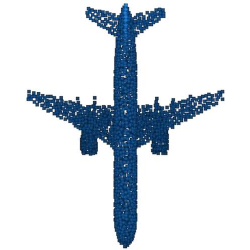} 
& \includegraphics[width=0.80cm,height=0.80cm,keepaspectratio]{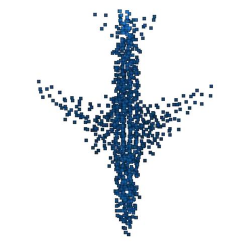} 
& \includegraphics[width=0.80cm,height=0.80cm,keepaspectratio]{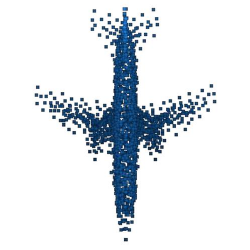} 
& \includegraphics[width=0.80cm,height=0.80cm,keepaspectratio]{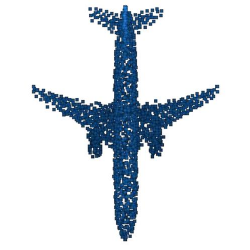} 
& \includegraphics[width=0.80cm,height=0.80cm,keepaspectratio]{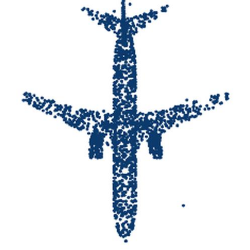}
& \includegraphics[width=0.80cm,height=0.80cm,keepaspectratio]{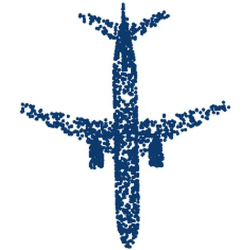} \\

\includegraphics[width=0.80cm,height=0.80cm,keepaspectratio]{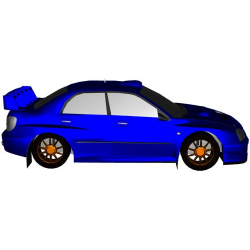}
& \includegraphics[width=0.80cm,height=0.80cm,keepaspectratio]{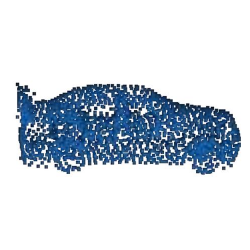} 
& \includegraphics[width=0.80cm,height=0.80cm,keepaspectratio]{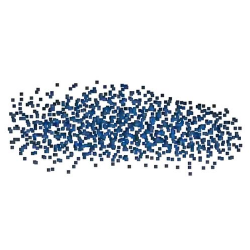} 
& \includegraphics[width=0.80cm,height=0.80cm,keepaspectratio]{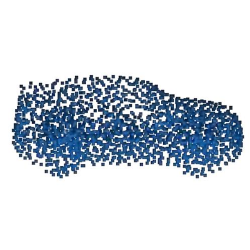} 
& \includegraphics[width=0.80cm,height=0.80cm,keepaspectratio]{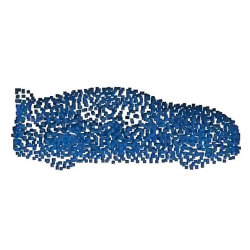} 
& \includegraphics[width=0.80cm,height=0.80cm,keepaspectratio]{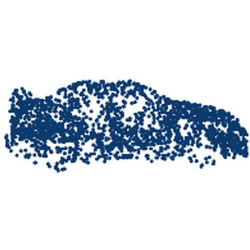}
& \includegraphics[width=0.80cm,height=0.80cm,keepaspectratio]{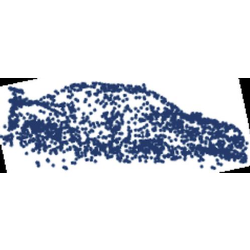} \\

\includegraphics[width=0.80cm,height=0.80cm,keepaspectratio]{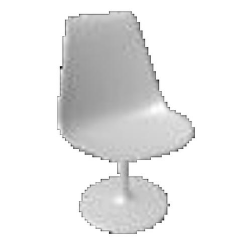}
& \includegraphics[width=0.80cm,height=0.80cm,keepaspectratio]{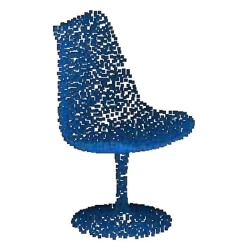} 
& \includegraphics[width=0.80cm,height=0.80cm,keepaspectratio]{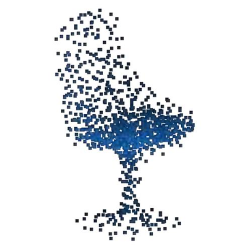} 
& \includegraphics[width=0.80cm,height=0.80cm,keepaspectratio]{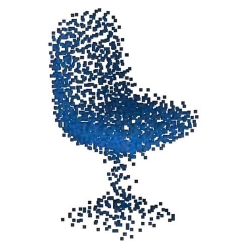} 
& \includegraphics[width=0.80cm,height=0.80cm,keepaspectratio]{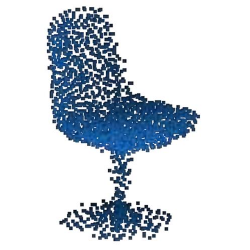} 
& \includegraphics[width=0.80cm,height=0.80cm,keepaspectratio]{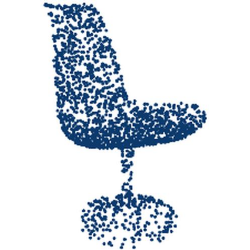}
& \includegraphics[width=0.80cm,height=0.80cm,keepaspectratio]{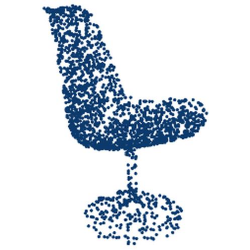} \\

% \includegraphics[width=0.80cm,height=0.80cm,keepaspectratio]{figures/3-4.pdf}
% & \includegraphics[width=0.80cm,height=0.80cm,keepaspectratio]{figures/3-4-gt.pdf} 
% & \includegraphics[width=0.80cm,height=0.80cm,keepaspectratio]{figures/3-4-psgn.pdf} 
% & \includegraphics[width=0.80cm,height=0.80cm,keepaspectratio]{figures/3-4-baseline.pdf} 
% & \includegraphics[width=0.80cm,height=0.80cm,keepaspectratio]{figures/3-4-3dlmnet.pdf} 
% & \includegraphics[width=0.80cm,height=0.80cm,keepaspectratio]{figures/3-4-mpse-fixed.pdf} \\

\includegraphics[width=0.80cm,height=0.80cm,keepaspectratio]{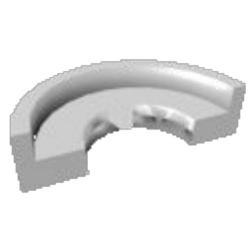}
& \includegraphics[width=0.80cm,height=0.80cm,keepaspectratio]{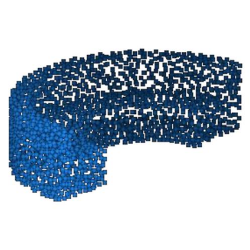} 
& \includegraphics[width=0.80cm,height=0.80cm,keepaspectratio]{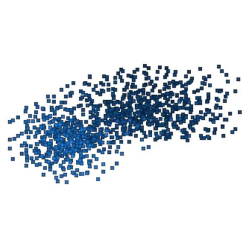} 
& \includegraphics[width=0.80cm,height=0.80cm,keepaspectratio]{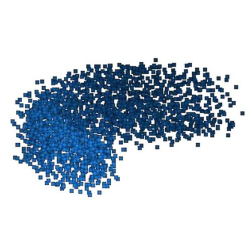} 
& \includegraphics[width=0.80cm,height=0.80cm,keepaspectratio]{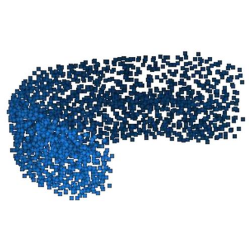} 
& \includegraphics[width=0.80cm,height=0.80cm,keepaspectratio]{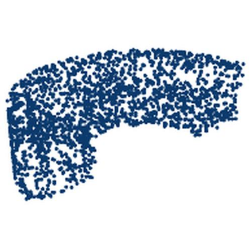}
& \includegraphics[width=0.80cm,height=0.80cm,keepaspectratio]{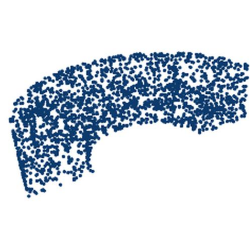} \\

\includegraphics[width=0.80cm,height=0.80cm,keepaspectratio]{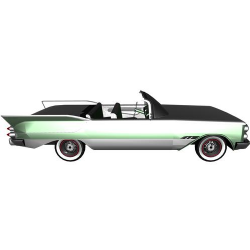}
& \includegraphics[width=0.80cm,height=0.80cm,keepaspectratio]{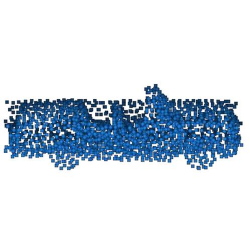} 
& \includegraphics[width=0.80cm,height=0.80cm,keepaspectratio]{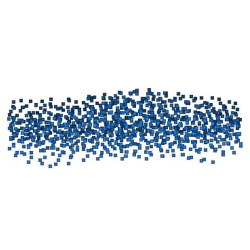} 
& \includegraphics[width=0.80cm,height=0.80cm,keepaspectratio]{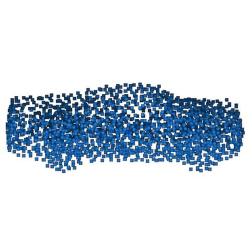} 
& \includegraphics[width=0.80cm,height=0.80cm,keepaspectratio]{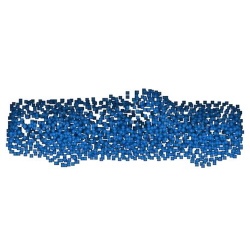} 
& \includegraphics[width=0.80cm,height=0.80cm,keepaspectratio]{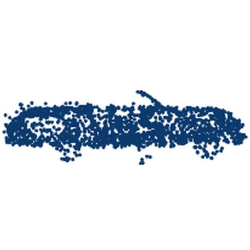}
& \includegraphics[width=0.80cm,height=0.80cm,keepaspectratio]{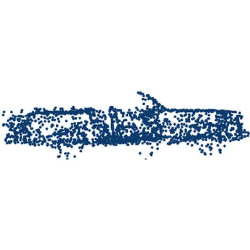} \\

% \includegraphics[width=0.80cm,height=0.80cm,keepaspectratio]{figures/3-7.pdf}
% & \includegraphics[width=0.80cm,height=0.80cm,keepaspectratio]{figures/3-7-gt.pdf} 
% & \includegraphics[width=0.80cm,height=0.80cm,keepaspectratio]{figures/3-7-psgn.pdf} 
% & \includegraphics[width=0.80cm,height=0.80cm,keepaspectratio]{figures/3-7-baseline.pdf} 
% & \includegraphics[width=0.80cm,height=0.80cm,keepaspectratio]{figures/3-7-3dlmnet.pdf} 
% & \includegraphics[width=0.80cm,height=0.80cm,keepaspectratio]{figures/3-7-mpse-fixed.pdf} \\

\includegraphics[width=0.80cm,height=0.80cm,keepaspectratio]{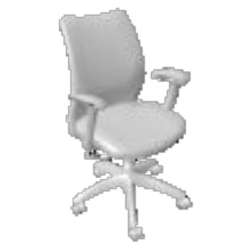}
& \includegraphics[width=0.80cm,height=0.80cm,keepaspectratio]{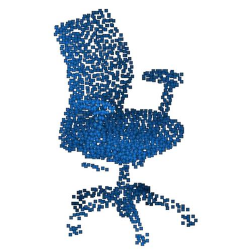} 
& \includegraphics[width=0.80cm,height=0.80cm,keepaspectratio]{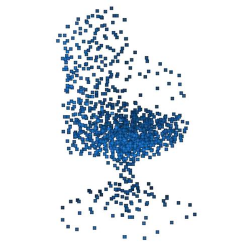} 
& \includegraphics[width=0.80cm,height=0.80cm,keepaspectratio]{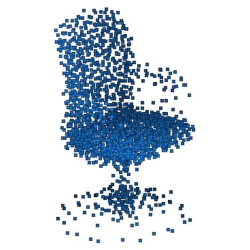} 
& \includegraphics[width=0.80cm,height=0.80cm,keepaspectratio]{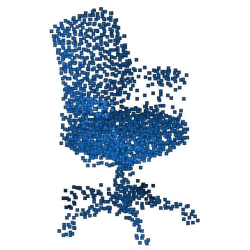} 
& \includegraphics[width=0.80cm,height=0.80cm,keepaspectratio]{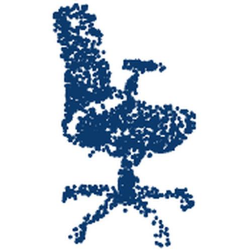}
& \includegraphics[width=0.80cm,height=0.80cm,keepaspectratio]{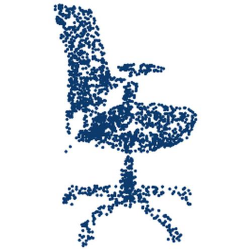} \\

\end{tabular}
\end{table}

%% file: tables/allmetrics.tex
\begin{table*}[]
\caption{Demonstration of the 3DMPE algorithm (last 2 columns) for 8 different 3D models from the Shapenet dataset. For each, 2048 points are used with 4 viewpoints. Projections are created with a raytracing algorithm.}
\label{tab:allmetrics}
\centering
\resizebox{0.95\textwidth}{!}{%
\begin{tabular}{|c|cccccc|cccccc|c|}
\hline
\multirow{2}{*}{Category} & \multicolumn{6}{c|}{Chamfer}           & \multicolumn{6}{c|}{EMD} \\ \cline{2-13}
            & Baseline \cite{mandikal20183D} & PSGN \cite{fan2017apoint} & 3D-LMNet \cite{mandikal20183D} & Our BL & 3DMPE(F) & 3DMPE(V) & Baseline & PSGN  & 3D-LMNet & Our BL & 3DMPE(F) & 3DMPE(V) \\ \hline
airplane    & 3.61     & 3.74                      & 3.34                           & 3.41  & \textbf{0.55} & 1.58 & 7.42   & 6.38  & 4.77  & 4.86  & \textbf{0.02} & 1.41 \\
bench       & 4.70     & 4.63                      & 4.55                           & 3.89  & \textbf{1.07} & 1.94 & 5.66   & 5.88  & 4.99  & 4.09  & \textbf{1.25} & 1.65 \\
car         & 4.67     & 5.20                      & 4.55                           & 4.34  & \textbf{1.02} & 4.01 & 4.74   & 4.87  & 4.10  & 3.8   & \textbf{0.01} & 4.64 \\
chair       & 6.51     & 6.39                      & 6.41                           & 5.41  & \textbf{1.85} & 4.28 & 8.99   & 9.63  & 8.02  & 3.73  & \textbf{1.03} & 1.21 \\
lamp        & 7.32     & 6.33                      & 7.10                           & 7.47  & \textbf{3.14} & 6.36 & 20.96  & 16.17 & 15.8  & 9.42  & \textbf{2.86} & 7.01 \\
rifle       & 2.99     & 2.91                      & 2.75                           & 2.22  & \textbf{2.11} & 2.14 & 9.30   & 8.48  & 6.08  & 2.45  & \textbf{0.72} & 1.94 \\
sofa        & 6.11     & 6.98                      & 5.85                           & 5.70  & \textbf{0.06} & 0.39 & 6.40   & 7.42  & 5.65  & 4.32  & \textbf{0.03} & 0.06 \\
table       & 6.16     & 6.00                      & 6.05                           & 6.10  & \textbf{1.27} & 2.01 & 9.51   & 8.40  & 7.82  & 4.85  & \textbf{1.99} & 2.68 \\ \hline
mean        & 5.99     & 5.54                      & 5.40                           & 4.82  & \textbf{1.56} & 2.84 & 7.82   & 7.20  & 7.00  & 4.69  & \textbf{0.99} & 2.58 \\ \hline
\end{tabular}%
}
\label{tab:shapenetmetrics}
\end{table*}

%% file: tables/pix3dallimages.tex
\begin{table*}[h]
\caption{Demonstration of the 3DMPE algorithm (last 2 columns) for different 3D models from the Pix3D dataset~\cite{pix3d}. For all experiments, 2048 points are used with 5 viewpoints. Projections are created with a raytracing algorithm.}
\label{tab:pix3dallimages}
\centering
{\small
\begin{tabular}{ccccccc}
Input & Ground Truth & PSGN & Baseline & 3D-LMNet & 3DMPE(F) & 3DMPE(V) \\ \hline

\includegraphics[width=1.5cm,height=1.5cm,keepaspectratio]{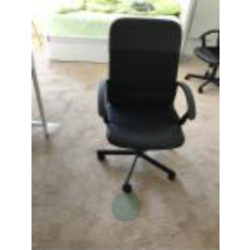}
& \includegraphics[width=1.5cm,height=1.5cm,keepaspectratio]{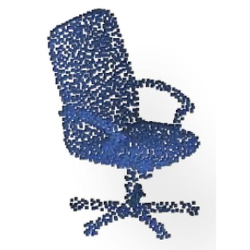} 
& \includegraphics[width=1.5cm,height=1.5cm,keepaspectratio]{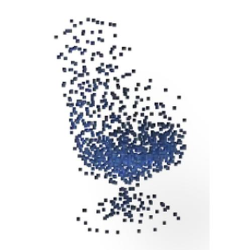} 
& \includegraphics[width=1.5cm,height=1.5cm,keepaspectratio]{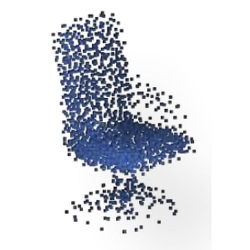} 
& \includegraphics[width=1.5cm,height=1.5cm,keepaspectratio]{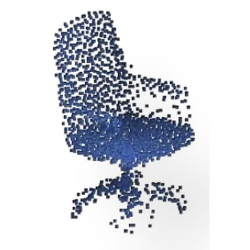} 
& \includegraphics[width=1.5cm,height=1.5cm,keepaspectratio]{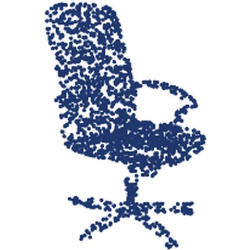}
& \includegraphics[width=1.5cm,height=1.5cm,keepaspectratio]{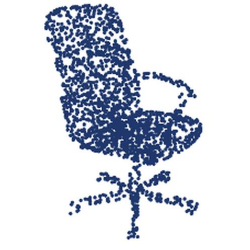} \\

\includegraphics[width=1.5cm,height=1.5cm,keepaspectratio]{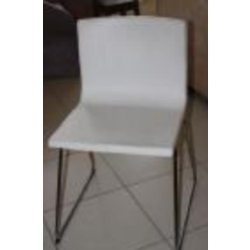}
& \includegraphics[width=1.5cm,height=1.5cm,keepaspectratio]{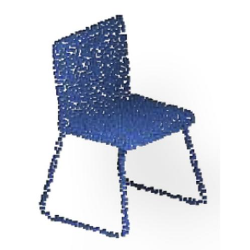} 
& \includegraphics[width=1.5cm,height=1.5cm,keepaspectratio]{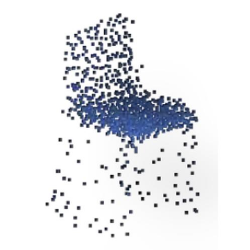} 
& \includegraphics[width=1.5cm,height=1.5cm,keepaspectratio]{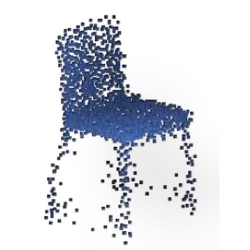} 
& \includegraphics[width=1.5cm,height=1.5cm,keepaspectratio]{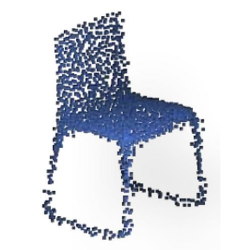} 
& \includegraphics[width=1.5cm,height=1.5cm,keepaspectratio]{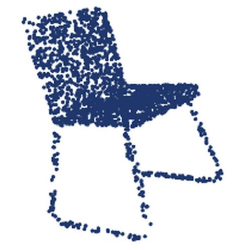}
& \includegraphics[width=1.5cm,height=1.5cm,keepaspectratio]{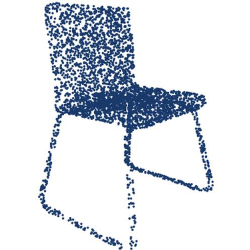} \\

\includegraphics[width=1.5cm,height=1.5cm,keepaspectratio]{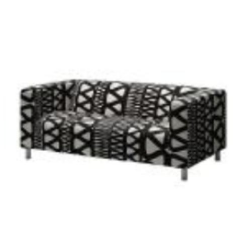}
& \includegraphics[width=1.5cm,height=1.5cm,keepaspectratio]{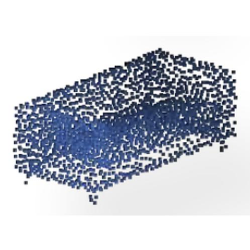} 
& \includegraphics[width=1.5cm,height=1.5cm,keepaspectratio]{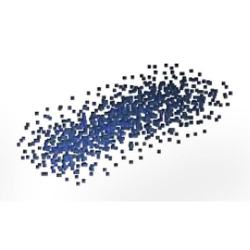} 
& \includegraphics[width=1.5cm,height=1.5cm,keepaspectratio]{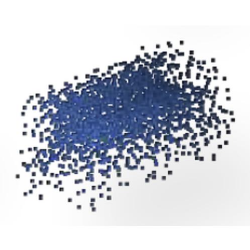} 
& \includegraphics[width=1.5cm,height=1.5cm,keepaspectratio]{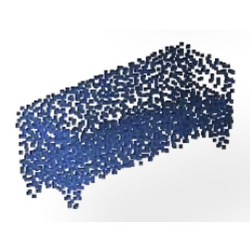} 
& \includegraphics[width=1.5cm,height=1.5cm,keepaspectratio]{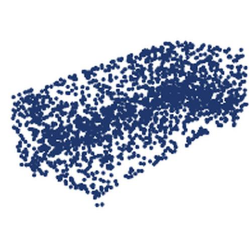}
& \includegraphics[width=1.5cm,height=1.5cm,keepaspectratio]{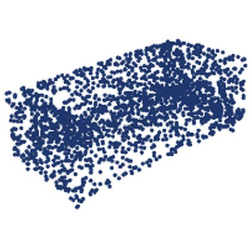} \\

\includegraphics[width=1.5cm,height=1.5cm,keepaspectratio]{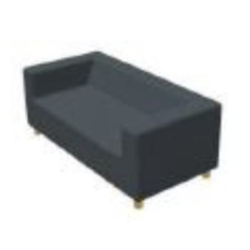}
& \includegraphics[width=1.5cm,height=1.5cm,keepaspectratio]{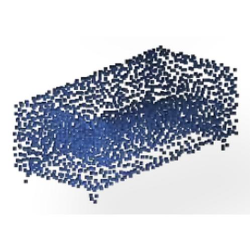} 
& \includegraphics[width=1.5cm,height=1.5cm,keepaspectratio]{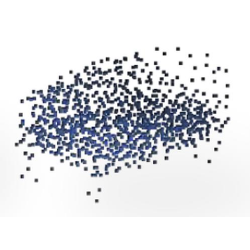} 
& \includegraphics[width=1.5cm,height=1.5cm,keepaspectratio]{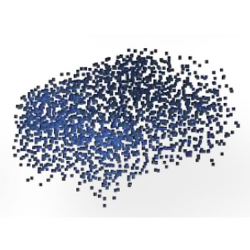} 
& \includegraphics[width=1.5cm,height=1.5cm,keepaspectratio]{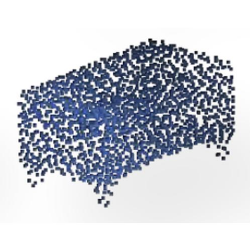} 
& \includegraphics[width=1.5cm,height=1.5cm,keepaspectratio]{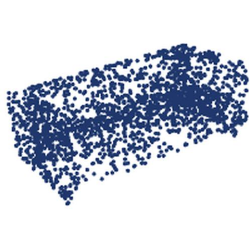}
& \includegraphics[width=1.5cm,height=1.5cm,keepaspectratio]{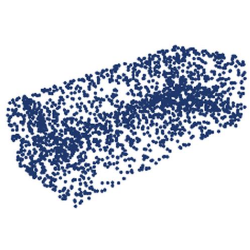} \\

\includegraphics[width=1.5cm,height=1.5cm,keepaspectratio]{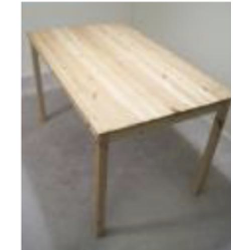}
& \includegraphics[width=1.5cm,height=1.5cm,keepaspectratio]{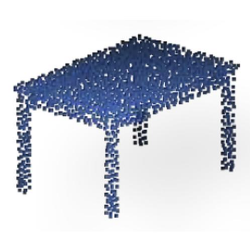} 
& \includegraphics[width=1.5cm,height=1.5cm,keepaspectratio]{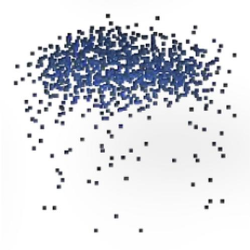} 
& \includegraphics[width=1.5cm,height=1.5cm,keepaspectratio]{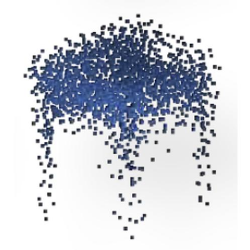} 
& \includegraphics[width=1.5cm,height=1.5cm,keepaspectratio]{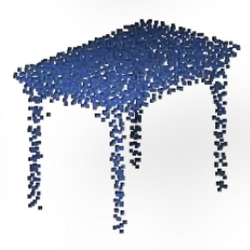} 
& \includegraphics[width=1.5cm,height=1.5cm,keepaspectratio]{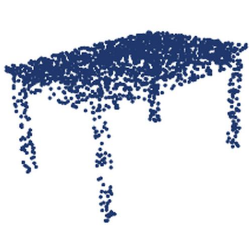}
& \includegraphics[width=1.5cm,height=1.5cm,keepaspectratio]{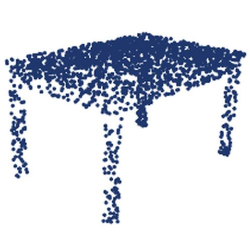} \\

\includegraphics[width=1.5cm,height=1.5cm,keepaspectratio]{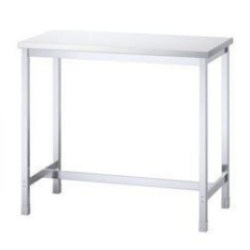}
& \includegraphics[width=1.5cm,height=1.5cm,keepaspectratio]{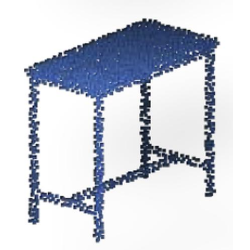} 
& \includegraphics[width=1.5cm,height=1.5cm,keepaspectratio]{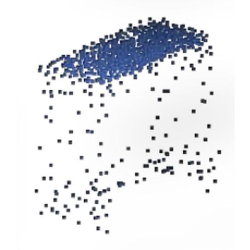} 
& \includegraphics[width=1.5cm,height=1.5cm,keepaspectratio]{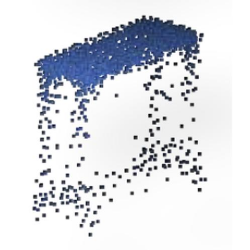} 
& \includegraphics[width=1.5cm,height=1.5cm,keepaspectratio]{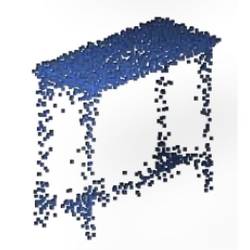} 
& \includegraphics[width=1.5cm,height=1.5cm,keepaspectratio]{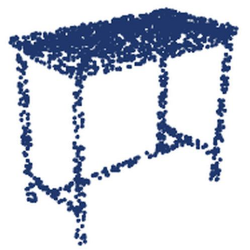}
& \includegraphics[width=1.5cm,height=1.5cm,keepaspectratio]{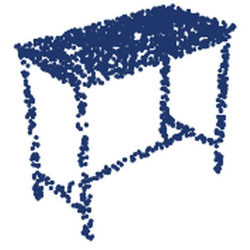} \\
\end{tabular}
}
\end{table*}

%% file: tables/allmetricspix3d.tex
\begin{table*}[]
\caption{Demonstration of the 3DMPE algorithm (last 2 columns) for 3 different 3D models from the Pix3D dataset. For all experiments, 2048 points are used with 4 viewpoints. Projections are created with raytracing algorithm.}
\label{tab:allmetricspix3d}
\centering
\resizebox{\textwidth}{!}{%
\begin{tabular}{|c|cccccc|cccccc|c|}
\hline
\multirow{2}{*}{Category} & \multicolumn{6}{c|}{Chamfer}           & \multicolumn{6}{c|}{EMD} \\ \cline{2-13}
            & Baseline \cite{mandikal20183D} & PSGN \cite{fan2017apoint} & 3D-LMNet \cite{mandikal20183D} & Our BL & 3DMPE(F) & 3DMPE(V) & Baseline & PSGN  & 3D-LMNet & Our BL & 3DMPE(F) & 3DMPE(V) \\ \hline
chair       & 7.52     & 8.05                      & 7.35                           & 2.50  & \textbf{1.21} & 2.15 & 11.17   & 12.55  & 9.14  & 1.70  & \textbf{1.53} & 1.74 \\
sofa        & 8.65     & 8.45                      & 8.18                           & 3.19  & \textbf{0.53} & 0.95 & 8.87    & 9.16   & 7.22  & 2.48  & \textbf{1.01} & 1.01 \\
table       & 11.23    & 10.82                     & 11.20                          & 4.36  & \textbf{3.19} & 3.98 & 15.71   & 15.16  & 12.73 & 4.24  & \textbf{3.65} & 4.11 \\ \hline
mean        & 9.13     & 9.11                      & 8.91                           & 3.35  & \textbf{1.64} & 2.36 & 11.92   & 12.29  & 9.70  & 2.81  & \textbf{2.06} & 2.29 \\ \hline
\end{tabular}%
}
\label{tab:pix3d_shapenetmetrics}
\end{table*}

%% file: 3DMPE.bib
@String(CVPR  = {IEEE Conf. Comput. Vis. Pattern Recog.})

@String(ICCV  = {Int. Conf. Comput. Vis.})

@String(ECCV  = {Eur. Conf. Comput. Vis.})

@String(NeurIPS = {Adv. Neural Inform. Process. Syst.})

@String(ICML  = {Int. Conf. Mach. Learn.})

@String(BMVC  = {Brit. Mach. Vis. Conf.})

@String(AAAI  = {AAAI})

@String(CVPR  = {CVPR})

@String(ICCV  = {ICCV})

@String(ECCV  = {ECCV})

@String(NeurIPS = {NeurIPS})

@String(ICML  = {ICML})

@String(BMVC  =	{BMVC})

@inproceedings{deul2010raytracing,
  title={Raytracing point clouds using geometric algebra},
  author={Deul, Crispin and Burger, Michael and Hildenbrand, Dietmar and Koch, Andreas},
  booktitle={submitted to the proceedings of the GraVisMa workshop, Plzen},
  year={2010},
  organization={Citeseer}
}

@article{weisstein1999triangle,
  title={Triangle point picking},
  author={Weisstein, Eric W},
  journal={https://mathworld. wolfram. com/},
  year={1999},
  publisher={Wolfram Research, Inc.}
}

@article{yao2007early,
  title={On early stopping in gradient descent learning},
  author={Yao, Yuan and Rosasco, Lorenzo and Caponnetto, Andrea},
  journal={Constructive Approximation},
  volume={26},
  number={2},
  pages={289--315},
  year={2007},
  publisher={Springer}
}

@article{zhu2022pde,
  title={PDE-Based 3D Surface Reconstruction from Multi-View 2D Images},
  author={Zhu, Zaiping and Iglesias, Andres and Zhou, Liqi and You, Lihua and Zhang, Jianjun},
  journal={Mathematics},
  volume={10},
  number={4},
  pages={542},
  year={2022},
  publisher={MDPI}
}

@book{Shirley2000Realistic,
  author    = {Peter Shirley},
  title     = {Realistic ray tracing},
  publisher = {A {K} Peters},
  year      = {2000},
  isbn      = {978-1-56881-110-9},
  bibsource = {dblp computer science bibliography, https://dblp.org}
}

@article{Robbins1951AStochastic,
author = {Herbert Robbins and Sutton Monro},
title = {{A Stochastic Approximation Method}},
volume = {22},
journal = {The Annals of Mathematical Statistics},
number = {3},
publisher = {Institute of Mathematical Statistics},
pages = {400 -- 407},
year = {1951}
}

@inproceedings{sift,
  author    = {David G. Lowe},
  title     = {Object Recognition from Local Scale-Invariant Features},
  booktitle = {Proceedings of the International Conference on Computer Vision, Kerkyra,
               Corfu, Greece, September 20-25, 1999},
  pages     = {1150--1157},
  publisher = {{IEEE} Computer Society},
  year      = {1999}
}

@inproceedings{realworlddataset,
  author       = {Rakesh Shrestha and
                  Siqi Hu and
                  Minghao Gou and
                  Ziyuan Liu and
                  Ping Tan},
  editor       = {Shai Avidan and
                  Gabriel J. Brostow and
                  Moustapha Ciss{\'{e}} and
                  Giovanni Maria Farinella and
                  Tal Hassner},
  title        = {A Real World Dataset for Multi-view 3D Reconstruction},
  booktitle    = {Computer Vision - {ECCV} 2022 - 17th European Conference, Tel Aviv,
                  Israel, October 23-27, 2022, Proceedings, Part {VIII}},
  series       = {Lecture Notes in Computer Science},
  volume       = {13668},
  pages        = {56--73},
  publisher    = {Springer},
  year         = {2022}
}

@article{ozyesil2017survey,
  author    = {Onur {\"{O}}zyesil and
               Vladislav Voroninski and
               Ronen Basri and
               Amit Singer},
  title     = {A survey of structure from motion},
  journal   = {Acta Numer.},
  volume    = {26},
  pages     = {305--364},
  year      = {2017}
}

@Manual{blender,
   title = {Blender - a 3D modelling and rendering package},
   author = {Blender Online Community},
   organization = {Blender Foundation},
   address = {Stichting Blender Foundation, Amsterdam},
   year = {2018},
   url = {http://www.blender.org}
 }

@misc{colmap_datasets,
  author = {Johannes L. Schoenberger},
  title = {4 Datasets in Colmap Documentation},
  year = {2023},
  url = {https://colmap.github.io/datasets.html},
  accessed = {2023-08-07}
}

@inproceedings{schonberger2016structure,
  author       = {Johannes L. Sch{\"{o}}nberger and
                  Jan{-}Michael Frahm},
  title        = {Structure-from-Motion Revisited},
  booktitle    = CVPR,
  pages        = {4104--4113},
  year         = {2016}
}

@inproceedings{wu20153d,
  author    = {Zhirong Wu and
               Shuran Song and
               Aditya Khosla and
               Fisher Yu and
               Linguang Zhang and
               Xiaoou Tang and
               Jianxiong Xiao},
  title     = {3D ShapeNets: {A} deep representation for volumetric shapes},
  booktitle = {{IEEE} Conference on Computer Vision and Pattern Recognition, {CVPR}
               2015, Boston, MA, USA, June 7-12, 2015},
  pages     = {1912--1920},
  publisher = {{IEEE} Computer Society},
  year      = {2015}
}

@article{shapenet2015,
  author    = {Angel X. Chang and
               Thomas A. Funkhouser and
               Leonidas J. Guibas and
               Pat Hanrahan and
               Qi{-}Xing Huang and
               Zimo Li and
               Silvio Savarese and
               Manolis Savva and
               Shuran Song and
               Hao Su and
               Jianxiong Xiao and
               Li Yi and
               Fisher Yu},
  title     = {ShapeNet: An Information-Rich 3D Model Repository},
  journal   = {CoRR},
  volume    = {abs/1512.03012},
  year      = {2015}
}

@article{hossain2021multi,
  author    = {Md. Iqbal Hossain and
               Vahan Huroyan and
               Stephen G. Kobourov and
               Raymundo Navarrete},
  title     = {Multi-Perspective, Simultaneous Embedding},
  journal   = {{IEEE} Trans. Vis. Comput. Graph.},
  volume    = {27},
  number    = {2},
  pages     = {1569--1579},
  year      = {2021}
}

@inproceedings{mandikal20183D,
  author    = {Priyanka Mandikal and
               Navaneet K. L. and
               Mayank Agarwal and
               Venkatesh Babu Radhakrishnan},
  title     = {3D-LMNet: Latent Embedding Matching for Accurate and Diverse 3D Point
               Cloud Reconstruction from a Single Image},
  booktitle = {British Machine Vision Conference 2018, {BMVC} 2018, Newcastle, UK,
               September 3-6, 2018},
  pages     = {55},
  publisher = {{BMVA} Press},
  year      = {2018}
}

@inproceedings{kulon2019single,
  author    = {Dominik Kulon and
               Haoyang Wang and
               Riza Alp G{\"{u}}ler and
               Michael M. Bronstein and
               Stefanos Zafeiriou},
  title     = {Single Image 3D Hand Reconstruction with Mesh Convolutions},
  booktitle = {30th British Machine Vision Conference 2019, {BMVC} 2019, Cardiff,
               UK, September 9-12, 2019},
  pages     = {45},
  publisher = {{BMVA} Press},
  year      = {2019}
}

@article{rubner2000earth,
  author    = {Yossi Rubner and
               Carlo Tomasi and
               Leonidas J. Guibas},
  title     = {The Earth Mover's Distance as a Metric for Image Retrieval},
  journal   = {Int. J. Comput. Vis.},
  volume    = {40},
  number    = {2},
  pages     = {99--121},
  year      = {2000}
}

@inproceedings{michalkiewicz2020asimple,
  author    = {Mateusz Michalkiewicz and
               Eugene Belilovsky and
               Mahsa Baktashmotlagh and
               Anders P. Eriksson},
  title     = {A Simple and Scalable Shape Representation for 3D Reconstruction},
  booktitle = {31st British Machine Vision Conference, {BMVC} 2020, Virtual Event, UK, September 7-10, 2020},
  publisher = {{BMVA} Press},
  year      = {2020}
}

@inproceedings{klokov2019probabilistic,
  author    = {Roman Klokov and
               Jakob Verbeek and
               Edmond Boyer},
  title     = {Probabilistic Reconstruction Networks for 3D Shape Inference from
               a Single Image},
  booktitle = {30th British Machine Vision Conference, {BMVC} 2019, Cardiff, UK, September 9-12, 2019},
  pages     = {165},
  publisher = {{BMVA} Press},
  year      = {2019}
}

@inproceedings{choy20163dr2n2,
  author    = {Christopher B. Choy and
               Danfei Xu and
               JunYoung Gwak and
               Kevin Chen and
               Silvio Savarese},
  title     = {3D-R2N2: {A} Unified Approach for Single and Multi-view 3D Object Reconstruction},
  booktitle = {Computer Vision - {ECCV} 2016 - 14th European Conference, Amsterdam,
               The Netherlands, October 11-14, 2016, Proceedings, Part {VIII}},
  series    = {Lecture Notes in Computer Science},
  volume    = {9912},
  pages     = {628--644},
  publisher = {Springer},
  year      = {2016}
}

@inproceedings{wu2016learning,
  author    = {Jiajun Wu and
               Chengkai Zhang and
               Tianfan Xue and
               Bill Freeman and
               Josh Tenenbaum},
  title     = {Learning a Probabilistic Latent Space of Object Shapes via 3D Generative-Adversarial
               Modeling},
  booktitle = {Advances in Neural Information Processing Systems 29: Annual Conference
               on Neural Information Processing Systems, Barcelona, Spain},
  pages     = {82--90},
  year      = {2016}
}

@INPROCEEDINGS{932923,
  author={Neubert, J. and Hammond, T. and Guse, N. and Do, Y. and Hu, Y. and Ferrier, N.},
  booktitle={Proceedings 2001 ICRA. IEEE International Conference on Robotics and Automation (Cat. No.01CH37164)}, 
  title={Automatic training of a neural net for active stereo 3D reconstruction}, 
  year={2001}
}

@article{Fischler1981RANSAC,
author = {Fischler, Martin A. and Bolles, Robert C.},
title = {Random Sample Consensus: A Paradigm for Model Fitting with Applications to Image Analysis and Automated Cartography},
year = {1981},
issue_date = {June 1981},
publisher = {Association for Computing Machinery},
address = {New York, NY, USA},
volume = {24},
number = {6},
issn = {0001-0782},
url = {https://doi.org/10.1145/358669.358692},
doi = {10.1145/358669.358692},
journal = {Commun. ACM},
month = {jun},
pages = {381–395},
numpages = {15}
}

@ARTICLE{8751135,
  author={Miclea, Vlad-Cristian and Nedevschi, Sergiu},
  journal={IEEE Transactions on Intelligent Transportation Systems}, 
  title={Real-Time Semantic Segmentation-Based Stereo Reconstruction}, 
  year={2020},
  volume={21},
  number={4},
  pages={1514-1524},
  doi={10.1109/TITS.2019.2913883}
}

@INPROCEEDINGS{1716121,
  author={Wen-Chang Cheng},
  booktitle={The 2006 IEEE International Joint Conference on Neural Network Proceedings}, 
  title={Neural-Network-Based Photometric Stereo for 3D Surface Reconstruction}, 
  year={2006},
  volume={},
  number={},
  pages={404-410},
  doi={10.1109/IJCNN.2006.246710}
}

@article{hamid2021stereo,
  title={Stereo matching algorithm based on hybrid convolutional neural network and directional intensity difference},
  author={Hamid, Mohd Saad and Manap, N and Hamzah, Rostam Affendi and Kadmin, Ahmad Fauzan},
  journal={Artificial intelligence (AI)},
  volume={14},
  pages={16},
  year={2021}
}

@article{ALTINGOVDE2022113460,
author={Okan Altingövde and Anastasiia Mishchuk and Gulnaz Ganeeva and Emad Oveisi and Cecile Hebert and Pascal Fua},
title = {3D reconstruction of curvilinear structures with stereo matching deep convolutional neural networks},
journal = {Ultramicroscopy},
volume = {234},
pages = {113460},
year = {2022}
}

@ARTICLE{8613838,
  author={Zhu, Hao and Jiao, Licheng and Ma, Wenping and Liu, Fang and Zhao, Wei},
  journal={IEEE Transactions on Neural Networks and Learning Systems}, 
  title={A Novel Neural Network for Remote Sensing Image Matching}, 
  year={2019}
}

@InProceedings{Seki_2017_CVPR,
author = {Seki, Akihito and Pollefeys, Marc},
title = {SGM-Nets: Semi-Global Matching With Neural Networks},
booktitle = {Proceedings of the IEEE Conference on Computer Vision and Pattern Recognition (CVPR)},
month = {July},
year = {2017}
}

@article{Zbontar2015Stereo,
  author    = {Jure Zbontar and
               Yann LeCun},
  title     = {Stereo Matching by Training a Convolutional Neural Network to Compare
               Image Patches},
  journal   = {CoRR},
  volume    = {abs/1510.05970},
  year      = {2015}
}

@InProceedings{Rocco_2017_CVPR,
author = {Rocco, Ignacio and Arandjelovic, Relja and Sivic, Josef},
title = {Convolutional Neural Network Architecture for Geometric Matching},
booktitle = {Proceedings of the IEEE Conference on Computer Vision and Pattern Recognition (CVPR)},
month = {July},
year = {2017}
}

@inproceedings{fan2017apoint,
  author    = {Haoqiang Fan and
               Hao Su and
               Leonidas J. Guibas},
  title     = {A Point Set Generation Network for 3D Object Reconstruction from a
               Single Image},
  booktitle = {2017 {IEEE} Conference on Computer Vision and Pattern Recognition,
               {CVPR} 2017, Honolulu, HI, USA, July 21-26, 2017},
  pages     = {2463--2471},
  publisher = {{IEEE} Computer Society},
  year      = {2017}
}

@inproceedings{yu2021local,
  author    = {Ziwei Yu and
               Linlin Yang and
               Shicheng Chen and
               Angela Yao},
  title     = {Local and Global Point Cloud Reconstruction for 3D Hand Pose Estimation},
  booktitle = {32nd British Machine Vision Conference 2021, {BMVC} 2021, Online,
               November 22-25, 2021},
  pages     = {388},
  publisher = {{BMVA} Press},
  year      = {2021}
}

@article{haming2010the,
  author    = {Klaus H{\"{a}}ming and
               Gabriele Peters},
  title     = {The structure-from-motion reconstruction pipeline - a survey with
               focus on short image sequences},
  journal   = {Kybernetika},
  volume    = {46},
  number    = {5},
  pages     = {926--937},
  year      = {2010}
}

@inproceedings{pix3d,
  author    = {Xingyuan Sun and
               Jiajun Wu and
               Xiuming Zhang and
               Zhoutong Zhang and
               Chengkai Zhang and
               Tianfan Xue and
               Joshua B. Tenenbaum and
               William T. Freeman},
  title     = {Pix3D: Dataset and Methods for Single-Image 3D Shape Modeling},
  booktitle = {2018 {IEEE} Conference on Computer Vision and Pattern Recognition,
               {CVPR} 2018, Salt Lake City, UT, USA, June 18-22, 2018},
  pages     = {2974--2983},
  publisher = {Computer Vision Foundation / {IEEE} Computer Society},
  year      = {2018}
}

@article{Zhang2020,
  author    = {Juyong Zhang and
               Yuxin Yao and
               Bailin Deng},
  title     = {Fast and Robust Iterative Closest Point},
  journal   = {{IEEE} Trans. Pattern Anal. Mach. Intell.},
  volume    = {44},
  number    = {7},
  pages     = {3450--3466},
  year      = {2022}
}

@article{10.2307/2132388,
  author    = {G. W. Stewart},
  title     = {On the Early History of the Singular Value Decomposition},
  journal   = {{SIAM} Rev.},
  volume    = {35},
  number    = {4},
  pages     = {551--566},
  year      = {1993}
}

@article{bianco2018evaluating,
  author    = {Simone Bianco and
               Gianluigi Ciocca and
               Davide Marelli},
  title     = {Evaluating the Performance of Structure from Motion Pipelines},
  journal   = {J. Imaging},
  volume    = {4},
  number    = {8},
  pages     = {98},
  year      = {2018}
}

@article{han2021image,
  author    = {Xian{-}Feng Han and
               Hamid Laga and
               Mohammed Bennamoun},
  title     = {Image-Based 3D Object Reconstruction: State-of-the-Art and Trends
               in the Deep Learning Era},
  journal   = {{IEEE} Trans. Pattern Anal. Mach. Intell.},
  volume    = {43},
  number    = {5},
  pages     = {1578--1604},
  year      = {2021}
}

@inproceedings{jiahui2020learning,
  author    = {Jiahui Lei and
               Srinath Sridhar and
               Paul Guerrero and
               Minhyuk Sung and
               Niloy J. Mitra and
               Leonidas J. Guibas},
  title     = {Pix2Surf: Learning Parametric 3D Surface Models of Objects from Images},
  booktitle = {Computer Vision - {ECCV} 2020 - 16th European Conference, Glasgow,
               UK, August 23-28, 2020, Proceedings, Part {XVIII}},
  series    = {Lecture Notes in Computer Science},
  volume    = {12363},
  pages     = {121--138},
  publisher = {Springer},
  year      = {2020}
}

@article{berger2017survey,
  author    = {Matthew Berger and
               Andrea Tagliasacchi and
               Lee M. Seversky and
               Pierre Alliez and
               Ga{\"{e}}l Guennebaud and
               Joshua A. Levine and
               Andrei Sharf and
               Cl{\'{a}}udio T. Silva},
  title     = {A Survey of Surface Reconstruction from Point Clouds},
  journal   = {Comput. Graph. Forum},
  volume    = {36},
  number    = {1},
  pages     = {301--329},
  year      = {2017}
}

@book{hartley2006multiple,
  author    = {Andrew Harltey and
               Andrew Zisserman},
  title     = {Multiple view geometry in computer vision {(2.} ed.)},
  publisher = {Cambridge University Press},
  year      = {2006}
}

@inproceedings{girdhar2016learning,
  author    = {Rohit Girdhar and
               David F. Fouhey and
               Mikel Rodriguez and
               Abhinav Gupta},
  title     = {Learning a Predictable and Generative Vector Representation for Objects},
  booktitle = {Computer Vision - {ECCV} 2016 - 14th European Conference, Amsterdam, The Netherlands, October 11-14, 2016, Proceedings, Part {VI}},
  series    = {Lecture Notes in Computer Science},
  volume    = {9910},
  pages     = {484--499},
  publisher = {Springer},
  year      = {2016}
}

@inproceedings{tatarchenko2017octree,
  author    = {Maxim Tatarchenko and
               Alexey Dosovitskiy and
               Thomas Brox},
  title     = {Octree Generating Networks: Efficient Convolutional Architectures
               for High-resolution 3D Outputs},
  booktitle = {{IEEE} International Conference on Computer Vision, {ICCV} 2017, Venice,
               Italy, October 22-29, 2017},
  pages     = {2107--2115},
  publisher = {{IEEE} Computer Society},
  year      = {2017}
}

@inproceedings{kar2017learning,
  author    = {Abhishek Kar and
               Christian H{\"{a}}ne and
               Jitendra Malik},
  title     = {Learning a Multi-View Stereo Machine},
  booktitle = {Advances in Neural Information Processing Systems 30: Annual Conference on Neural Information Processing Systems 2017, December 4-9, 2017, Long Beach, CA, {USA}},
  pages     = {365--376},
  year      = {2017}
}

@inproceedings{tulsiani2017multi,
  author    = {Shubham Tulsiani and
               Tinghui Zhou and
               Alexei A. Efros and
               Jitendra Malik},
  title     = {Multi-view Supervision for Single-View Reconstruction via Differentiable
               Ray Consistency},
  booktitle = {2017 {IEEE} Conference on Computer Vision and Pattern Recognition,
               {CVPR} 2017, Honolulu, HI, USA, July 21-26, 2017},
  pages     = {209--217},
  publisher = {{IEEE} Computer Society},
  year      = {2017}
}

@inproceedings{insafutdinov2018unsupervised,
  author    = {Eldar Insafutdinov and
               Alexey Dosovitskiy},
  title     = {Unsupervised Learning of Shape and Pose with Differentiable Point Clouds},
  booktitle = {Advances in Neural Information Processing Systems 31: Annual Conference
               on Neural Information Processing Systems 2018, NeurIPS 2018, December
               3-8, 2018, Montr{\'{e}}al, Canada},
  pages     = {2807--2817},
  year      = {2018}
}

@inproceedings{lin2018learning,
  author    = {Chen{-}Hsuan Lin and
               Chen Kong and
               Simon Lucey},
  title     = {Learning Efficient Point Cloud Generation for Dense 3D Object Reconstruction},
  booktitle = {Proceedings of the Thirty-Second {AAAI} Conference on Artificial Intelligence,
               (AAAI-18), the 30th innovative Applications of Artificial Intelligence
               (IAAI-18), and the 8th {AAAI} Symposium on Educational Advances in
               Artificial Intelligence (EAAI-18), New Orleans, Louisiana, USA, February
               2-7, 2018},
  pages     = {7114--7121},
  publisher = {{AAAI} Press},
  year      = {2018}
}

@inproceedings{sridhar2019multiview,
  author    = {Srinath Sridhar and
               Davis Rempe and
               Julien Valentin and
               Sofien Bouaziz and
               Leonidas J. Guibas},
  title     = {Multiview Aggregation for Learning Category-Specific Shape Reconstruction},
  booktitle = {Advances in Neural Information Processing Systems 32, Vancouver, BC, Canada},
  pages     = {2348--2359},
  year      = {2019}
}

@inproceedings{mandikal2019dense,
  author    = {Priyanka Mandikal and
               Venkatesh Babu Radhakrishnan},
  title     = {Dense 3D Point Cloud Reconstruction Using a Deep Pyramid Network},
  booktitle = {{IEEE} Winter Conference on Applications of Computer Vision, {WACV}
               2019, Waikoloa Village, HI, USA, January 7-11, 2019},
  pages     = {1052--1060},
  publisher = {{IEEE}},
  year      = {2019}
}

@article{shepard1962analysis,
    title={The analysis of proximities: multidimensional scaling with an unknown distance function},
    author={Shepard, Roger N},
    journal={Psychometrika},
    volume={27},
    number={2},
    pages={125--140},
    year={1962},
    publisher={Springer}
}

@inproceedings{malitsky2020adaprive,
  author    = {Yura Malitsky and
               Konstantin Mishchenko},
  title     = {Adaptive Gradient Descent without Descent},
  booktitle = {Proceedings of the 37th International Conference on Machine Learning,
               {ICML} 2020, 13-18 July 2020, Virtual Event},
  series    = {Proceedings of Machine Learning Research},
  volume    = {119},
  pages     = {6702--6712},
  publisher = {{PMLR}},
  year      = {2020}
}

@article{zheng2019graph,
  author    = {Jonathan X. Zheng and
               Samraat Pawar and
               Dan F. M. Goodman},
  title     = {Graph Drawing by Stochastic Gradient Descent},
  journal   = {{IEEE} Trans. Vis. Comput. Graph.},
  volume    = {25},
  number    = {9},
  pages     = {2738--2748},
  year      = {2019}
}

@inproceedings{wang2024dust3r,
  author       = {Shuzhe Wang and
                  Vincent Leroy and
                  Yohann Cabon and
                  Boris Chidlovskii and
                  J{\'{e}}r{\^{o}}me Revaud},
  title        = {DUSt3R: Geometric 3D Vision Made Easy},
  booktitle    = {{IEEE/CVF} Conference on Computer Vision and Pattern Recognition,
                  {CVPR} 2024, Seattle, WA, USA, June 16-22, 2024},
  pages        = {20697--20709},
  publisher    = {{IEEE}},
  year         = {2024}
}

@inproceedings{leroy2024mast3r,
  author       = {Vincent Leroy and
                  Yohann Cabon and
                  J{\'{e}}r{\^{o}}me Revaud},
  title        = {Grounding Image Matching in 3D with MASt3R},
  booktitle    = {Computer Vision - {ECCV} 2024 - 18th European Conference, Milan, Italy,
                  September 29-October 4, 2024, Proceedings, Part {LXXII}},
  series       = {Lecture Notes in Computer Science},
  volume       = {15130},
  pages        = {71--91},
  publisher    = {Springer},
  year         = {2024}
}

@inproceedings{wang2025vggt,
  author       = {Jianyuan Wang and
                  Minghao Chen and
                  Nikita Karaev and
                  Andrea Vedaldi and
                  Christian Rupprecht and
                  David Novotn{\'{y}}},
  title        = {{VGGT:} Visual Geometry Grounded Transformer},
  booktitle    = {{IEEE/CVF} Conference on Computer Vision and Pattern Recognition,
                  {CVPR} 2025, Nashville, TN, USA, June 11-15, 2025},
  pages        = {5294--5306},
  publisher    = {Computer Vision Foundation / {IEEE}},
  year         = {2025}
}
